\documentclass[10pt,twocolumn,letterpaper]{article}

\usepackage[pagenumbers]{cvpr} 

\definecolor{cvprblue}{rgb}{0.21,0.49,0.74}
\usepackage[pagebackref,breaklinks,colorlinks,allcolors=cvprblue]{hyperref}
\usepackage{hyphenat}

\usepackage{graphicx}
\usepackage{adjustbox}
\usepackage{array}
\usepackage{xcolor}
\usepackage{longtable}
\usepackage{tcolorbox}
\tcbuselibrary{breakable}

\def\paperID{***} 
\def\confName{CVPR}
\def\confYear{2026}

\title{SigVLP: Sigmoid Volume-Language Pre-Training for Self-Supervised CT-Volume Adaptive Representation Learning}

\author{
Jiayi Wang$^{1}$ \quad
Hadrien Reynaud$^{1}$ \quad
Ibrahim Ethem Hamamci$^{2,3,4}$ \quad
Sezgin Er$^{2,4}$ \\
Suprosanna Shit$^{2,3}$ \quad
Bjoern Menze$^{2,3}$ \quad
Bernhard Kainz$^{5,1}$ \\[2mm]
$^{1}$Friedrich-Alexander University Erlangen-Nürnberg, Germany \\
$^{2}$Department of Quantitative Biomedicine, University of Zurich, Switzerland \\
$^{3}$ETH AI Center, ETH Zurich, Switzerland \\
$^{4}$International School of Medicine, Istanbul Medipol University, Turkey \\
$^{5}$Department of Computing, Imperial College London, UK \\[2mm]
{\tt\small \{jiayi.w.wang, hadrien.reynaud\}@fau.de} \\
{\tt\small \{ibrahim.hamamci, bjoern.menze, suprosanna.shit\}@uzh.ch} \\
{\tt\small sezgin.er@std.medipol.edu.tr \quad b.kainz@imperial.ac.uk}
}

\begin{document}
\maketitle
\begin{abstract}

Large-scale, volumetric medical imaging datasets typically aggregate scans from different vendors and devices, resulting in highly variable resolution, slice thicknesses, and numbers of slices per study. Consequently, training representation models usually requires cropping or interpolating along the z-axis to obtain fixed-size blocks, which inevitably causes information loss.
We propose a new training approach to overcome this limitation. Instead of absolute position embeddings, we interpret volumes as sequences of 3D chunks and adopt Rotary Position Embeddings, allowing us to treat the z-axis as an unconstrained temporal dimensions.
Building on this idea, we introduce a new vision-language model: SigVLP.
In SigVLP, we implement Rotary Position Embedding as the positional encoding method, which is applied directly within the attention operation, generating input-conditioned sine and cosine weights on the fly. This design ensures consistent alignment between query and key projections and adapts to any input sizes.
To allow for variable input size during training, we sample Computed Tomography volumes in chunks and pair them with localized organ-wise textual observations. Compared to using entire reports for conditioning, chunkwise alignment provides finer-grained supervision, enabling the model to establish stronger correlations between the text and volume representations, 
thereby improving the precision of text-to-volume alignment.
Our models are trained with the Muon optimizer and evaluated on a diverse set of downstream tasks, including zero-shot abnormality and organ classification, segmentation, and retrieval tasks.

\end{abstract}

\begin{figure}[t]
    \centering
    \includegraphics[width=1.1\linewidth]{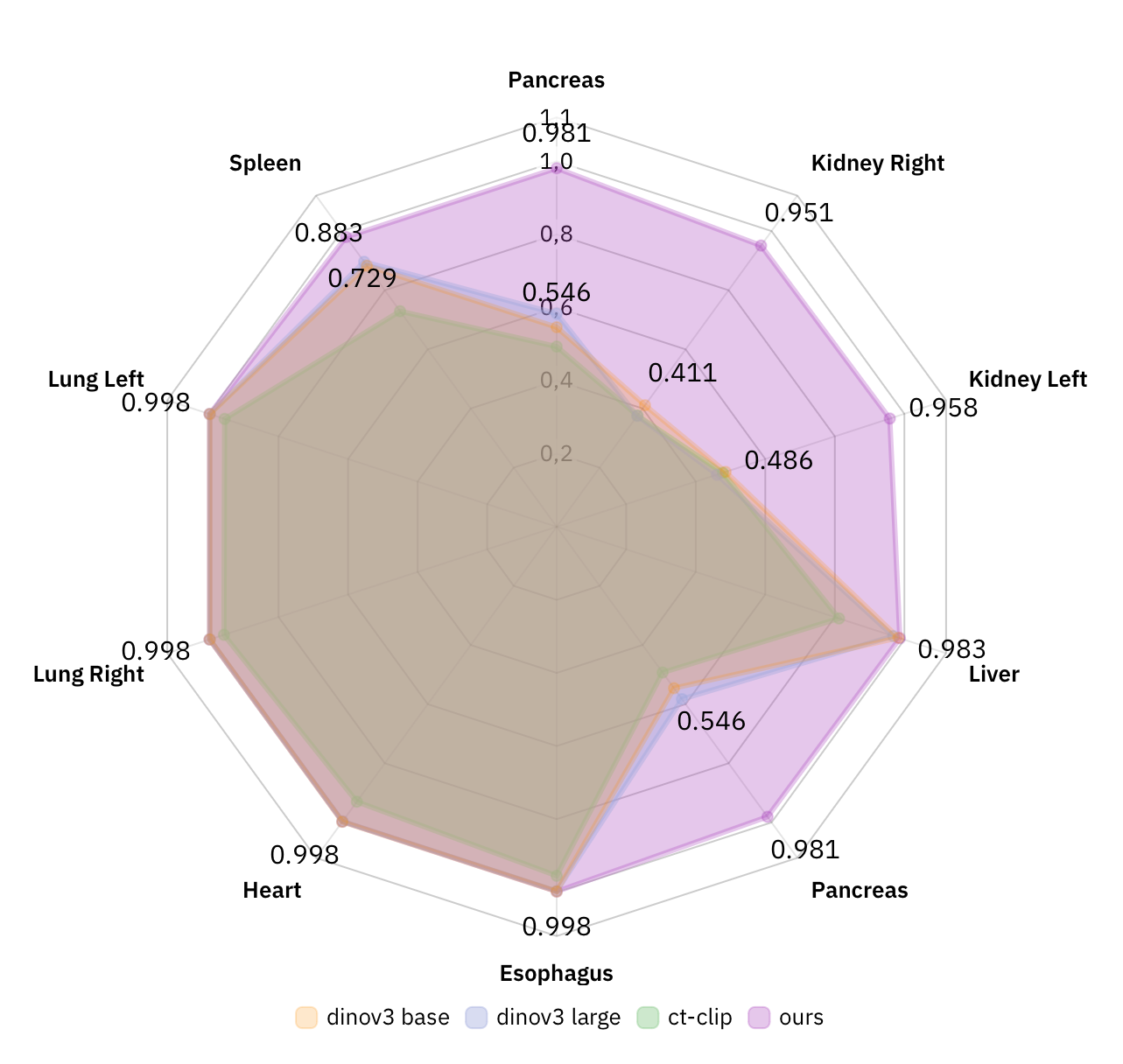}
    \caption{Radar chart of average Precision for organ presence classification using Linear Probes: comparison between SigVLP (ours), CT-Clip and DINOv3 variants (Base, Large) with identical linear probe structures.}
    \label{fig:teaser}
\end{figure}

\begin{figure*}[t]
    \centering
    \includegraphics[width=1.0\linewidth]{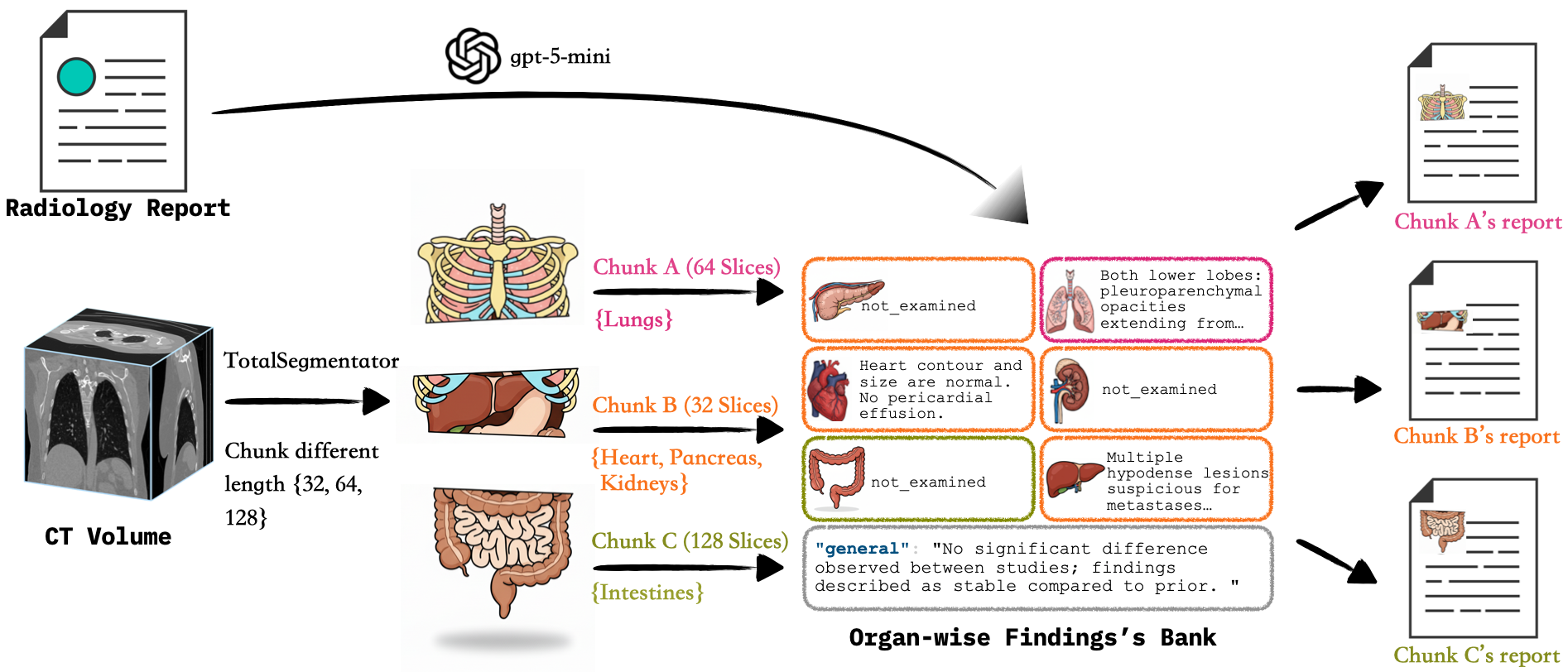}
    \vspace{-2mm}
    \caption{Our approach for Organ-wise Radiology Report Generation: The original CT volume is segmented into organ masks~\cite{wasserthal2023totalsegmentator,xu2025cads}. The volume is then split into blocks of different lengths, where the masks indicate which organs should be included for report generation. Organ-specific findings are extracted with GPT-5 mini to construct an organ findings bank, stored as individual entries. A general description is appended to summarize the entire volume.}
    \label{fig:keypoint}
\end{figure*}

\section{Introduction}
\label{sec:intro} 

Strong universal embeddings are the foundation of modern Vision-Language Models (VLMs) such as CLIP~\cite{radford2021learning}, enabling semantic alignment between visual and textual domains. In the natural image domain, such models have transformed downstream learning by allowing general-purpose representations that can be leveraged across diverse tasks~\cite{li2021albef, li2022blip}, ranging from object detection to image captioning, alleviating the need for expensive training operations. Translating this success to medical imaging, however, remains an open challenge.

Recent medical adaptations such as ConVIRT~\cite{zhang2022convirt}, CT-CLIP~\cite{hamamci2024generatect, hamamci2024ct2rep} and MedCLIP~\cite{wang2022medclip} have attempted to replicate this paradigm for clinical data. Yet, these models are limited by both the scale and structure of available training data and compute scale. For example, CT-CLIP is primarily trained on chest CT data, leading to narrow representation spaces that fail to generalize across modalities, organs, and institutions. Similarly, vision-language pretraining on datasets such as MIMIC-CXR~\cite{johnson2019mimic} captures only 2D chest radiographs with short, templated reports, offering limited representation depth for complex 3D anatomy. As a result, embeddings from these models often encode local visual or textual cues rather than holistic, cross-anatomical semantics that are essential for volumetric reasoning. An alternative line of work avoids full 3D processing by selecting a few representative slices, which improves efficiency at the cost of spatial completeness~\cite{wang2024enhancing}. Additionally, early 3D medical VLM attempts that enforce multi-view consistency still struggle to preserve true volumetric coherence~\cite{liu2023t3d}.

Nevertheless, most of these methods for volumetric medical image analysis relies heavily on large-scale datasets. However, due to strict privacy regulations, fully open-access and standardized datasets remain scarce. For instance, CT-RATE~\cite{hamamci2024developing} offers over 40,000 full volumetric 3D CT scans, in comparison to MIMIC-CXR, which includes nearly 377,000 chest radiographs from more than 65,000 patients, but only in 2D. Although both datasets are widely used in computer vision research, they also illustrate a fundamental challenge: heterogeneity in data representation. Nonetheless, CT-RATE provides complete 3D volumes, whereas MIMIC-CXR includes only cropped (single or dual) 2D views.

In addition to limited medical data, another core obstacle in medical imaging lies in the lack of cross-institutional standardization. Scans  from different vendors and devices often differ significantly in XY spacing, rescale intercept, rescale slope, particularly along the z-axis (slice-wise direction in 3D volumes). 
While normalization methods such as interpolation or cropping can be applied, they systematically lead to data loss. Additionally, most scanners will output fixed-size x-y slices, while the z-axis is much more variable due to patient size and slice thickness.
This limitation is significant when transferring pretrained models from large-scale datasets to organ-specific tasks, such as kidney tumor segmentation (KiTS)~\cite{heller2020kits19}, liver tumor segmentation (LiTS)~\cite{bilic2019lits}, or brain tumor segmentation (BraTS)~\cite{bakas2017brats}. Recent 3D VLMs exemplify this issue by resampling every study to a fixed 3D grid to match architectural constraints~\cite{xin2025med3dvlm,ates2025dcformer}. While it simplifies batching and training, this process leads to the loss of clinically relevant details. 


On the contrary to the need to accommodate variable slices, most methods are limited by reliance on fixed slice numbers. Such restriction is often enforced due to architectural constraints in Transformer-based models~\cite{han2022survey}, especially when using absolute positional embeddings~\cite{kazemnejad2023impact}. These embeddings typically require a fixed-length token sequence, whether for visual or textual input, a limitation that becomes problematic when handling 3D volumes with varying slice counts. In contrast, Rotary Positional Encoding (RoPE)~\cite{su2023roformer} has shown strong potential in handling sequence variability in video-based Transformers, where it effectively encodes temporal relationships without requiring fixed-length inputs. A concurrent direction addresses variable depth by first encoding 2D slices and then aggregating them with a dedicated Z-former, which reduces memory yet reconstructs 3D context only post hoc~\cite{lee2024read}. Our approach maintains within-chunk spatio-temporal structure directly in the volumetric encoder.

Motivated by advancements in video representation learning, particularly from works like CTFlow~\cite{wang2025ctflow} and EchoFlow~\cite{reynaud2025echoflow}, we treat 3D medical volumes as sequences of 3D chunks, analogous to video frames, but extended to 3D. In this setting, Rotary Positional Embeddings are useful, as they assign each token a unique rotationally-invariant positional signature. Rather than using fixed positional indices, RoPE captures relative dependencies across tokens through a rotation matrix applied to query-key projections, enabling better long-range context modeling. This mechanism is particularly powerful in long-sequence modeling tasks, and has been widely adopted in Large Language Models (LLMs) such as LLaMA~\cite{touvron2023llama}, Qwen~\cite{bai2023qwen}, and Kimi-K2~\cite{team2025kimi}. While 1D RoPE suffices for long textual sequences, volumetric settings require 3D positional schemes~\cite{ma20253d}.
%
To address this, various RoPE extensions, like VideoRoPE~\cite{wei2025videorope} VideoRoPE++~\cite{wei2025videoropepp}, TAD-RoPE~\cite{gao2024tcllava} and M-RoPE~\cite{wang2024qwen2vl} have been developed to better model spatial layouts and temporal continuity in 2D and 3D data. 

Building upon these insights, we propose a dynamic, chunk-wise training pipeline that eliminates the fixed-length constraint along the $z$-axis. Instead of aligning the entire visual volume with full-length textual reports, we introduce a fine-grained, region-specific alignment strategy using enhanced, re-structured reports. Our approach extends the SigLIPv2 architecture~\cite{tschannen2025siglip}, which builds on a pairwise sigmoid objective shown to stabilize large-scale vision-language pretraining~\cite{zhai2023sigmoid}.

To account for variations in scan lengths, we train the model on 3D chunks, where the total number of slices varies between 32, 64, and 128. 
We leverage the state-of-the-art Muon optimizer~\cite{liu2025muon,jordan2024muon} to facilitate stable and efficient learning across variable-length inputs. Drawing inspiration from the organ-wise alignment strategies~\cite{shui2025organclip}, we segment entire volumes into smaller anatomical blocks. Correspondingly, we crop the original radiology report into organ-specific segments to ensure meaningful alignment between text and visual chunks. 

For this, we first use a lightweight language model (GPT-5-mini)  
to decompose radiology reports organ-wise.
Next, anatomical segmentation masks, generated by~\cite{wasserthal2023totalsegmentator,xu2025cads}, allow us to accurately determine the anatomical contents of each chunk and match it to the corresponding report fragment. We then reconstruct reports, based on the content of each sampled chunk on the fly.

We evaluate our embedding space on downstream tasks such as 
\textit{zero-shot abnormality classification} and \textit{disease diagnosis}, 
which are standard benchmarks for assessing vision-language alignment. Experimental results show that our method significantly improves performance across these tasks. In summary, our contributions are:

\begin{itemize}
    \item \textbf{On-the-fly subvolume-observationss alignement.} We design a train-time method to allow retrieval of relevant clinical observations for the organs present in any sampled subvolume, allowing optimal alignment of our text and volume encoders outputs.
    \item \textbf{Large-scale volumetric vision-language pretraining.} We advance volumetric foundation modeling by pretraining SigVLP on a large publicly available corpus of 3D CT volumes (CT-RATE), demonstrating for the first time that anatomically consistent vision-language alignment can emerge at scale. SigVLP unifies spatial continuity and semantic abstraction through adopted ideas from Large Language models, such as RoPe~\cite{wei2025videorope,wei2025videoropepp} and Muon~\cite{jordan2024muon}. 
    \item \textbf{Organ-wise clinical observations dataset.} We release an open-source dataset of organ-wise clinical observations, automatically extracted from free-text radiology reports using a custom LLM-assisted text analysis approach, on top of the publicly available CT-RATE dataset.
    \item \textbf{Downstream performance improvements.} We explore the expressivity of our learned representations by evaluation our embeddings on a wide range of downstream tasks, demonstrating the benefits of our approach.
\end{itemize}

\section{Related Works}
\label{sec:related}

Recent advances in medical vision-language pretraining (Med-VLP) have mainly targeted 2D chest X-rays (CXR), using contrastive learning to align images with reports~\cite{zhang2022convirt,yan2022clinicalbert,boecking2022textsemantics,huang2024knowledge}. While effective, such global alignment overlooks fine-grained associations between image regions and report segments~\cite{chen2023adamatch}. To enrich textual representation, MedKLIP~\cite{wu2023medklip} and KAD~\cite{zhang2023kad} embed domain knowledge, and Imitate~\cite{liu2024imitate} applies hierarchical alignment between visual and structured text features. MedCLIP~\cite{wang2022medclip} and PTUnifier~\cite{chen2023ptunifier} train on unpaired corpora, while PairAug~\cite{xie2024pairaug} expands supervision through synthetic pairing.

Med-VLP has since extended to 3D imaging, particularly CT~\cite{hamamci2024developing,zhang2025velvetmed,shui2025organclip,blankemeier2024merlin}. VELVET-Med~\cite{zhang2025velvetmed} and CT-CLIP~\cite{hamamci2024developing} align CT volumes with radiology reports, and Merlin~\cite{blankemeier2024merlin} integrates structured electronic health records data for abdomen-level modeling. To overcome the spatial rigidity of fixed-grid inputs, T3D~\cite{liu2023t3d} introduces a multi-view 3D CLIP but lacks volumetric continuity. DCFormer~\cite{ates2025dcformer} and Med3DVLM~\cite{xin2025med3dvlm} decompose volumes hierarchically yet still resample scans to uniform grids; the latter also employs a pairwise sigmoid contrastive loss loss~\cite{zhai2023sigmoid}, validating its effectiveness. MS-VLM~\cite{lee2024read} aggregates 2D slice features via a Z-former to support variable-length scans but reconstructs 3D context only post-hoc. Our framework instead preserves volumetric coherence through dynamic chunk-wise processing rather than slice aggregation or fixed decomposition.

Some works pursue fine-grained, organ-level alignment. CT-GLIP~\cite{lin2024ct} builds organ-level image-text pairs to localize structures and align each with text reports. fVLM~\cite{shui2025organclip} similarly performs anatomy-level contrastive learning for detailed CT understanding. These methods show the utility of localized supervision but collapse scans into pooled organ features, losing z-axis continuity. Our approach retains volumetric context by aligning variable-length subvolumes with relevant report segments, enabling dynamic RoPE-based modeling without resampling.

Recent grounding-oriented works explore partial supervision and multi-modal weak labels: MedFinder~\cite{chen2024bimcv} aligns CT sub-regions with report phrases for retrieval, and PET/CT grounding~\cite{huemann2025vision} fuses modalities at fine granularity. These demonstrate growing interest in spatially grounded medical VLPs, though they rely on curated landmarks or multi-stage training.
\section{Method}
\label{sec:method} 

\noindent\textbf{Preprocessing.}
We convert each full free-text report from the CT-RATE dataset to an organ-indexed dictionary using a lightweight LLM. After evaluating GPT~5~\cite{openai2025chatgpt5}, GPT~5~Mini~\cite{openai2025chatgpt5}, GPT~5~Nano~\cite{openai2025chatgpt5}, DeepSeek~R1~\cite{guo2025deepseekr1} and Qwen~1.5B~\cite{bai2023qwen}, we selected GPT~5~Mini because it yielded the best cost-quality trade-off. For each organ $o\in\mathcal{O}$ the record $\mathrm{Findings}(o)=(s_o,f_o)$ with $s_o\in\{\texttt{normal},\texttt{abnormal},\texttt{not\_examined}\}$ and free-text $f_o\in\Sigma_{alnum}$. An optional global note is $g=\mathrm{Findings}(\texttt{general})$. We enforce $s_o=\texttt{not\_examined}\Rightarrow f_o=\text{``not\_examined''}$. During training, given a 4D organ mask $\mathbf{M}\in\{0,1\}^{T\times H\times W\times|\mathcal{O}|}$, we sample chunks with length $\ell\sim\mathcal{U}(\{32,64,128\})$ and starting z-axis slice  $s\sim\mathcal{U}(\{0,\dots,T-\ell\})$. We define $\mathcal{C}=\{s,\dots,s+\ell-1\}$, and retrieve organs present in the chunks such that
\[
\mathcal{O}_{\mathcal{C}}=\{\,o\in\mathcal{O}\mid \max_{t\in\mathcal{C},x,y}\mathbf{M}(t,x,y,o)=1\,\}.
\]

\noindent\textbf{Train-time Chunk Composition.}
If $\mathcal{O}_{\mathcal{C}}=\varnothing$, we return ``No target structures were detected in this CT block.'' Otherwise, we partition the organs by status
\[
\begin{aligned}
\mathcal{A}&=\{\,o\in\mathcal{O}_{\mathcal{C}}:\ s_o=\texttt{abnormal},\ f_o\neq\emptyset\,\},\\
\mathcal{N}&=\{\,o\in\mathcal{O}_{\mathcal{C}}:\ s_o=\texttt{normal},\ f_o\neq\emptyset\,\},\\
\mathcal{X}&=\{\,o\in\mathcal{O}_{\mathcal{C}}:\ s_o=\texttt{not\_examined}\,\}.
\end{aligned}
\]

Let the $\mathrm{Join}(\cdot)$ operation generate a coma-separated list and $\oplus$ denote the whitespace concatenation. We compose
\[
\begin{aligned}
S_{\mathcal{X}}&=\big(\mathrm{List}(\mathcal{X})+\text{``were not examined.''}\big)\,\mathbf{1}[|\mathcal{X}|>0],\\
S_{\mathcal{N}}&=\mathrm{Join}\big(\{f_o\}_{o\in\mathcal{N}}\big),\\
S_{\mathcal{A}}&=\mathrm{Join}\big(\{f_o\}_{o\in\mathcal{A}}\big),\\
S_{g}&=g\cdot\mathbf{1}\!\big[g\not\in\{\emptyset,\text{``not\_examined''}\}\big].
\end{aligned}
\]

The final description
$D(\mathcal{C})=S_{\mathcal{X}}\oplus S_{\mathcal{N}}\oplus S_{\mathcal{A}}\oplus S_g$
yields organ-aware, chunk-aligned supervision at multiple ``temporal'' scales $\ell\in\{32,64,128\}$. 
This chunk-wise approach is effective for volumetric vision-language learning: rather than aligning a single global report to an entire scan, we establish fine-grained, spatially localized correspondences between subvolumes and their relevant textual findings. Overlapping chunks ensure anatomical continuity across boundaries and mitigate label leakage, while fixed-length subvolumes normalize variable scan depths and enable dynamic batching. Thus, each 3D CT scan is composed of dozens of well-grounded supervision pairs,  increasing the density and quality of learning signals. It allows us to capture both global and local anatomical semantics, learn position-aware embeddings without resampling, and achieve improved generalization across heterogeneous imaging protocols.

\noindent\textbf{Rotary Position Embedding.}
To support variable-length volumes, we remove additive absolute positional embeddings and apply RoPE directly inside the attention operation. Let the hidden states be
$\mathbf{H}\!\in\!\mathbb{R}^{B\times L\times C}$ with $B$ batch size, $L$ sequence length, and $C$ channels. We compute projections
\begin{align}
\mathbf{Q}=\mathbf{H}\mathbf{W}^Q,\quad
\mathbf{K}=\mathbf{H}\mathbf{W}^K,\quad
\mathbf{V}=\mathbf{H}\mathbf{W}^V,
\end{align}
and reshape to $B\times H\times L\times d$ with $H$ heads and $d=C/H$.

For position $t\!\in\!\{0,\ldots,L\!-\!1\}$ and paired index
$r\!\in\!\{0,\ldots,\frac{d}{2}\!-\!1\}$, we define a geometric frequency with base $b\!>\!1$:
\begin{equation}
\omega_r = b^{-\frac{2r}{d}},\qquad \theta_{t,r} = t\,\omega_r,
\end{equation}
and we use RoPE base $b=1000$.
We write a head's query $\mathbf{q}_t\!\in\!\mathbb{R}^d$ as a pair
$\big(q^{(0)}_{t,r}, q^{(1)}_{t,r}\big)$ (even/odd dimensions). RoPE rotates each pair:
\begin{align}
\widetilde{q}^{(0)}_{t,r} &= q^{(0)}_{t,r}\cos\theta_{t,r} - q^{(1)}_{t,r}\sin\theta_{t,r},\\
\widetilde{q}^{(1)}_{t,r} &= q^{(0)}_{t,r}\sin\theta_{t,r} + q^{(1)}_{t,r}\cos\theta_{t,r},
\end{align}
and analogously for keys to obtain $\widetilde{\mathbf{k}}_t$. Values are left unchanged.
Stacking all pairs yields $\widetilde{\mathbf{Q}}, \widetilde{\mathbf{K}}\in\mathbb{R}^{B\times H\times L\times d}$.

\noindent\textbf{Attention.}
We compute the scaled dot-product attention with an 
additive mask $\mathbf{M}$):
\begin{align}
\mathbf{A} &= \operatorname{softmax}\!\left(\frac{\widetilde{\mathbf{Q}}\;\widetilde{\mathbf{K}}^{\!\top}}{\sqrt{d}} + \mathbf{M}\right),\\
\mathbf{O} &= \mathbf{A}\,\mathbf{V};  
\mathbf{O}_{\text{proj}} = \operatorname{Concat}_{h=1}^{H}(\mathbf{O}_h)\,\mathbf{W}_O,
\end{align}
where $\mathbf{O}_h\in\mathbb{R}^{B\times L\times d}$ is the output of head $h$, 
$H$ is the total number of heads, and $\mathbf{W}_O\in\mathbb{R}^{(H d)\times C}$ 
is the standard output projection matrix that linearly combines all heads 
and maps the concatenated attention output back to the model dimension $C$.
In this implementation, RoPE is applied to $\mathbf{Q}/\mathbf{K}$ only, computed on-the-fly from the current $L$, thus enabling dynamic sequence lengths with no additional  parameters and negligible overhead.

\section{Experiments}
\label{sec:results} 

\begin{figure*}[htbp]
    \centering
    \begin{subfigure}{0.24\textwidth}
        \includegraphics[width=\linewidth]{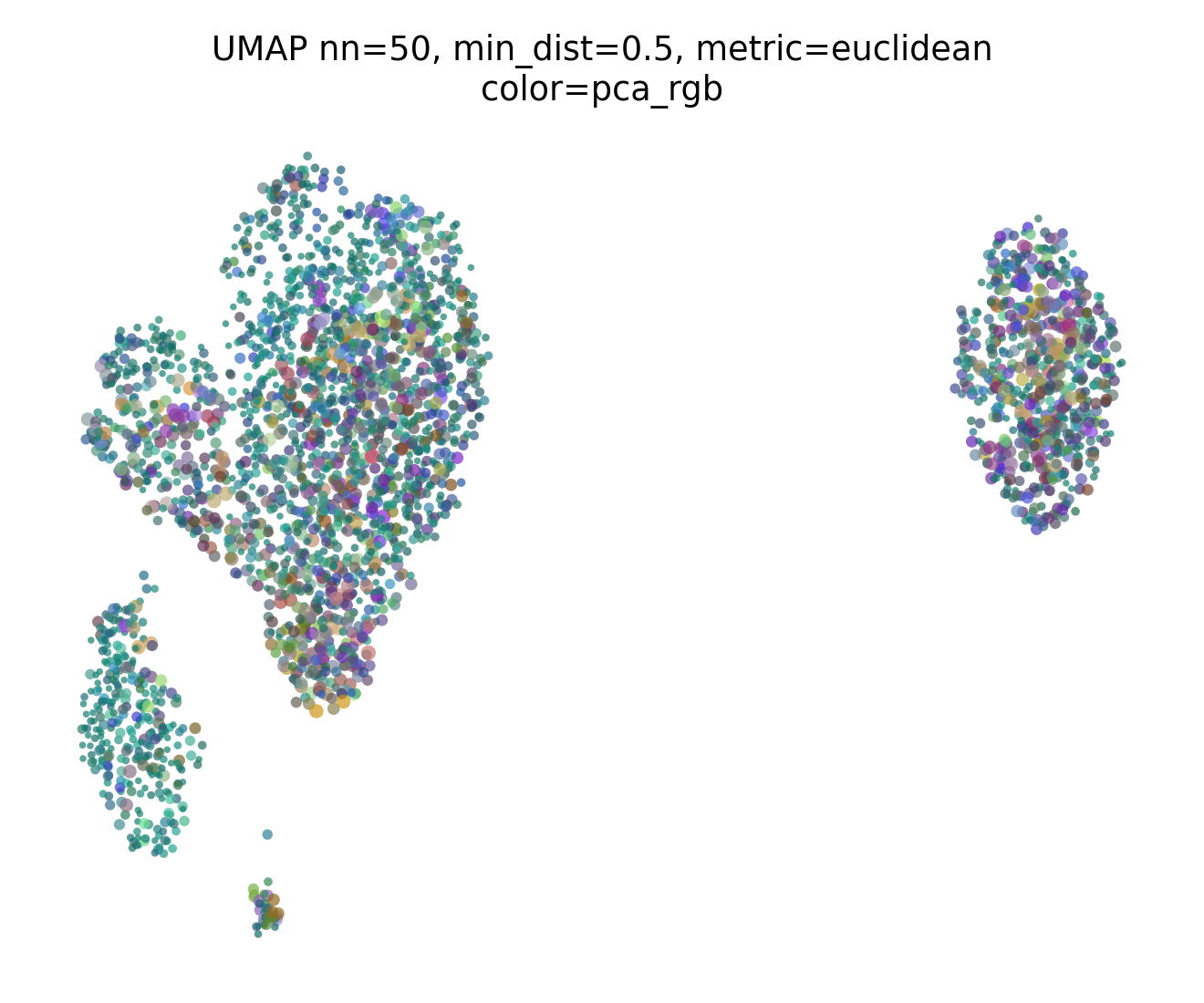}
        \caption{DINOv3-base}
    \end{subfigure}
    \begin{subfigure}{0.24\textwidth}
        \includegraphics[width=\linewidth]{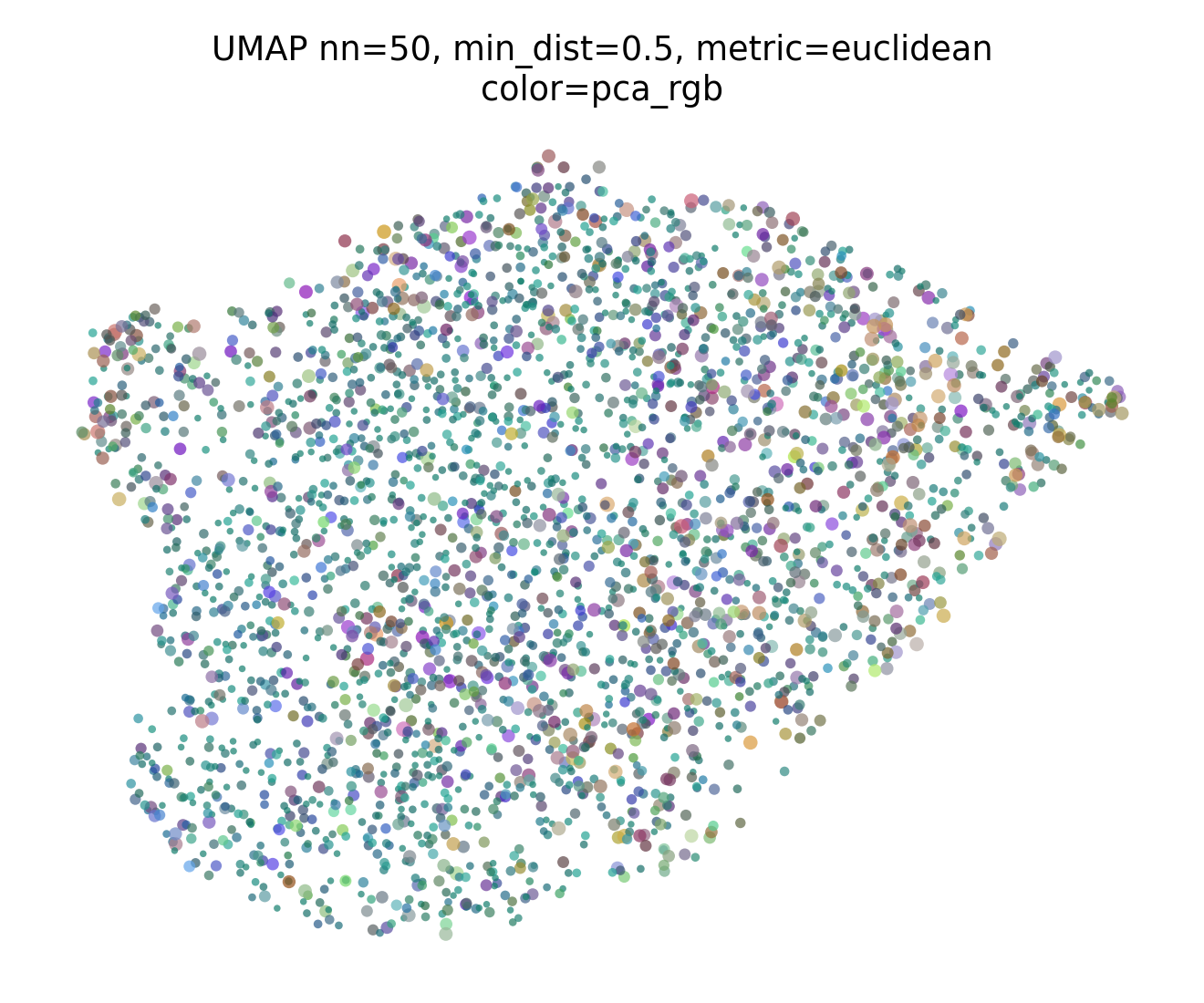}
        \caption{CT-CLIP}
    \end{subfigure}
    \begin{subfigure}{0.24\textwidth}
        \includegraphics[width=\linewidth]{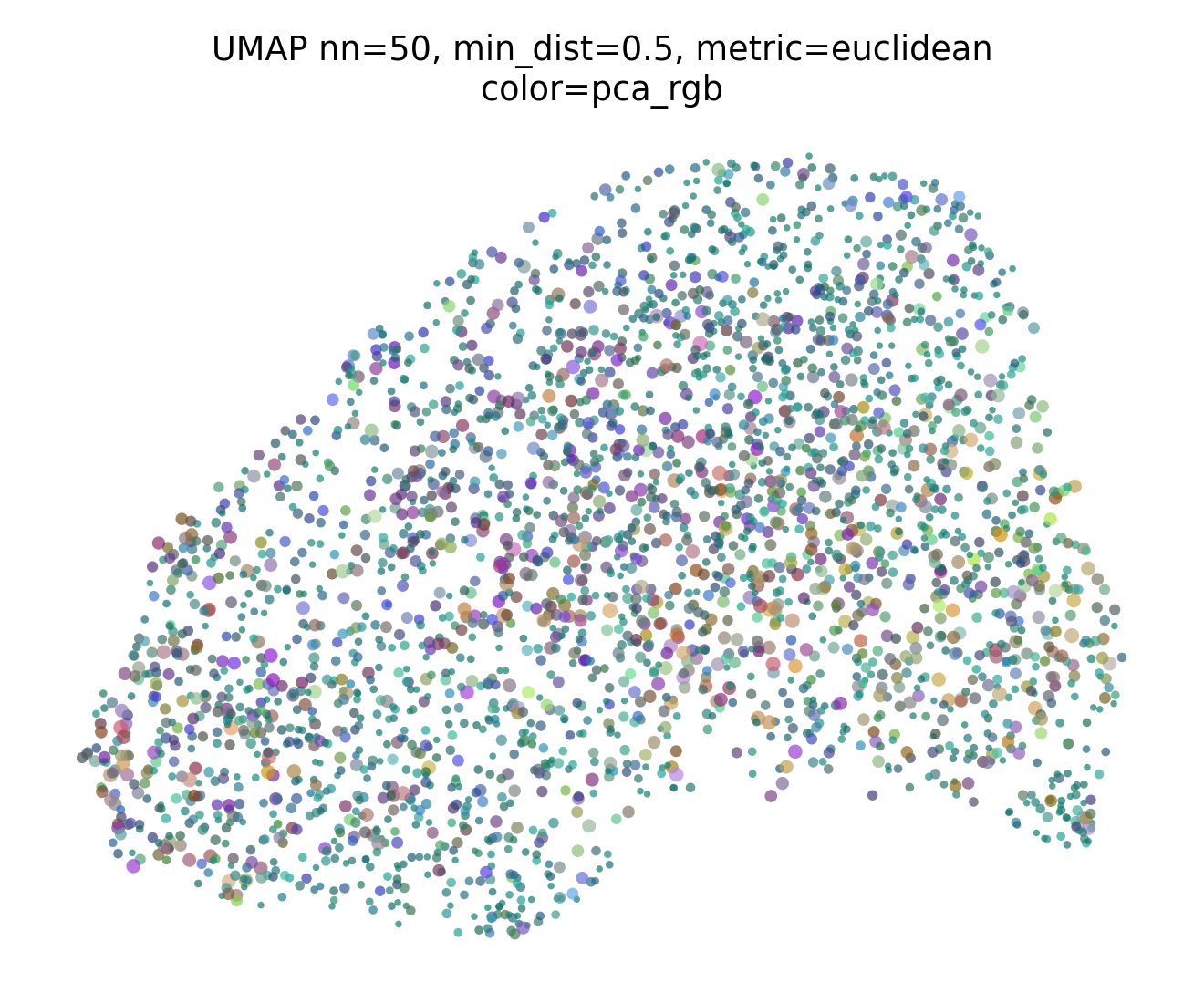}
        \caption{CT-Vocab (vocabulary-finetuned)}
    \end{subfigure}
    \begin{subfigure}{0.24\textwidth}
        \includegraphics[width=\linewidth]{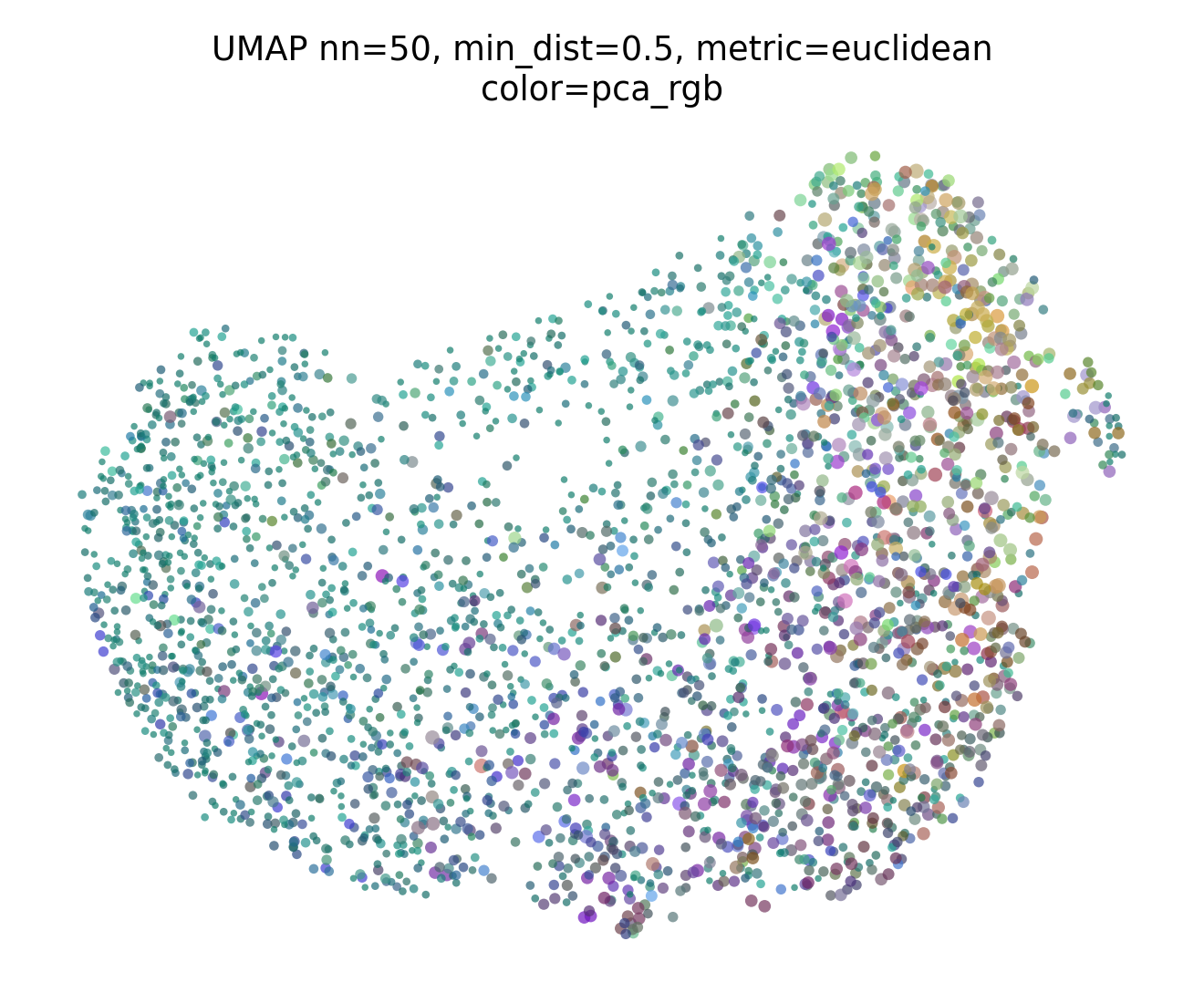}
        \caption{CT-LiPro (classification-tuned)}
    \end{subfigure}

    \begin{subfigure}{0.24\textwidth}
        \includegraphics[width=\linewidth]{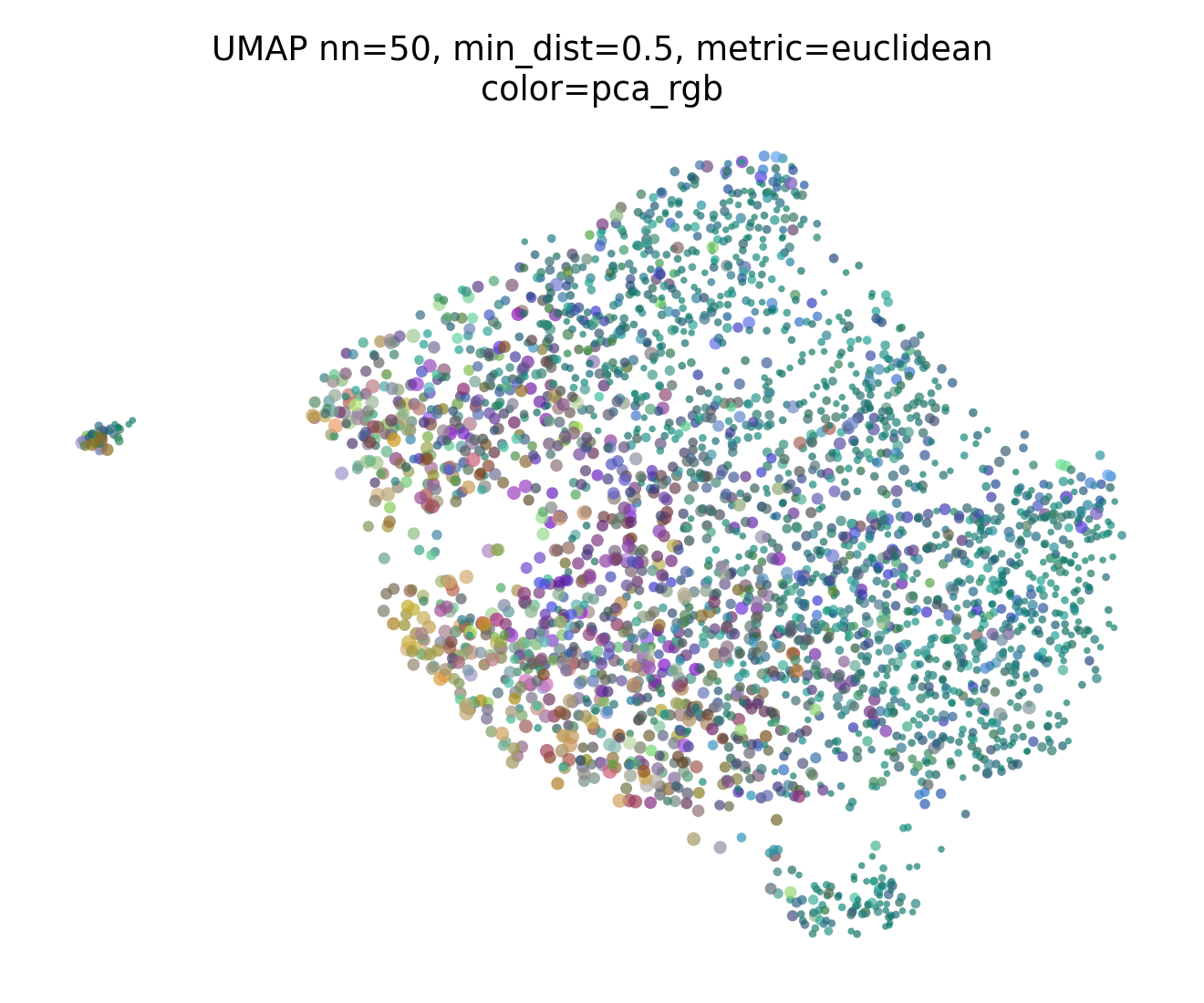}
        \caption{Ours (2k steps)}
    \end{subfigure}
    \begin{subfigure}{0.24\textwidth}
        \includegraphics[width=\linewidth]{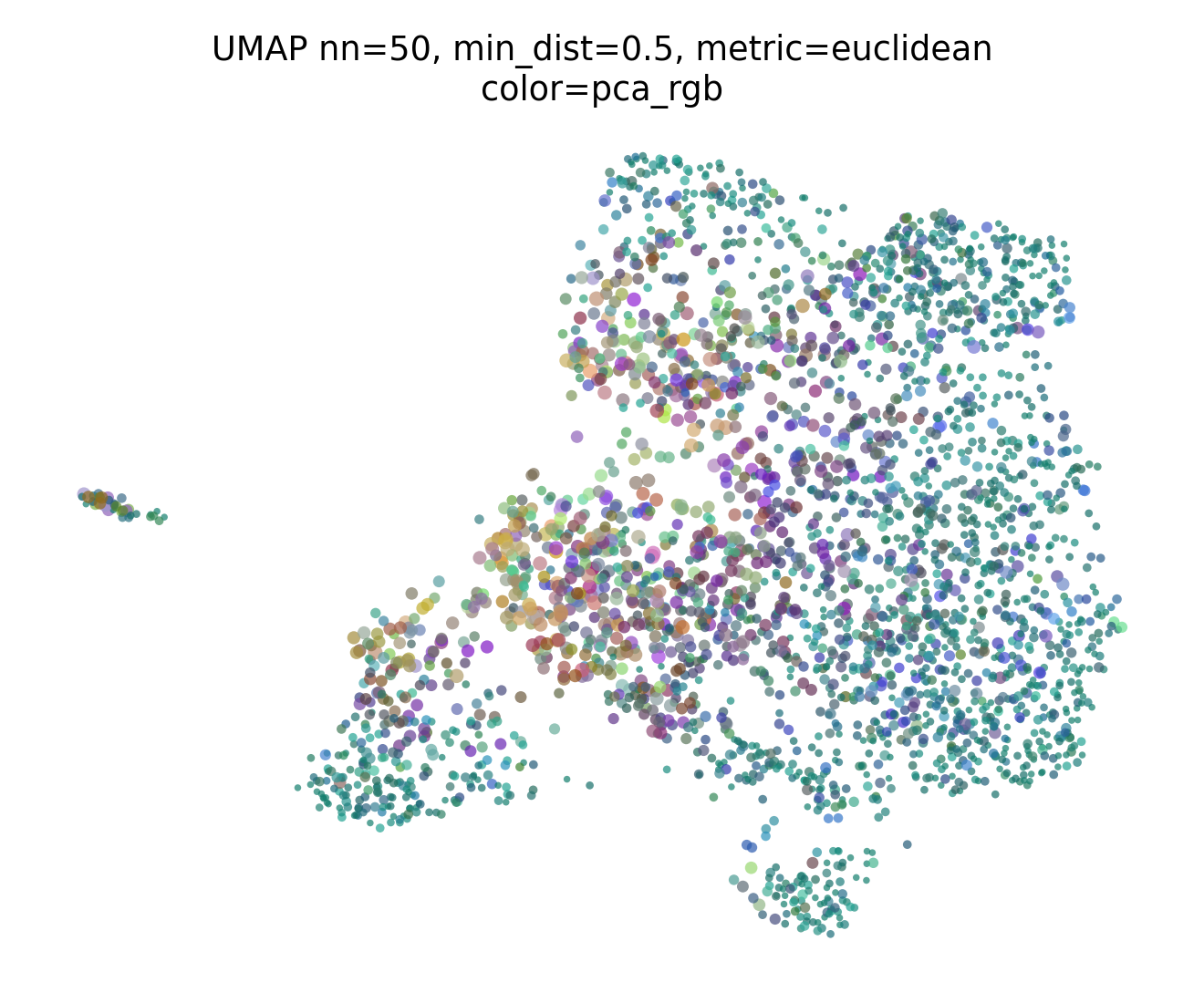}
        \caption{Ours (4k steps)}
    \end{subfigure}
    \begin{subfigure}{0.24\textwidth}
        \includegraphics[width=\linewidth]{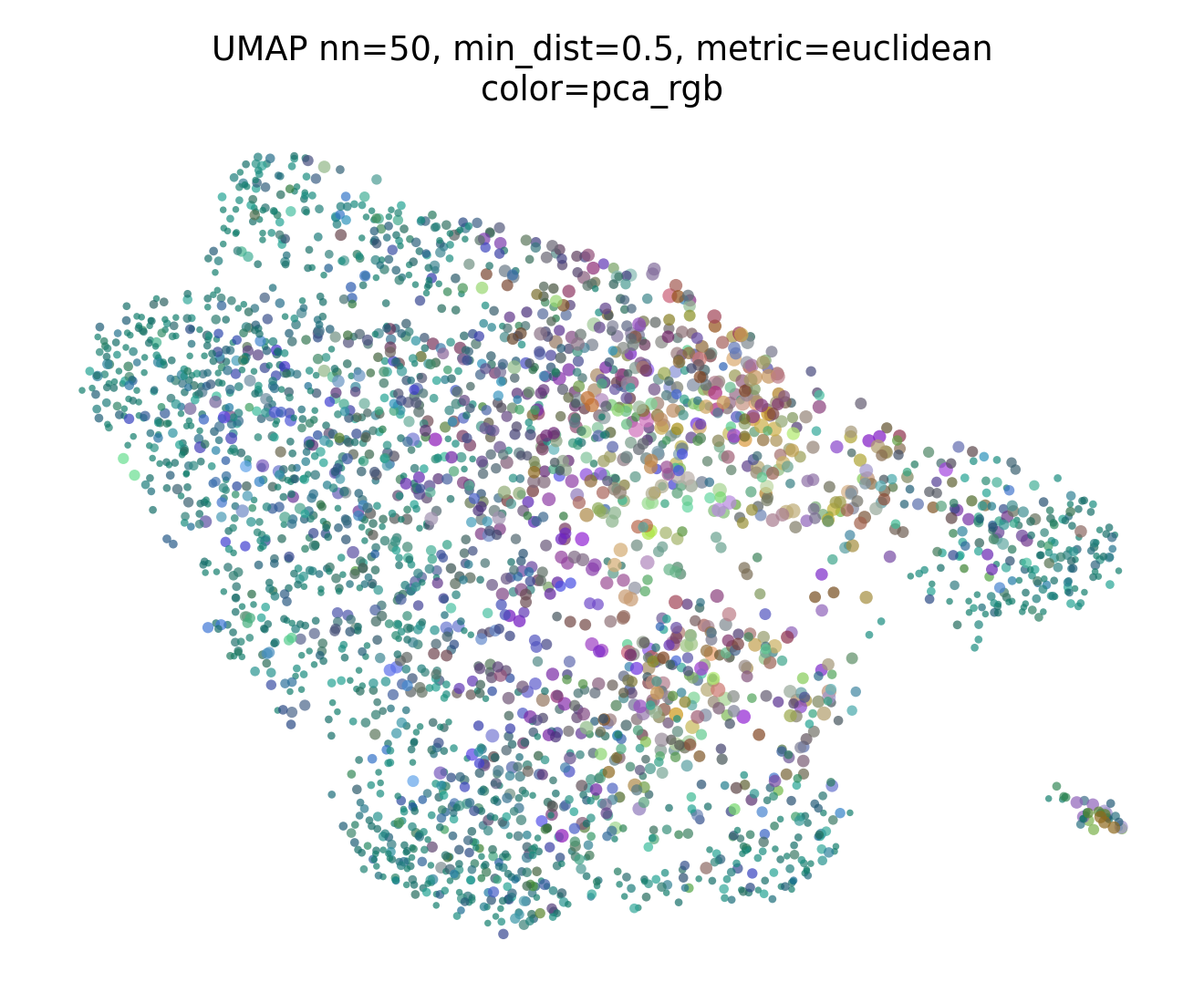}
        \caption{Ours (6k steps)}
    \end{subfigure}
    \begin{subfigure}{0.24\textwidth}
        \includegraphics[width=\linewidth]{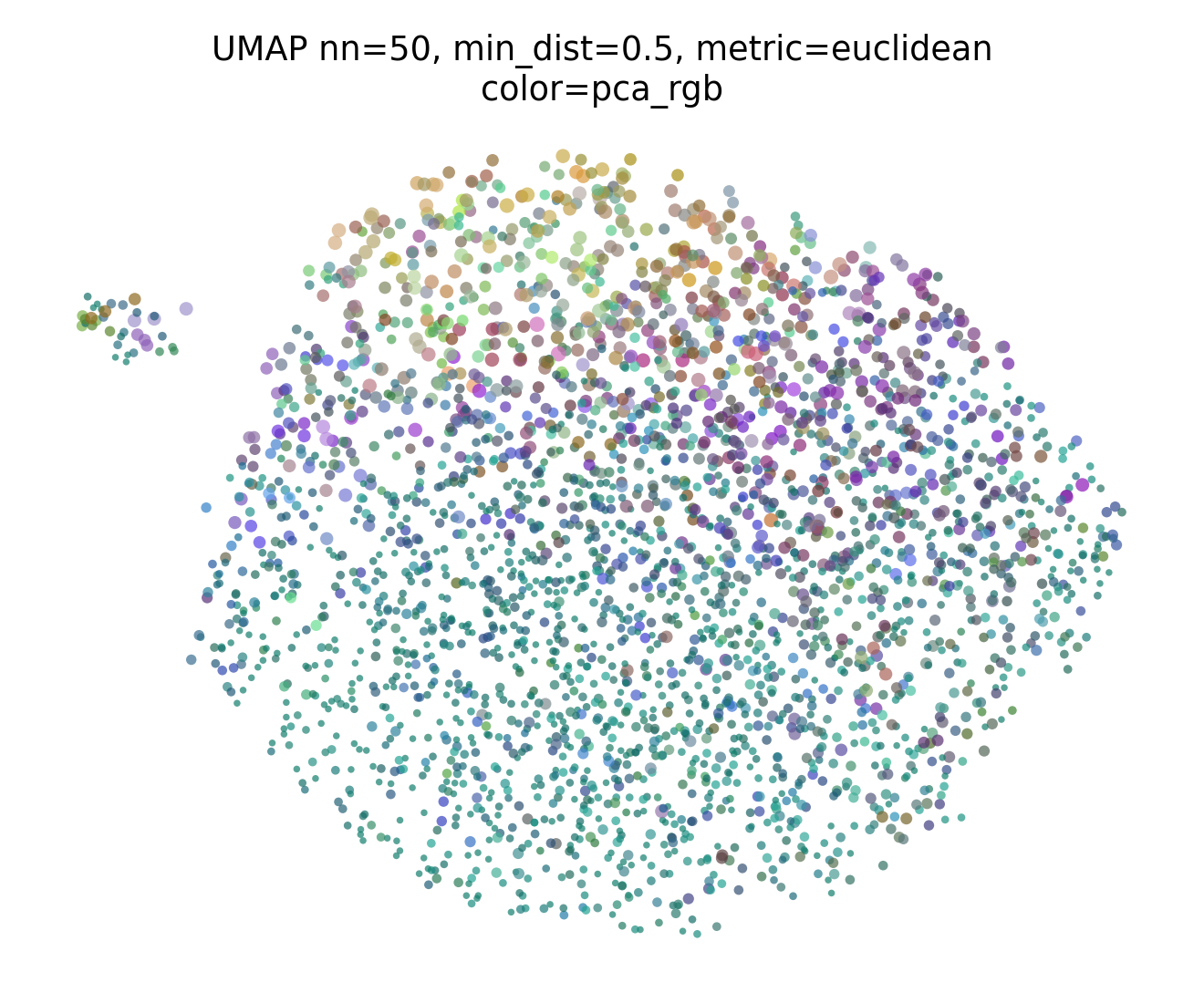}
        \caption{Ours (230k steps)}
    \end{subfigure}

    \caption{\textbf{UMAP~\cite{mcinnes2018umap} visualization of evaluation embeddings} across baselines and our model. 
    Colors indicate abnormality classes, with similar hues corresponding to semantically related labels. 
    \textit{Top row:} DINOv3-base, CT-CLIP, CT-Vocab (vocabulary-finetuned), CT-LiPro (classification-finetuned). 
    \textit{Bottom row:} Our method at 2k, 4k, 6k, and 234{,}930 training steps.}
    \label{fig:big_4x2}
\end{figure*}

\label{sec:impl}
We use the \textbf{CT\mbox{-}RATE}~\cite{hamamci2024foundation} dataset for all our experiments, using \(\,40\mathrm{k}+\) CT volumes for training and a fully isolated \(\,3\mathrm{k}+\) set for validation (no patient overlap). Training was done on \(16\) nodes, with \(4\) \(\times\) H100 GPUs per node and a per\hyp{}GPU batch size of \(8\), giving a global batch of
$B_{\text{global}} \;=\; 8 \times 4 \times 16 \;=\; 512.$
We use a learning rate of \(1\!\times\!10^{-3}\) and weight decay \(1\!\times\!10^{-4}\).
To reduce inter\hyp{}scanner variation, all slices are resized in-plane to \(256\times256\).
Text supervision uses the dataset\hyp{}provided, volume\hyp{}associated radiology reports; anatomy grounding uses high\hyp{}confidence masks from~\cite{wasserthal2023totalsegmentator}.

Our downstream tasks include
zero-shot abnormality classification using whole-chunk representations to assess CLIP-style text-vision alignment, linear-probe classification and segmentation on frozen embeddings to measure anatomical precision and the separability of organ-specific features, as well as chunk-wise organ presence classification. We also evaluate text-image retrieval, which directly assess localized visual-textual correspondence and the effectiveness
of our on-the-fly chunk-text alignment pretraining strategy. Our primary emphasis is on the vision component: we test whether a chunk-wise training strategy enables fine-grained, detail-level alignment. We compare CT-specific variants, CT-CLIP (baseline), CT-Vocab (vocabulary fine-tuned), and CT-LiPro (classification fine-tuned), and include DINOv3-base \cite{simeoni2025dinov3, duriantaco2025dinov3clip} as a standalone, general-purpose visual backbone to gauge out-of-domain generalization.
Fig.~\ref{fig:big_4x2} presents eight UMAP~\cite{mcinnes2018umap} embedding projection plots: CT-CLIP~\cite{hamamci2024developing}, CT-Vocab~\cite{hamamci2024developing}, CT-LiPro~\cite{hamamci2024foundation}, DINOv3-base~\cite{simeoni2025dinov3}, and our model at four different training stages. In each plot, similar hues correspond to semantically similar labels. We observe that DINOv3 produces separated clusters while most CT-specific baselines exhibit few isolated clusters. 
During training, after around 500 optimization steps, both training loss and validation metrics stabilize, yet representation structure continues to evolve as the training progresses. At $2{,}000$ steps (Fig.~\ref{fig:big_4x2}(e)), the projection already shows clear color separation, resembling the fine-tuned CT-LiPro.
At $234{,}930$ steps, our embeddings are projected as a continuous and uniform hue, indicating that the model has learned a smooth, structured representation of the underlying feature space.
We interrupted training at $234{,}930$ ($\approx 120$M examples seen) as scaling laws tells us that at that stage, only order of magnitudes more training would impact the quality of the model.

\begin{table}[ht]
\centering
\footnotesize
\setlength{\tabcolsep}{5pt}
\caption{Retrieval performance between 200-slice 3D CT volumes and corresponding radiology reports.
The baseline uses the initial \textbf{SigLIP2} weights released in~\cite{tschannen2025siglip}.
For our proposed \textbf{SigVLP} method, we employ reconstructed reports generated through our pipeline.
At each evaluation step, 100 samples are retrieved from a total of 1,500 candidates, and all recall values are computed based on the top-100 retrieval results.
}
\resizebox{\columnwidth}{!}{
\begin{tabular}{lcccc}
\hline
Model & R@5$\uparrow$ & R@10$\uparrow$ & R@50$\uparrow$ & MeanRank$\downarrow$ \\
\hline
SigLIPv2-L \cite{tschannen2025siglip}      & 0.047{\tiny$\pm$0.007} & 0.101{\tiny$\pm$0.009} & 0.504{\tiny$\pm$0.011} & 50.53{\tiny$\pm$0.39} \\
CT-Clip \cite{hamamci2024developing}       & 0.204{\tiny$\pm$0.038} & 0.348{\tiny$\pm$0.052} & 0.835{\tiny$\pm$0.049} & 26.01{\tiny$\pm$3.01} \\
CT-Vocab \cite{hamamci2024developing}      & 0.055{\tiny$\pm$0.012} & 0.107{\tiny$\pm$0.015} & 0.517{\tiny$\pm$0.013} & 49.12{\tiny$\pm$0.80} \\
Dinov3-Clip \cite{duriantaco2025dinov3clip}   & 0.047{\tiny$\pm$0.020} & 0.107{\tiny$\pm$0.027} & 0.501{\tiny$\pm$0.024} & 50.53{\tiny$\pm$1.06} \\
Ours & \textbf{0.636}{\tiny$\pm$0.050} & \textbf{0.769}{\tiny$\pm$0.036} & \textbf{0.978}{\tiny$\pm$0.010} & \textbf{8.23}{\tiny$\pm$0.947} \\
\hline
\end{tabular}
}
\label{tab:clip_comparison}
\end{table}


\noindent\textbf{{Retrieval Performace between 3D Volume and Radiology Report.}}
In Table~\ref{tab:clip_comparison}, we present the retrieval performance of different CLIP variants, including CT-specific CLIPs and general 2D image CLIPs such as DINOv3-CLIP and our baseline SigLIP2. For the 2D models, we averaged slice-level features,  and used them as input, but these models performed poorly, indicating their limited ability to capture meaningful medical information. However, 3D models like CT-CLIP achieved higher results, indicating that it could retrieve valid matches.
Our SigVLP demonstrates a significant and statistically valid improvement over all previous methods, with a MeanRank of $8.23\pm0.947$ compared to the CT-Clip's $26.01\pm3.01$.

\begin{table*}[t]
\centering
\footnotesize
\caption{ Classification: linear classification on 18 abnormalities. Segmentation: linear MLP segmentation, where  Segmentation (Dice) shows the results for three representative organs, and Segmentation (Overall) shows mDice and mIoU over 12 organs. For both classification and segmentation, performance is summarized using per-organ mean values and standard deviations. Classification results report standard deviations obtained via bootstrap sampling. Segmentation results report standard deviations across 12 organs. Zero values indicate that no valid mask was produced (all background label), not missing entries.}
\resizebox{\textwidth}{!}{
\begin{tabular}{lccccc|ccc|cc}
\toprule
& \multicolumn{5}{c|}{Classification} & \multicolumn{3}{c|}{Segmentation (Dice)} & \multicolumn{2}{c}{Segmentation (Overall)} \\
\cmidrule(lr){2-6} \cmidrule(lr){7-9} \cmidrule(lr){10-11}
Backbone 
& Precision & Recall & F1 & Accuracy & AUROC 
& Task01 & Task02 & Task03 & mDice & mIoU \\
&  &  &  &  &  
& (lung\_left) & (Aorta) & (Stomach) &  &  \\
\midrule
SigLIPv2-L~\cite{tschannen2025siglip} & 0.362{\tiny $\pm$0.041} & 0.518{\tiny $\pm$0.038} & 0.388{\tiny $\pm$0.028} & 0.756{\tiny $\pm$0.013} & 0.706{\tiny $\pm$0.028} & 0.000 & 0.000 & 0.000 & 0.000{\tiny $\pm$0.000} & 0.000{\tiny $\pm$0.000} \\
CT-CLIP~\cite{hamamci2024developing} & 0.288{\tiny $\pm$0.065} & 0.424{\tiny $\pm$0.044} & 0.305{\tiny $\pm$0.048} & 0.764{\tiny $\pm$0.016} & 0.639{\tiny $\pm$0.031} & 0.105 & 0.000 & 0.000 & 0.013{\tiny $\pm$0.029} & 0.007{\tiny $\pm$0.015} \\
CT-Vocab~\cite{hamamci2024developing} & 0.300{\tiny $\pm$0.060} & 0.463{\tiny $\pm$0.031} & 0.333{\tiny $\pm$0.036} & 0.765{\tiny $\pm$0.019} & 0.653{\tiny $\pm$0.042} & 0.005 & 0.000 & 0.000 & 0.000{\tiny $\pm$0.000} & 0.000{\tiny $\pm$0.000} \\
DINOv3-S~\cite{simeoni2025dinov3}   & 0.330{\tiny $\pm$0.025} & \textbf{0.815}{\tiny $\pm$0.032} & \textbf{0.455}{\tiny $\pm$0.029} & 0.516{\tiny $\pm$0.011} & 0.651{\tiny $\pm$0.023} & 0.816 & 0.002 & 0.237 & 0.292{\tiny $\pm$0.257} & 0.226{\tiny $\pm$0.201} \\
DINOv3-B~\cite{simeoni2025dinov3}   & 0.332{\tiny $\pm$0.032} & 0.712{\tiny $\pm$0.052} & 0.437{\tiny $\pm$0.036} & 0.553{\tiny $\pm$0.019} & 0.649{\tiny $\pm$0.031} & \textbf{0.876} & 0.278 & 0.340 & 0.448{\tiny $\pm$0.289} & 0.362{\tiny $\pm$0.251} \\
DINOv3-L~\cite{simeoni2025dinov3}  & 0.342{\tiny $\pm$0.022} & 0.714{\tiny $\pm$0.032} & 0.446{\tiny $\pm$0.022} & 0.569{\tiny $\pm$0.014} & 0.654{\tiny $\pm$0.033} & 0.849 & 0.117 & 0.199 & 0.316{\tiny $\pm$0.305} & 0.250{\tiny $\pm$0.240} \\
\midrule
Ours (2000 step)  & 0.355{\tiny $\pm$0.020} & 0.632{\tiny $\pm$0.029} & 0.418{\tiny $\pm$0.016} & 0.702{\tiny $\pm$0.020} & \textbf{0.709}{\tiny $\pm$0.024} & 0.794 & 0.376 & 0.255 & 0.374{\tiny $\pm$0.261} & 0.287{\tiny $\pm$0.215} \\
Ours (4000 step)  & 0.408{\tiny $\pm$0.042} & 0.551{\tiny $\pm$0.032} & 0.429{\tiny $\pm$0.034} & 0.727{\tiny $\pm$0.008} & 0.708{\tiny $\pm$0.024} & 0.800 & 0.318 & 0.346 & 0.392{\tiny $\pm$0.300} & 0.304{\tiny $\pm$0.255} \\
Ours (230000 step) & \textbf{0.429}{\tiny $\pm$0.034} & 0.428{\tiny $\pm$0.028} & 0.410{\tiny $\pm$0.028} & \textbf{0.778}{\tiny $\pm$0.013} & 0.683{\tiny $\pm$0.016} & 0.786 & \textbf{0.471} & \textbf{0.489} & \textbf{0.487}{\tiny $\pm$0.253} & \textbf{0.381}{\tiny $\pm$0.203} \\
\bottomrule
\end{tabular}
}
\label{tab:whole_body}
\end{table*}

\begin{figure}[htb]
    \centering
    \setlength{\tabcolsep}{2pt}
    \renewcommand{\arraystretch}{1.0}
    \resizebox{\linewidth}{!}{
    \begin{tabular}{c c c c}
        & \textbf{Ground Truth} & \textbf{DINOv3-base} & \textbf{Ours} \\

        \raisebox{20pt}{\rotatebox[origin=c]{90}{\textbf{Slice 125}}} &
        \includegraphics[width=0.28\textwidth]{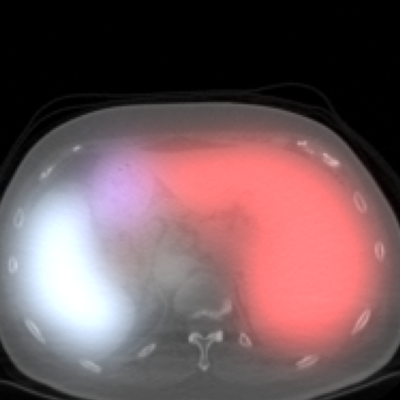} &
        \includegraphics[width=0.28\textwidth]{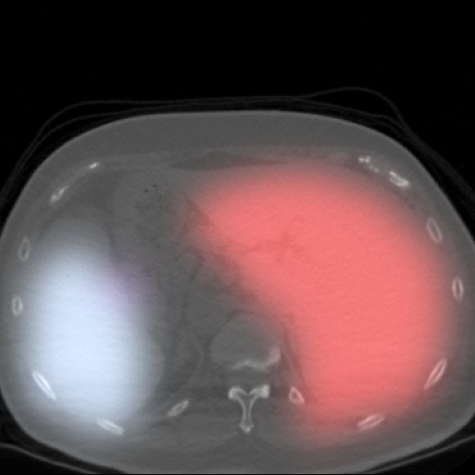} &
        \includegraphics[width=0.28\textwidth]{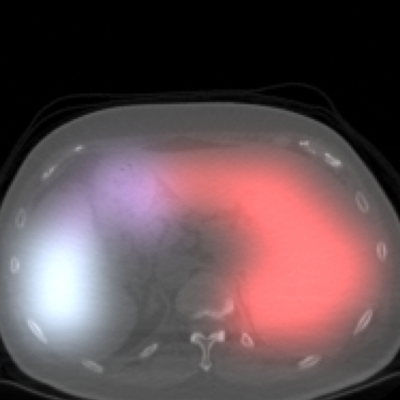} \\

        \raisebox{20pt}{\rotatebox[origin=c]{90}{\textbf{Slice 250}}} &
        \includegraphics[width=0.28\textwidth]{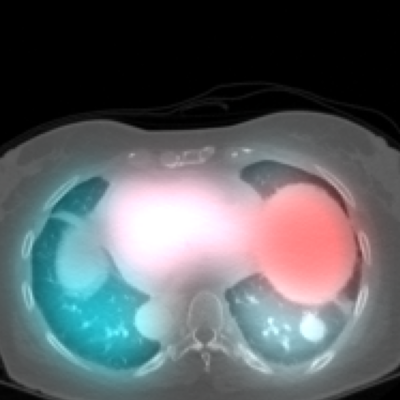} &
        \includegraphics[width=0.28\textwidth]{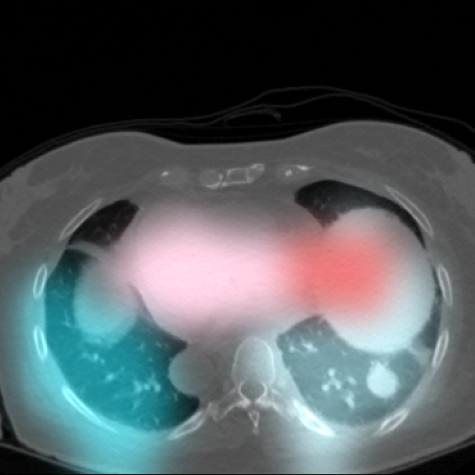} &
        \includegraphics[width=0.28\textwidth]{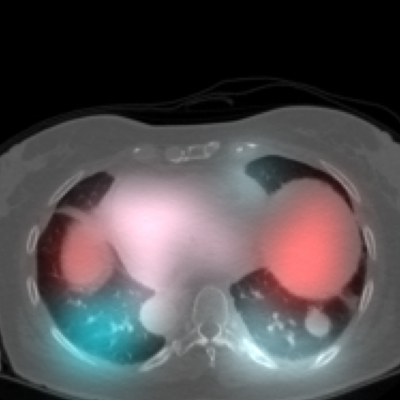} \\

        \raisebox{20pt}{\rotatebox[origin=c]{90}{\textbf{Slice 390}}} &
        \includegraphics[width=0.28\textwidth]{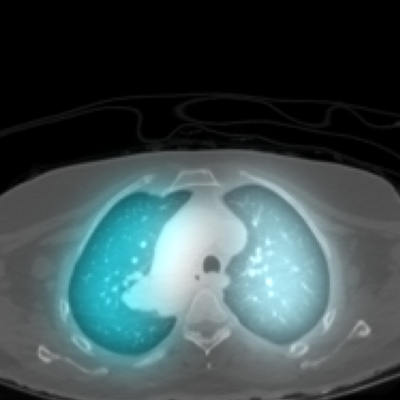} &
        \includegraphics[width=0.28\textwidth]{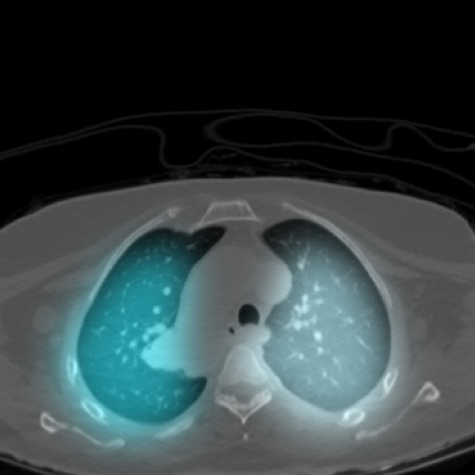} &
        \includegraphics[width=0.28\textwidth]{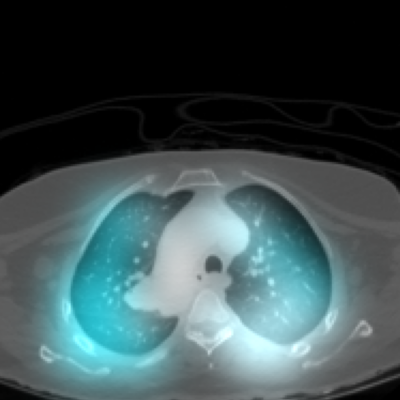} \\

        \raisebox{20pt}{\rotatebox[origin=c]{90}{\textbf{Slice 460}}} &
        \includegraphics[width=0.28\textwidth]{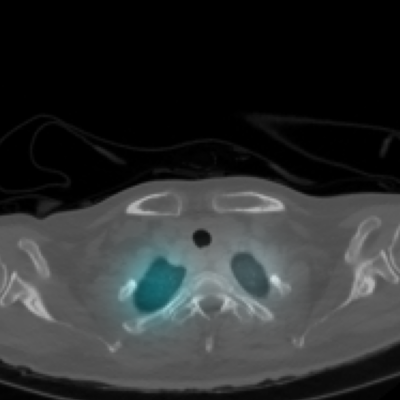} &
        \includegraphics[width=0.28\textwidth]{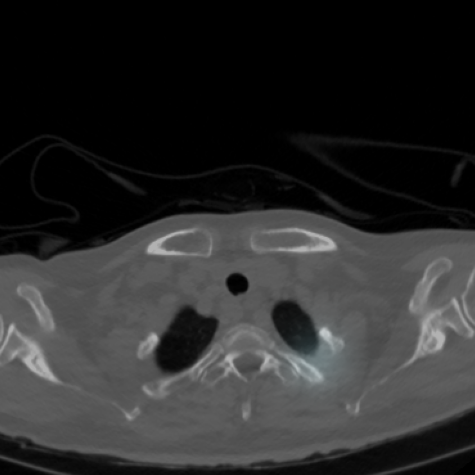} &
        \includegraphics[width=0.28\textwidth]{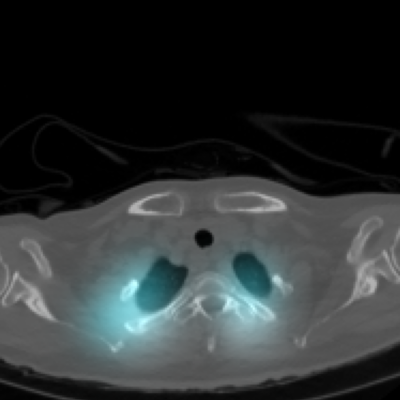} \\
    \end{tabular}
    }

    \caption{Qualitative comparison of segmentation overlays across four slices (125, 250, 390, 460) in an example volume. 
    Columns correspond to Ground Truth, DINOv3-base, and our method. Rows correspond to different slices, with vertically centered rotated labels for compact visualization.}
    \label{fig:segmentation_overlays}
\end{figure}

\noindent\textbf{Whole-Body Abnormality Classification and Organ-Wise Segmentation.}
Our downstream results reveal a clear scale-dependent trade-off in the learned features. In the linear-probe classification experiments in Table~\ref{tab:whole_body}, DINOv3 attains near-perfect recall of 0.996 but only 0.21 precision, which indicates features that are highly sensitive yet weakly selective. Our final model achieves the best precision of 0.435 and the highest accuracy of 0.80, which reflects a clean separation of true positives from hard negatives.

The linear-probe segmentation, performed in a downsampled patch-level space and reported in Table~\ref{tab:whole_body} (Segmentation (Dice) and (Overall)), further exposes the impact of object scale. DINOv3 performs  better on large organs where coarse cues suffice.
For medium and small structures such as the aorta or the stomach, our chunk-aligned volumetric features'   expressivity
surpass DINOv3's. On the aorta, the Dice score improves from 0.278 to 0.471, a $1.7\times$ improvement. Our model achieves the best mean Dice, supporting the conclusion that it captures finer anatomical details from volumetric data.

Qualitative comparisons in Fig.~\ref{fig:segmentation_overlays} echo these trend. DINOv3 covers a large fraction of pixels on large organs such as the lung and the spleen, which yields broad coverage but less precise boundaries. Our model produces tighter masks on smaller structures such as the stomach, and the predicted contours follow the ground truth more closely. 
DINOv3 is attractive for exhaustive screening where high recall is the priority, whereas our features are preferable for precise delineation and margin-sensitive decisions.
These results demonstrate, again, that the embeddings produced by our model retain much more information for small structures while retaining large-scale structure information. 


\begin{table*}[htbp]
\centering
\caption{Comparison of classification (linear probe) and zero-shot retrieval performance across different slice numbers.
The upper part shows 12-organ presence binary classification, where the reported standard deviations are organ-wise.
The lower part presents zero-shot retrieval performance, with standard deviations obtained via bootstrap sampling (subset size = 50 samples per bootstrap iteration).}
\resizebox{\textwidth}{!}{
\begin{tabular}{l|ccccc|ccccc|ccccc}
\toprule
\multicolumn{16}{c}{\textbf{Classification (Linear Probe)}} \\
\midrule
Method 
& \multicolumn{5}{c|}{32 slices} 
& \multicolumn{5}{c|}{64 slices} 
& \multicolumn{5}{c}{128 slices} \\ 
\cmidrule(lr){2-6} \cmidrule(lr){7-11} \cmidrule(lr){12-16}
& Precision & Recall & F1 & AUC & mAP 
& Precision & Recall & F1 & AUC & mAP 
& Precision & Recall & F1 & AUC & mAP \\
\midrule
SigLIPv2-L~\cite{tschannen2025siglip}  & 0.499{\tiny $\pm$0.399} & 0.494{\tiny $\pm$0.491} & 0.496{\tiny $\pm$0.367} & 0.523{\tiny $\pm$0.251} & 0.542{\tiny $\pm$0.345} & 0.599{\tiny $\pm$0.393} & 0.614{\tiny $\pm$0.397} & 0.607{\tiny $\pm$0.332} & 0.630{\tiny $\pm$0.215} & 0.592{\tiny $\pm$0.364} & 0.665{\tiny $\pm$0.349} & 0.767{\tiny $\pm$0.380} & 0.712{\tiny $\pm$0.381} & 0.652{\tiny $\pm$0.189} & 0.705{\tiny $\pm$0.324} \\
CT-CLIP~\cite{hamamci2024developing}    & 0.792{\tiny $\pm$0.396} & 0.668{\tiny $\pm$0.491} & 0.725{\tiny $\pm$0.440} & 0.840{\tiny $\pm$0.076} & 0.661{\tiny $\pm$0.204} & 0.832{\tiny $\pm$0.427} & 0.745{\tiny $\pm$0.500} & 0.786{\tiny $\pm$0.455} & 0.879{\tiny $\pm$0.100} & 0.696{\tiny $\pm$0.268} & 0.851{\tiny $\pm$0.421} & 0.871{\tiny $\pm$0.471} & 0.861{\tiny $\pm$0.437} & 0.895{\tiny $\pm$0.075} & 0.699{\tiny $\pm$0.288} \\
CT-Vocab~\cite{hamamci2024developing}   & 0.819{\tiny $\pm$0.360} & 0.661{\tiny $\pm$0.467} & 0.731{\tiny $\pm$0.420} & 0.850{\tiny $\pm$0.060} & 0.668{\tiny $\pm$0.187} & 0.937{\tiny $\pm$0.441} & 0.711{\tiny $\pm$0.495} & 0.809{\tiny $\pm$0.481} & 0.894{\tiny $\pm$0.106} & 0.682{\tiny $\pm$0.306} & 0.851{\tiny $\pm$0.421} & 0.871{\tiny $\pm$0.471} & 0.861{\tiny $\pm$0.437} & 0.877{\tiny $\pm$0.081} & 0.671{\tiny $\pm$0.299} \\
Dinov3-S~\cite{simeoni2025dinov3}       & \textbf{0.982}{\tiny $\pm$0.436} & 0.919{\tiny $\pm$0.458} & 0.950{\tiny $\pm$0.446} & 0.990{\tiny $\pm$0.120} & 0.517{\tiny $\pm$0.387} & 0.942{\tiny $\pm$0.437} & 0.893{\tiny $\pm$0.458} & 0.917{\tiny $\pm$0.446} & 0.979{\tiny $\pm$0.122} & 0.567{\tiny $\pm$0.388} & 0.915{\tiny $\pm$0.363} & 0.924{\tiny $\pm$0.404} & 0.920{\tiny $\pm$0.380} & 0.966{\tiny $\pm$0.123} & 0.726{\tiny $\pm$0.350} \\
Dinov3-B~\cite{simeoni2025dinov3}       & 0.958{\tiny $\pm$0.431} & 0.942{\tiny $\pm$0.450} & 0.950{\tiny $\pm$0.441} & 0.993{\tiny $\pm$0.097} & 0.629{\tiny $\pm$0.354} & 0.936{\tiny $\pm$0.349} & 0.920{\tiny $\pm$0.327} & 0.928{\tiny $\pm$0.320} & \textbf{0.987}{\tiny $\pm$0.090} & 0.650{\tiny $\pm$0.312} & 0.934{\tiny $\pm$0.325} & 0.931{\tiny $\pm$0.317} & 0.933{\tiny $\pm$0.320} & 0.978{\tiny $\pm$0.100} & 0.763{\tiny $\pm$0.364} \\
Dinov3-L~\cite{simeoni2025dinov3}       & 0.976{\tiny $\pm$0.422} & 0.925{\tiny $\pm$0.458} & 0.950{\tiny $\pm$0.446} & 0.992{\tiny $\pm$0.097} & 0.516{\tiny $\pm$0.380} & \textbf{0.975}{\tiny $\pm$0.430} & 0.837{\tiny $\pm$0.458} & 0.901{\tiny $\pm$0.447} & 0.983{\tiny $\pm$0.109} & 0.572{\tiny $\pm$0.376} & 0.925{\tiny $\pm$0.378} & 0.935{\tiny $\pm$0.439} & 0.930{\tiny $\pm$0.418} & 0.977{\tiny $\pm$0.095} & 0.761{\tiny $\pm$0.369} \\
\textbf{Ours} & 0.963{\tiny $\pm$0.272} & \textbf{0.960}{\tiny $\pm$0.300} & \textbf{0.961}{\tiny $\pm$0.308} & \textbf{0.995}{\tiny $\pm$0.004} & \textbf{0.977}{\tiny $\pm$0.030} & 0.940{\tiny $\pm$0.267} & \textbf{0.948}{\tiny $\pm$0.300} & \textbf{0.944}{\tiny $\pm$0.298} & 0.986{\tiny $\pm$0.011} & \textbf{0.896}{\tiny $\pm$0.029} & \textbf{0.949}{\tiny $\pm$0.284} & \textbf{0.960}{\tiny $\pm$0.302} & \textbf{0.954}{\tiny $\pm$0.311} & \textbf{0.986}{\tiny $\pm$0.018} & \textbf{0.903}{\tiny $\pm$0.027} \\
\midrule
\multicolumn{16}{c}{\textbf{Free-form Image to Report Retrieval}} \\
\midrule
Method
& \multicolumn{5}{c|}{32 slices} 
& \multicolumn{5}{c|}{64 slices} 
& \multicolumn{5}{c}{128 slices} \\ 
\cmidrule(lr){2-6} \cmidrule(lr){7-11} \cmidrule(lr){12-16}
& mAP & R@1 & R@5 & R@10 & NDCG@10 
& mAP & R@1 & R@5 & R@10 & NDCG@10 
& mAP & R@1 & R@5 & R@10 & NDCG@10 \\
\midrule
SigLIPv2-L~\cite{tschannen2025siglip} & 0.098{\tiny $\pm$0.015} & 0.024{\tiny $\pm$0.015} & 0.108{\tiny $\pm$0.024} & 0.234{\tiny $\pm$0.057} & 0.104{\tiny $\pm$0.024} & 0.101{\tiny $\pm$0.028} & 0.026{\tiny $\pm$0.027} & 0.120{\tiny $\pm$0.048} & 0.234{\tiny $\pm$0.059} & 0.108{\tiny $\pm$0.037} & 0.087{\tiny $\pm$0.016} & 0.014{\tiny $\pm$0.016} & 0.108{\tiny $\pm$0.035} & 0.210{\tiny $\pm$0.034} & 0.090{\tiny $\pm$0.017} \\
CT-CLIP~\cite{hamamci2024developing} & 0.115{\tiny $\pm$0.018} & 0.040{\tiny $\pm$0.015} & 0.140{\tiny $\pm$0.047} & 0.262{\tiny $\pm$0.034} & 0.124{\tiny $\pm$0.020} & 0.127{\tiny $\pm$0.015} & 0.022{\tiny $\pm$0.014} & 0.188{\tiny $\pm$0.035} & 0.308{\tiny $\pm$0.024} & 0.147{\tiny $\pm$0.017} & 0.113{\tiny $\pm$0.016} & 0.026{\tiny $\pm$0.020} & 0.156{\tiny $\pm$0.033} & 0.282{\tiny $\pm$0.043} & 0.127{\tiny $\pm$0.019} \\
CT-Vocab~\cite{hamamci2024developing} & 0.093{\tiny $\pm$0.001} & 0.020{\tiny $\pm$0.001} & 0.110{\tiny $\pm$0.013} & 0.230{\tiny $\pm$0.010} & 0.102{\tiny $\pm$0.003} & 0.093{\tiny $\pm$0.003} & 0.022{\tiny $\pm$0.006} & 0.108{\tiny $\pm$0.013} & 0.208{\tiny $\pm$0.022} & 0.096{\tiny $\pm$0.007} & 0.090{\tiny $\pm$0.001} & 0.020{\tiny $\pm$0.001} & 0.102{\tiny $\pm$0.011} & 0.196{\tiny $\pm$0.020} & 0.090{\tiny $\pm$0.006} \\
CT-LiPro~\cite{hamamci2024foundation} & 0.106{\tiny $\pm$0.013} & 0.034{\tiny $\pm$0.009} & 0.116{\tiny $\pm$0.025} & 0.228{\tiny $\pm$0.042} & 0.109{\tiny $\pm$0.020} & 0.120{\tiny $\pm$0.013} & 0.028{\tiny $\pm$0.013} & 0.168{\tiny $\pm$0.013} & 0.256{\tiny $\pm$0.039} & 0.073{\tiny $\pm$0.015} & 0.084{\tiny $\pm$0.011} & 0.014{\tiny $\pm$0.013} & 0.092{\tiny $\pm$0.020} & 0.154{\tiny $\pm$0.027} & 0.073{\tiny $\pm$0.015} \\
Dinov3-Clip~\cite{duriantaco2025dinov3clip} & 0.091{\tiny $\pm$0.008} & 0.020{\tiny $\pm$0.009} & 0.104{\tiny $\pm$0.015} & 0.212{\tiny $\pm$0.022} & 0.095{\tiny $\pm$0.011} & 0.092{\tiny $\pm$0.003} & 0.022{\tiny $\pm$0.006} & 0.104{\tiny $\pm$0.020} & 0.204{\tiny $\pm$0.012} & 0.093{\tiny $\pm$0.005} & 0.094{\tiny $\pm$0.004} & 0.022{\tiny $\pm$0.006} & 0.106{\tiny $\pm$0.013} & 0.216{\tiny $\pm$0.023} & 0.098{\tiny $\pm$0.006} \\
\textbf{Ours} & \textbf{0.641}{\tiny $\pm$0.037} & \textbf{0.508}{\tiny $\pm$0.055} & \textbf{0.804}{\tiny $\pm$0.041} & \textbf{0.908}{\tiny $\pm$0.030} & \textbf{0.700}{\tiny $\pm$0.031} & \textbf{0.646}{\tiny $\pm$0.056} & \textbf{0.490}{\tiny $\pm$0.071} & \textbf{0.864}{\tiny $\pm$0.060} & \textbf{0.942}{\tiny $\pm$0.042} & \textbf{0.715}{\tiny $\pm$0.052} & \textbf{0.638}{\tiny $\pm$0.046} & \textbf{0.498}{\tiny $\pm$0.065} & \textbf{0.808}{\tiny $\pm$0.049} & \textbf{0.912}{\tiny $\pm$0.039} & \textbf{0.704}{\tiny $\pm$0.044} \\
\bottomrule
\end{tabular}}
\label{tab:classification_retrieval}
\end{table*}

\noindent\textbf{Chunk-Wise and Organ-Wise Linear-Probe Classification and Free-Form Retrieval.}
During training, we chunk each volume into blocks and generate masks and reports at the block level. In Table~\ref{tab:classification_retrieval}, the first part evaluates a vision-only setting, where we attach a linear probe to the generated embeddings and perform 12-organ classification. The second part reports CLIP-style retrieval results, evaluated on the first 50 validation samples from the validation subset.

Chunk-wise downstream tasks are highly relevant for volumetric data, as organs change gradually along the axial axis, so blocks preserve coherent local anatomy and avoid whole-volume label dilution. Supervision is local by nature (patching and organ-wise observations), and aligning it at the block level reduces label leakage and improves feature-label consistency. Fixed-length blocks normalize variable scan lengths, allowing batching. They also enable class re-balancing and hard-example mining while yielding localizable and interpretable outputs. For cross-modal retrieval, text can target the most relevant spatio-temporal neighborhoods, and boundary uncertainty is mitigated by overlapping chunks with consistency and multi-shot supervision.

In Table~\ref{tab:classification_retrieval} Classification (Linear Probe), we average slice embeddings along the z-axis for the 2D models. DINOv3 achieves good performance, especially when the number of slices is small. However, as the number of slices increases, its performance degrades noticeably, even though the model is applied once per slice. In contrast, our model consistently improves as the number of slices grows, while being applied once per block. This trend indicates that in a true 3D setting, our SigVLP extracts more accurate and spatially consistent information. Averaging 2D features across slices, even for large models such as DINOv3, inevitably loses valuable 3D context when the number of slices increases.

In Table~\ref{tab:classification_retrieval} Free-form Image to Report Retrieval, we evaluate the ability of models to perform text-image multimodal alignment at the chunk level. DINOv3-CLIP, which combines OpenCLIP’s text encoder with a lightweight adapter on top of DINOv3, performs well on general natural image datasets (\emph{e.g.}, COCO~\cite{lin2014microsoft}). However, it struggles in medical imaging scenarios, where spatial and anatomical precision are crucial. CT-CLIP, a domain-specific CLIP variant, performs better when the input volume is close to the full-volume setting it was trained on (\emph{e.g.}, 128 slices). In both cases, our model demonstrates a substantial advantage. While CT-CLIP attempts to captures the global semantics of the entire CT volume, our modelgoes further — it understands where each part belongs and what it represents. --–

\section{Ablation Study}
\label{sec:ablatioon}

We present ablations exploring different configurations, including variations in RoPE parameter $b$ (which controls the positional rotation frequency in embedding space), different numbers of input slices, and various model training stages. 

\begin{figure}[t]
    \centering
    \includegraphics[width=0.99\linewidth]{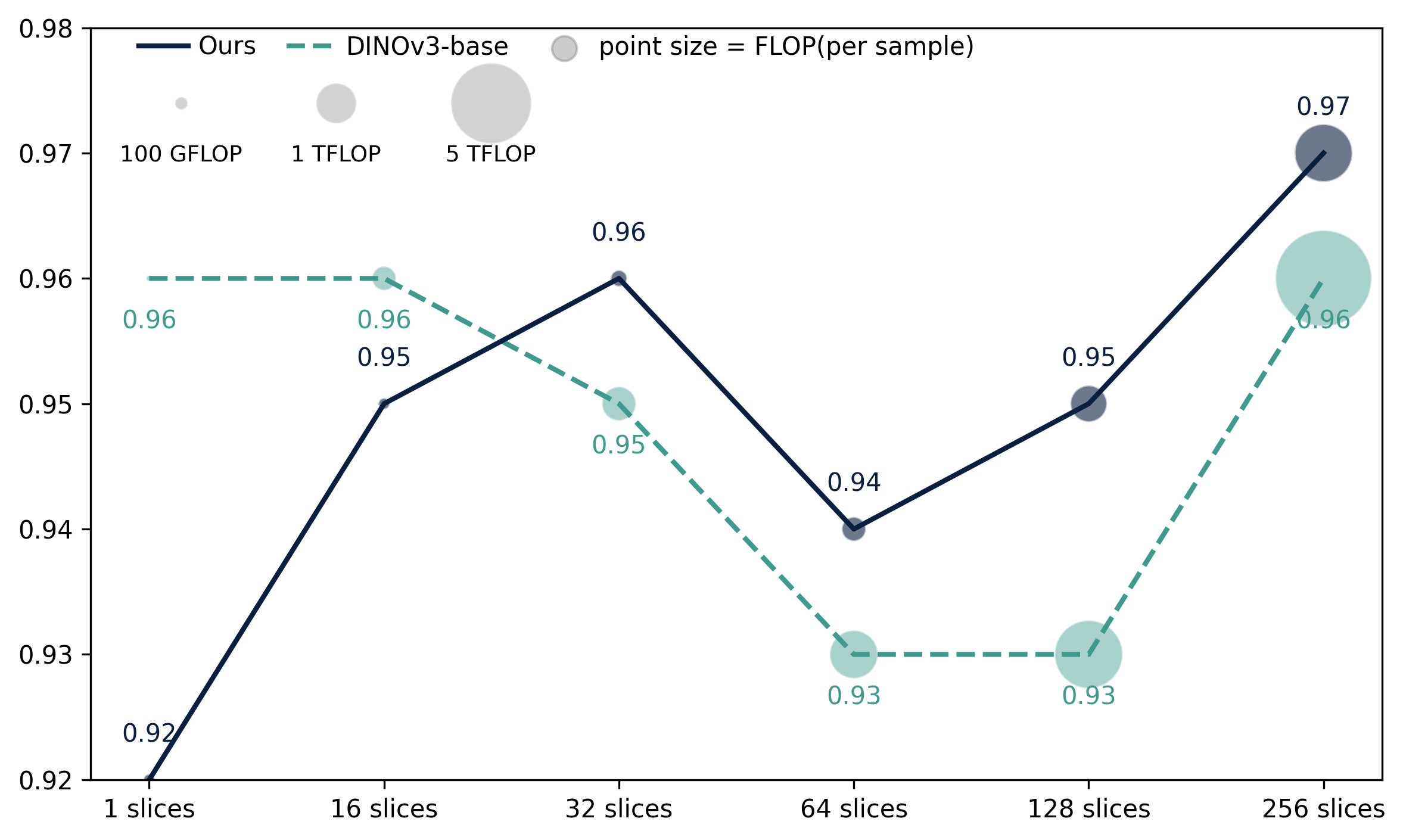}
    \caption{Classification Linear probe F1 score vs. number of slices. Dashed line: DINOv3-Base; solid line: SigVLP; point size shows required floating-point operations (FLOP).  }
    \label{fig:F1-curve}
\end{figure}

%

\noindent\textbf{Input Slice Count Variation.}
To explore the limits of our model's 3D-focused architecture, we evaluate its performance on a 2D image benchmark. Due to architectural constraints, we rely on slice-repetition to fill-in the 16-slice patch-encoder's input.
%
Results are shown in Fig.~\ref{fig:F1-curve}. We observe that even with a single image as input, our model achieves 92\% F1 score, beating many of the baselines shown in Table~\ref{tab:classification_retrieval} at other slice counts.
Nonetheless, the F1 score is clearly lower in the extreme 1-slice case, compared to the 16-slices to 256-slices results.
Additionally, we compared our model to DINOv3-base (second best model in Table~\ref{tab:classification_retrieval} after ours).
We observe that, DINOv3-base has a clear advantage when evaluated on a single slice, but that advantage quickly disappears as the number of slices increases. This is despite the fact that DINOv3-base is evaluated once per slice, and the results are ensembled, giving it a clear advantage over our model which is only ran once per block of slices.





\noindent\textbf{Variation of Rotary Embedding default in evaluation.} 
During training, we set $b_1 = 1000$. While the default $b$ for RoPE in video transformers is typically $10000$, our data chunks are relatively short and of moderate resolution (\emph{i.e.}, $256\text{ resolution} / 16\text{ patch size} \times 128\text{ slices} / 16\text{ patch size} = 2048$ embeddings). Using $b = 10000$ would result in excessively low-frequency positional encoding, leaving much of the representational range underutilized. Therefore, setting $b_1 = 1000$ better matches our spatio-temporal scale.

In this evaluation, we examine the effect of different $b$ values. Although these settings are only changed at inference, we are interested in how RoPE frequency scaling influences performance. As shown in Table~\ref{tab:ours2000-theta-slices}, smaller $b$ values yield slightly better results. This can be attributed to their their ability to preserve higher-frequency variations and maintain finer spatio–temporal structure within short-range context.

\begin{table}[ht]
\centering
\footnotesize
\caption{Performance of SigVLP's report-to-volume retrieval  across slice counts and $b$ values with max. rank 100.
}
\setlength{\tabcolsep}{5pt}
\begin{tabular}{lcccc}
\hline
$b$ & R@5$\uparrow$ & R@10$\uparrow$ & R@50$\uparrow$ & MeanRank$\downarrow$ \\
\hline
$0.5b_1$ 
& 0.399{\tiny$\pm$0.049} & 0.547{\tiny$\pm$0.037} & 0.915{\tiny$\pm$0.026} & 16.69{\tiny$\pm$1.70} \\
$b_1$ 
& 0.636{\tiny$\pm$0.050} & 0.769{\tiny$\pm$0.036} & 0.978{\tiny$\pm$0.010} & 
8.23{\tiny$\pm$0.947} \\
$2b_1$ 
& 0.351{\tiny$\pm$0.052} & 0.480{\tiny$\pm$0.048} & 0.899{\tiny$\pm$0.025} & 19.44{\tiny$\pm$1.85} \\
\hline
\end{tabular}
\label{tab:ours2000-theta-slices}
\end{table}

\begin{figure}[t]
  \centering
  \setlength{\tabcolsep}{2pt}

  \newcommand{\vlabel}[4][0pt]{%
    \raisebox{#1}{\rotatebox[origin=c]{90}{\footnotesize\textbf{#2}}}%
  }

  \begin{tabular}{@{}c c c c@{}}
    & {\footnotesize\textbf{32 slices}} & {\footnotesize\textbf{64 slices}} & {\footnotesize\textbf{128 slices}} \\
    \vlabel[24pt]{step 2000} &
    \includegraphics[width=0.29\linewidth]{"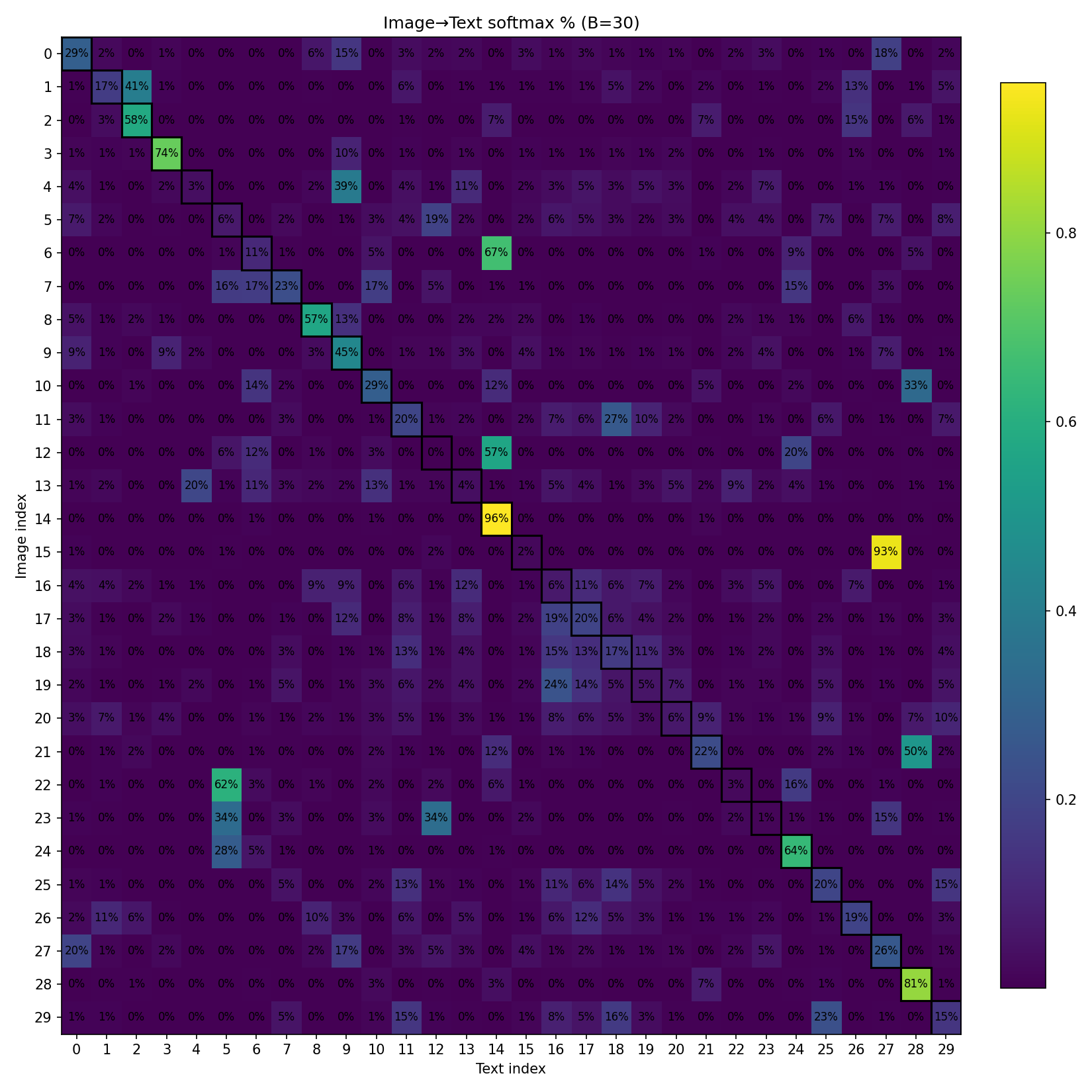"} &
    \includegraphics[width=0.29\linewidth]{"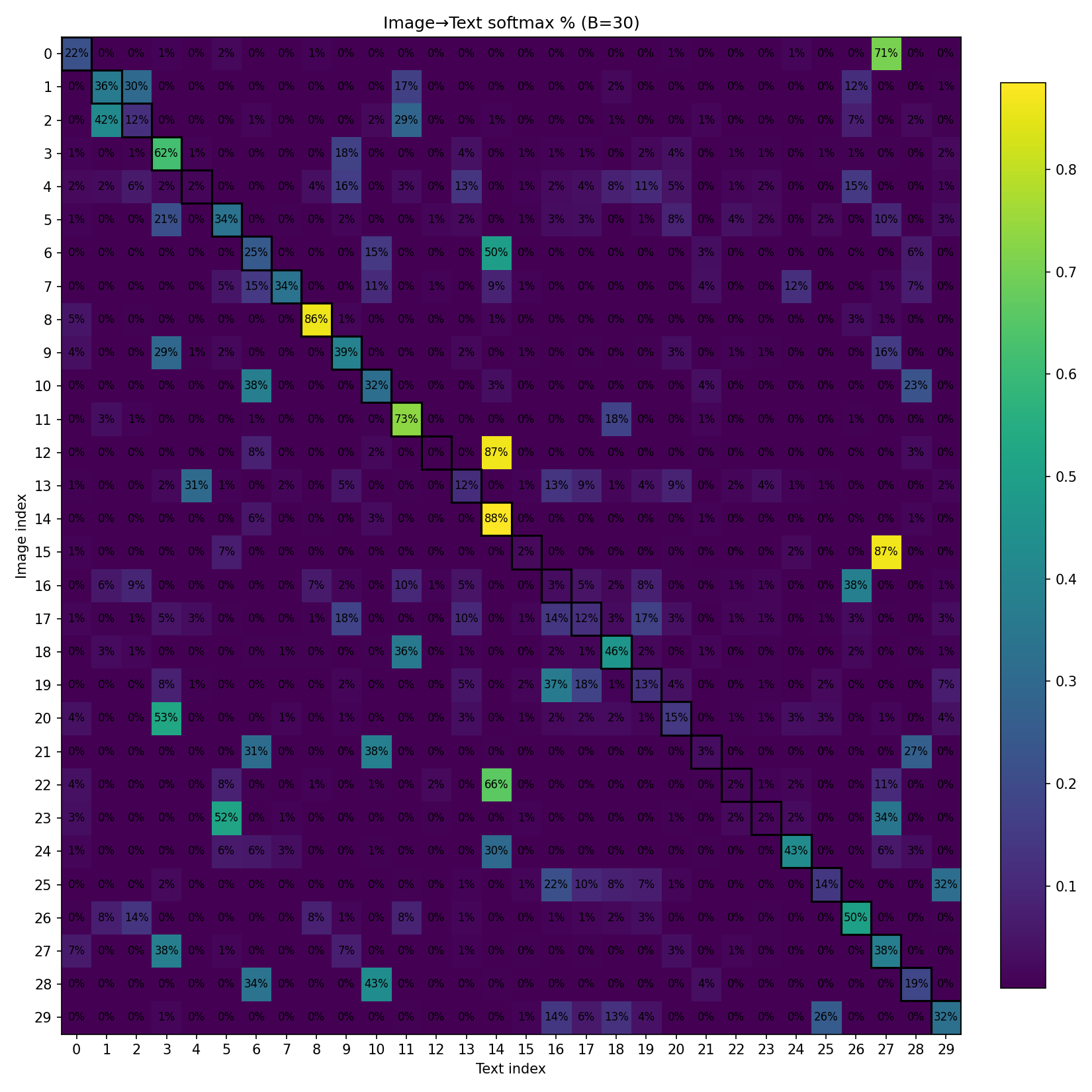"} &
    \includegraphics[width=0.29\linewidth]{"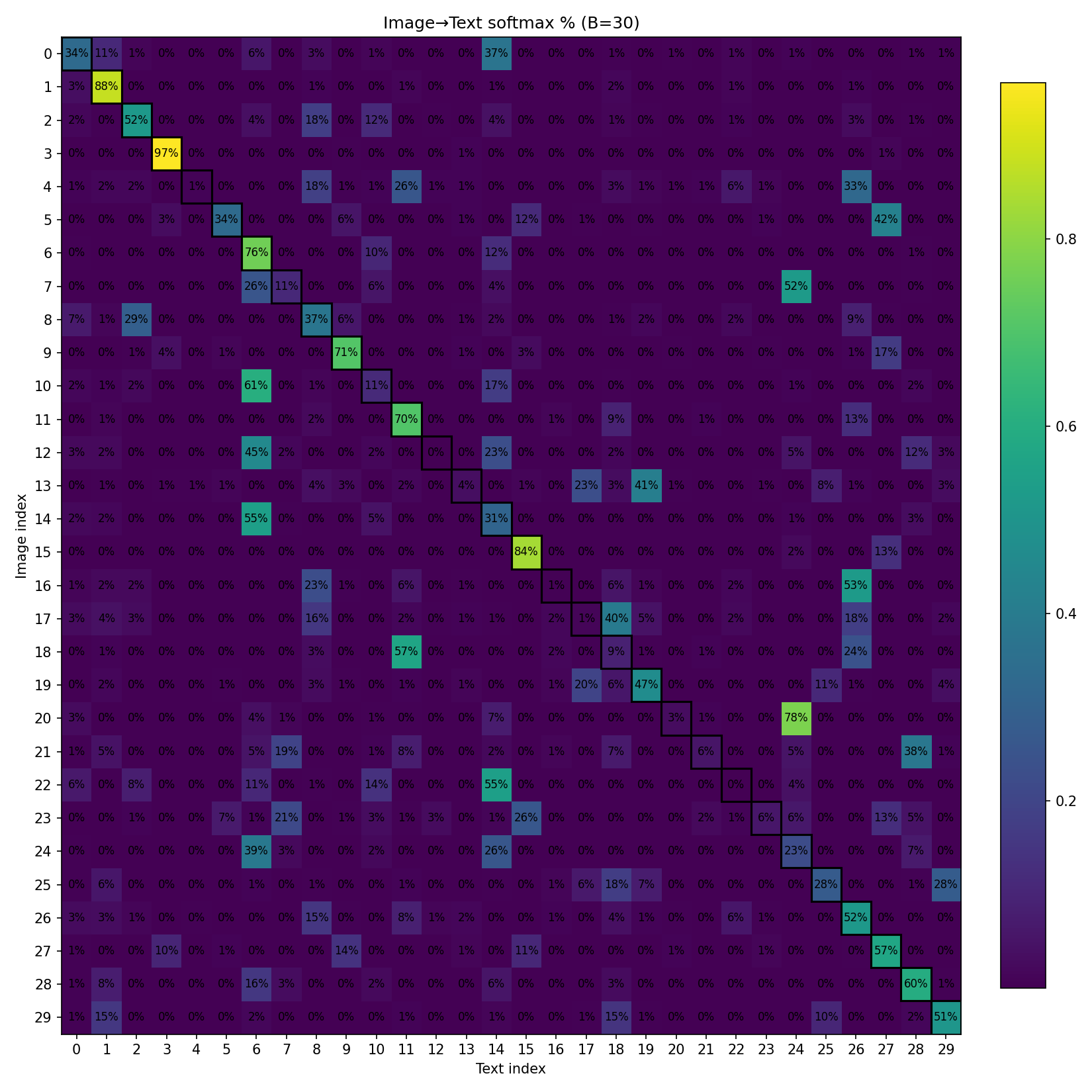"} \\
    \vlabel[24pt]{step 4000} &
    \includegraphics[width=0.29\linewidth]{"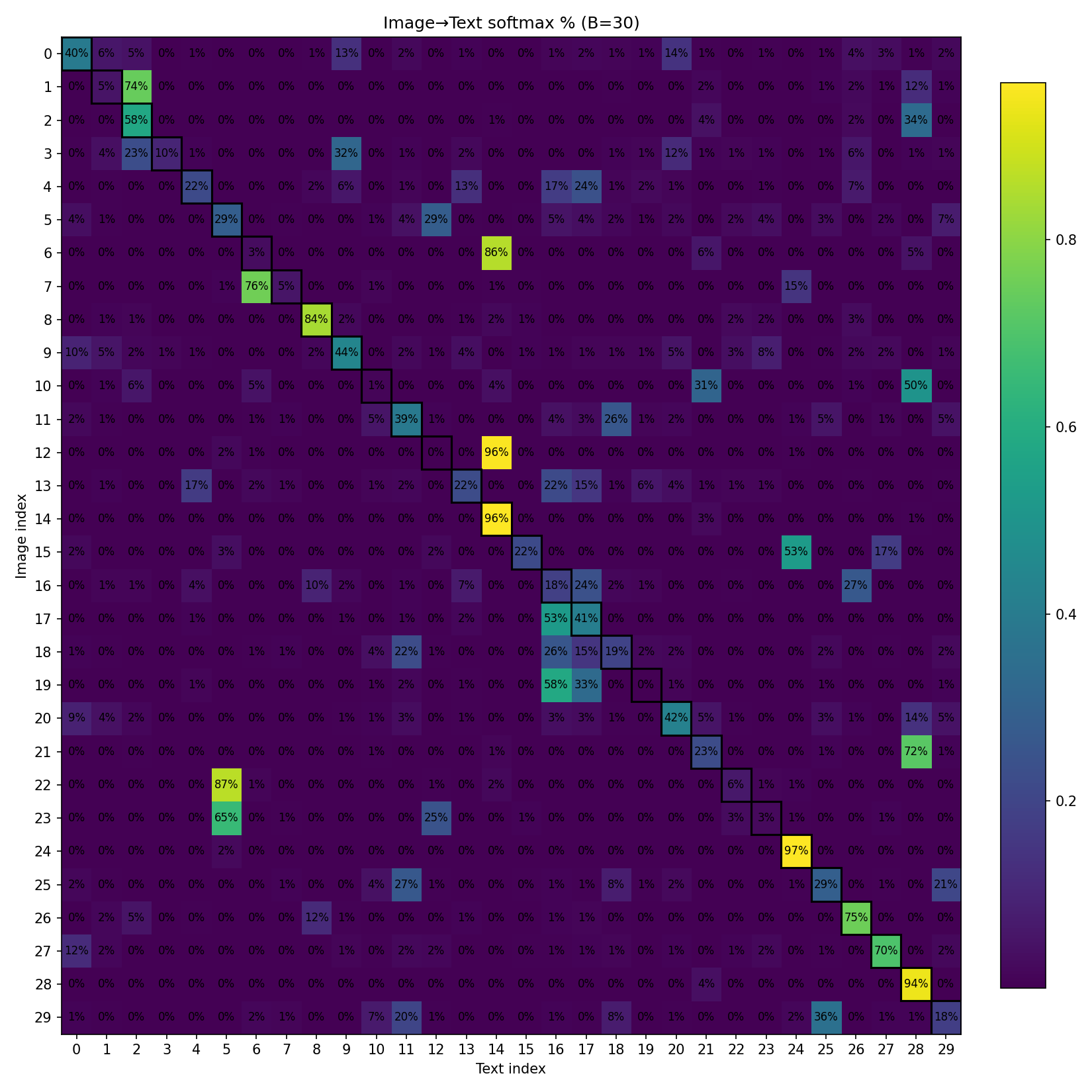"} &
    \includegraphics[width=0.29\linewidth]{"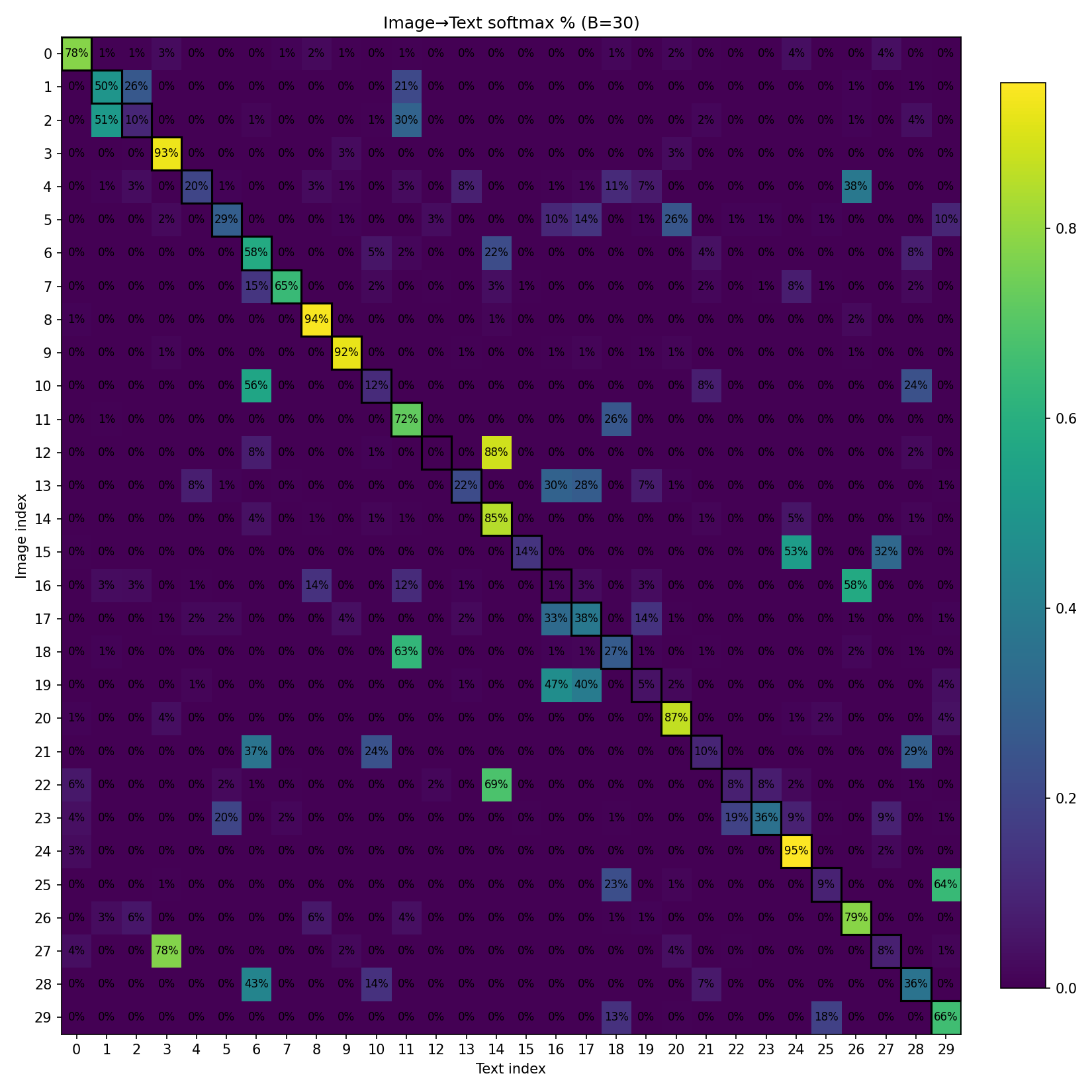"} &
    \includegraphics[width=0.29\linewidth]{"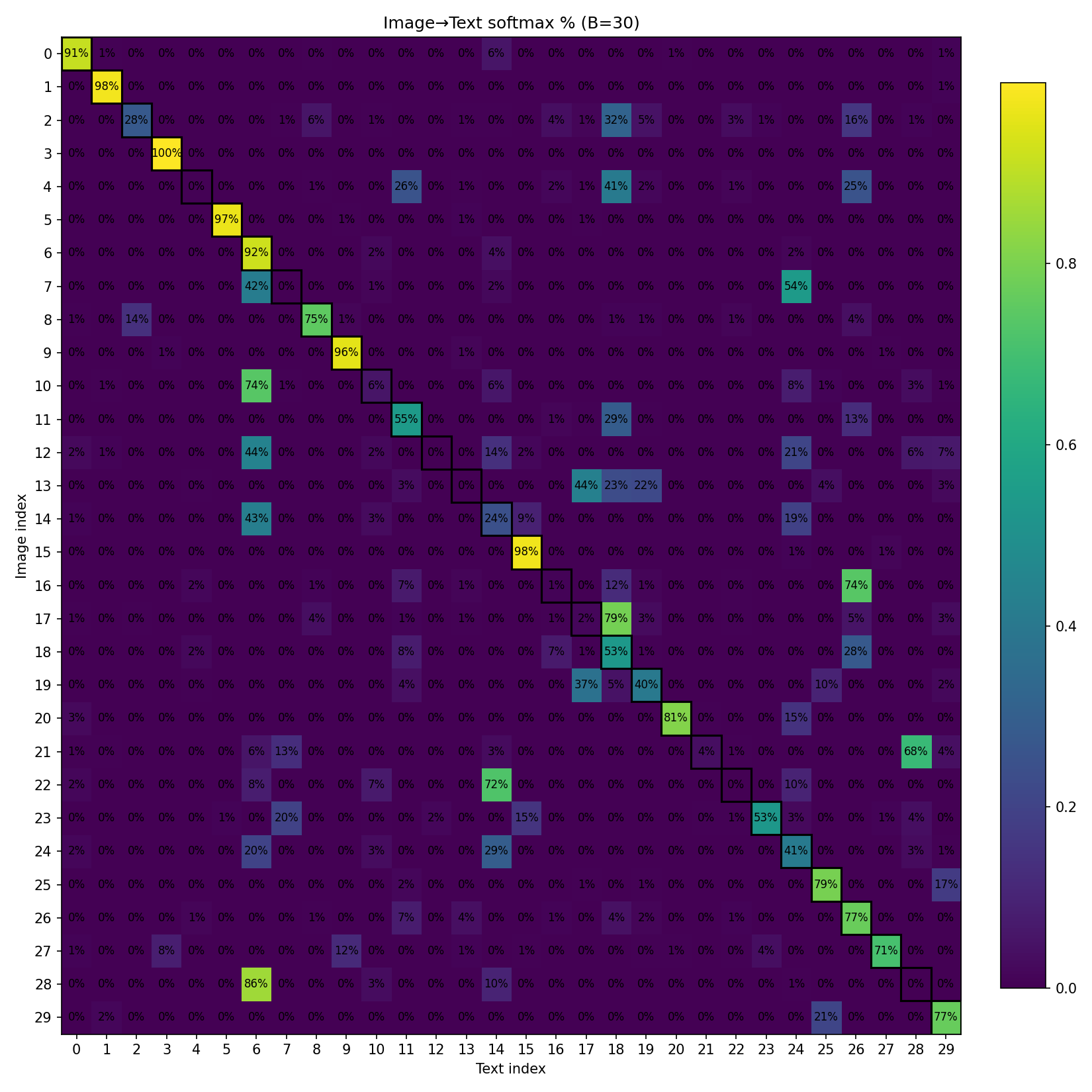"} \\
    \vlabel[24pt]{step 230000} &
    \includegraphics[width=0.29\linewidth]{"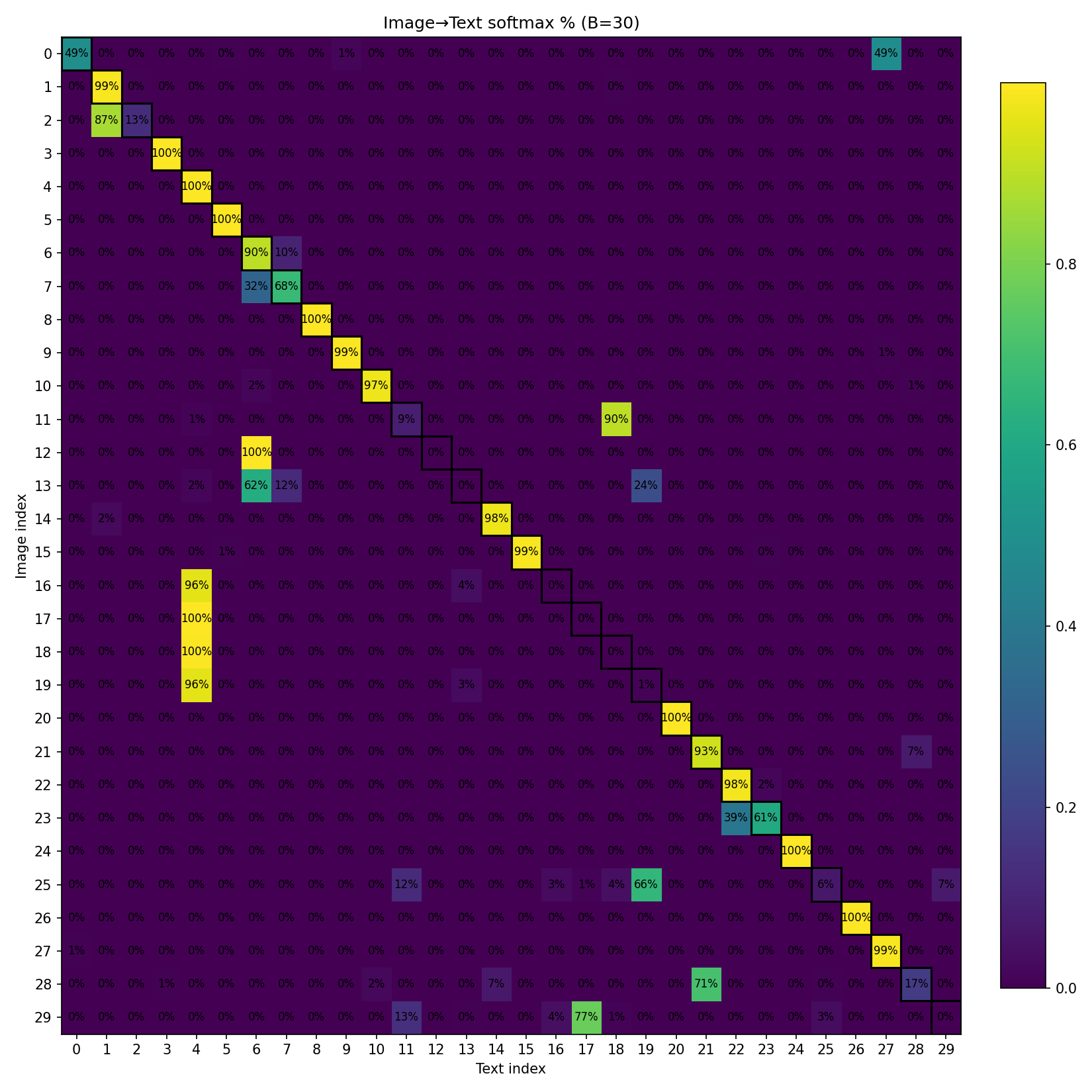"} &
    \includegraphics[width=0.29\linewidth]{"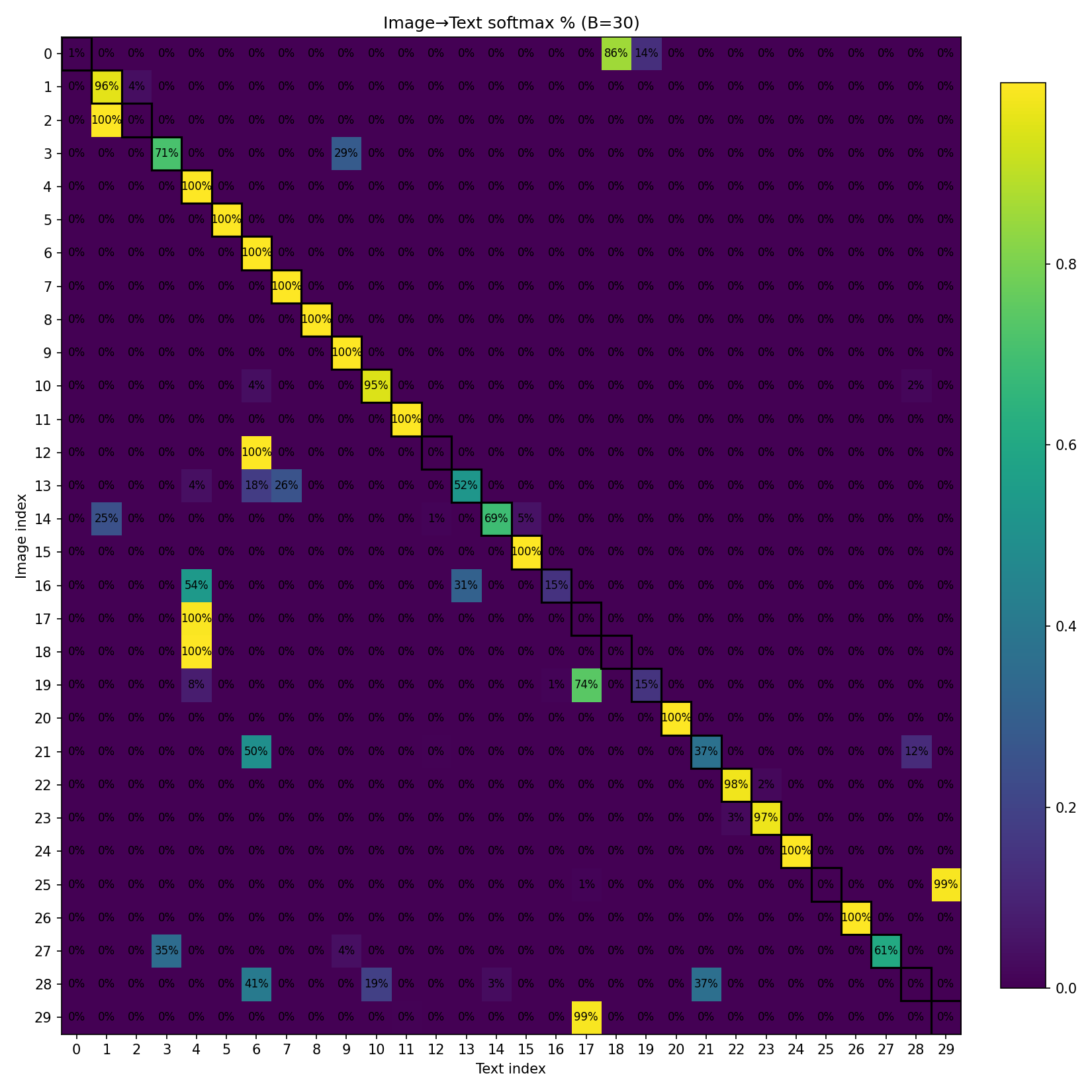"} &
    \includegraphics[width=0.29\linewidth]{"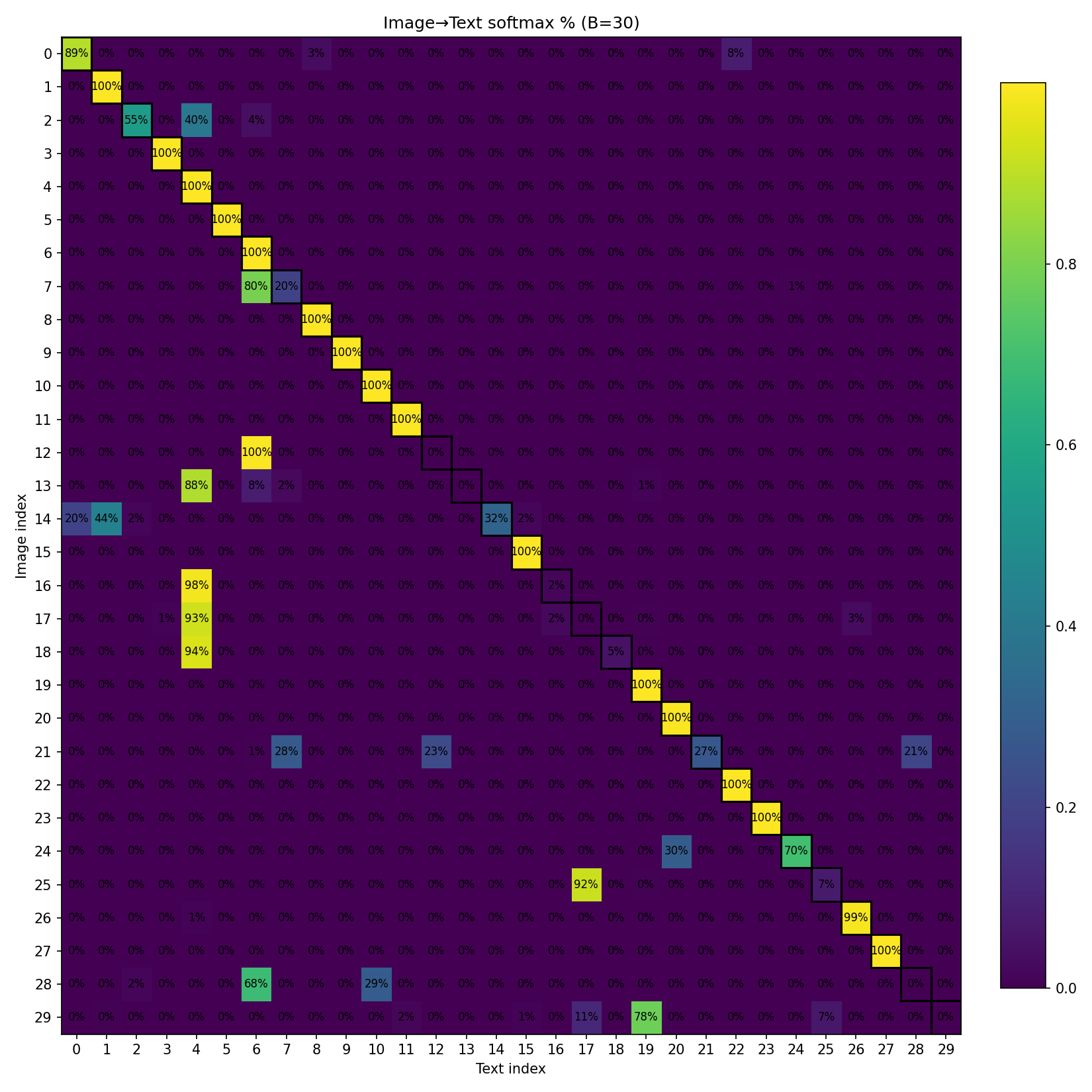"} \\
  \end{tabular}

  \vspace{0.25em}
  \caption{Retrieval of 50 paired samples across model sizes.}
  \label{fig:methods_slices_grid_singlecol}
  \vspace{-0.5em}
\end{figure}

\noindent\textbf{Retrieval experiments across different input slice counts and model training stages.}
In Fig.~\ref{fig:methods_slices_grid_singlecol}, we present the retrieval heatmaps of 50 text-image pairs from the validation subset while changing the model training stage and input slice counts. The diagonal entries correspond to the ground-truth alignments. As model training progresses, high-similarity scores become more concentrated along the diagonal with fewer off-diagonal activations, yielding a cleaner and more distinct alignment pattern. This indicates that, with more training the ambiguity decreases, achieving higher precision cross-modal matching. For our final model, the diagonal concentration is the most pronounced.

\section{Conclusion}
\label{sec:conclusion}

We have introduced SigVLP, a volumetric vision-language pretraining approach that removes the fixed-depth constraint of conventional 3D architectures by integrating RoPE directly into the attention mechanism. This enables flexible handling of variable-length CT volumes while preserving anatomical continuity and local spatial semantics. By coupling volumetric chunk-wise sampling with organ-aware textual alignment, SigVLP achieves fine-grained visual-textual correspondence and robust generalization across heterogeneous datasets. Our evaluation demonstrates consistent gains over strong baselines such as DINOv3 and CT-CLIP 
with pronounced improvements in localized, block-wise evaluation. 
Our findings suggest that adaptive positional encoding and localized supervision can form a scalable foundation for volumetric vision-language models enabling foundation models that jointly reason over anatomy and language. 
Source code and pretrained weights will be released on GitHub and Hugging Face by the time of the conference.
{
    \small
    \bibliographystyle{ieeenat_fullname}
    \bibliography{main}

@String(CVPR= {IEEE Conf. Comput. Vis. Pattern Recog.})

@String(ICCV= {Int. Conf. Comput. Vis.})

@String(ECCV= {Eur. Conf. Comput. Vis.})

@String(NIPS= {Adv. Neural Inform. Process. Syst.})

@String(ICML = {Int. Conf. Mach. Learn.})

@String(AAAI = {AAAI})

@String(CVPR  = {CVPR})

@String(ICCV  = {ICCV})

@String(ICML  = {ICML})

@String(ECCV  = {ECCV})

@String(NIPS  = {NeurIPS})

@inproceedings{li2021albef,
  title     = {{Align before Fuse: Vision and Language Representation Learning with Momentum Distillation}},
  author    = {Li, Junnan and Selvaraju, Ramprasaath and Gotmare, Akhilesh and Joty, Shafiq and Xiong, Caiming and Hoi, Steven Chu Hong},
  booktitle = NIPS,
  volume    = {34},
  pages     = {9694--9705},
  year      = {2021},
  url       = {https://proceedings.neurips.cc/paper_files/paper/2021/file/505259756244493872b7709a8a01b536-Paper.pdf}
}

@inproceedings{li2022blip,
  title     = {{BLIP: Bootstrapping Language-Image Pre-training for Unified Vision-Language Understanding and Generation}},
  author    = {Li, Junnan and Li, Dongxu and Xiong, Caiming and Hoi, Steven C.H.},
  booktitle = ICML,
  pages     = {12888--12900},
  year      = {2022},
  volume    = {162},
  publisher = {PMLR},
  url       = {https://proceedings.mlr.press/v162/li22n.html}
}

@inproceedings{zhang2022convirt,
  title     = {{ConVIRT: Contrastive Learning of Medical Visual Representations from Paired Images and Text}},
  author    = {Zhang, Yuhao and Jiang, Hang and Miura, Yasuhide and Manning, Christopher D. and Langlotz, Curtis P.},
  booktitle = {Proc. Mach. Learn. Healthc. Conf. (MLHC)},
  pages     = {2--25},
  year      = {2022},
  publisher = {PMLR},
  url       = {https://proceedings.mlr.press/v182/zhang22a.html}
}

@article{heller2020kits19,
  title     = {{The KiTS19 Challenge Data: 300 Kidney Tumor Cases with Clinical Context, CT Semantic Segmentations, and Surgical Outcomes}},
  author    = {Heller, Nicholas and Sathianathen, Niranjan and Kalapara, Arveen and Walczak, Edward and Moore, Keenan and Kaluzniak, Heather and Rosenberg, Joel and Blake, Paul and Rengel, Zachary and Oestreich, Makinna and Dean, Joshua and Tradewell, Michael and Shah, Aneri and Tejpaul, Resha and Edgerton, Zachary and Peterson, Matthew and Raza, Shaneabbas and Regmi, Subodh and Papanikolopoulos, Nikolaos and Weight, Christopher},
  journal   = {arXiv preprint arXiv:1904.00445},
  year      = {2020},
  url       = {https://arxiv.org/abs/1904.00445}
}

@article{bilic2019lits,
  title     = {{The Liver Tumor Segmentation Benchmark (LiTS)}},
  author    = {Bilic, Patrick and Christ, Philipp F. and Vorontsov, Eugene and Chlebus, Grzegorz and Chen, Hao and Dou, Qi and Fu, Chi-Tung and Han, Xiaohong and Heng, Pheng-Ann and Hesser, J{\"u}rgen and others},
  journal   = {arXiv preprint arXiv:1901.04056},
  year      = {2019},
  url       = {https://arxiv.org/abs/1901.04056}
}

@article{bakas2017brats,
  title     = {{Advancing The Cancer Genome Atlas Glioma MRI Collections with Expert Segmentation Labels and Radiomic Features}},
  author    = {Bakas, Spyridon and Akbari, Hamed and Sotiras, Aristeidis and Bilello, Michel and Rozycki, Matthew and Kirby, Justin S. and Freymann, John B. and Farahani, Keyvan and Davatzikos, Christos},
  journal   = {Sci. Data},
  volume    = {4},
  pages     = {170117},
  year      = {2017},
  doi       = {10.1038/sdata.2017.117},
  url       = {https://www.nature.com/articles/sdata2017117}
}

@inproceedings{yan2022clinicalbert,
  title     = {{Clinical-BERT: Vision-Language Pre-training for Radiograph Diagnosis and Reports Generation}},
  author    = {Yan, Bin and Pei, Minghui},
  booktitle = AAAI,
  volume    = {36},
  number    = {3},
  pages     = {2982--2990},
  year      = {2022},
  doi       = {10.1609/aaai.v36i3.20204},
  url       = {https://doi.org/10.1609/aaai.v36i3.20204}
}

@inproceedings{boecking2022textsemantics,
  title     = {{Making the Most of Text Semantics to Improve Biomedical Vision--Language Processing}},
  author    = {Boecking, Benedikt and Usuyama, Naoto and Bannur, Shruthi and Castro, Daniel C. and Schwaighofer, Anton and Hyland, Stephanie and Wetscherek, Marc and Naumann, Tristan and Nori, Aditya and Alvarez-Valle, Jorge and Poon, Hoifung and Oktay, Ozan},
  booktitle = {Lecture Notes in Computer Science (LNCS)},
  pages     = {1--21},
  year      = {2022},
  publisher = {Springer},
  doi       = {10.1007/978-3-031-20059-5_1},
  url       = {https://doi.org/10.1007/978-3-031-20059-5_1}
}

@inproceedings{huang2024knowledge,
  title     = {{Exploring Vision Language Pretraining with Knowledge Enhancement via Large Language Model}},
  author    = {Tung, Cheng and Lin, Yu and Yin, Jianfeng and Ye, Qi and Chen, Hongming},
  booktitle = {Lecture Notes in Computer Science (LNCS)},
  year      = {2024},
  publisher = {Springer},
  doi       = {10.1007/978-3-031-67751-9_7},
  url       = {https://link.springer.com/chapter/10.1007/978-3-031-67751-9_7}
}

@inproceedings{wu2023medklip,
  title     = {{MedKLIP: Medical Knowledge Enhanced Language-Image Pre-Training for X-ray Diagnosis}},
  author    = {Wu, Jian and Zhang, Xiaohong and Zhang, Yicheng and Wang, Yujie and Xie, Weidi},
  booktitle = ICCV,
  year      = {2023},
  pages     = {19493--19503},
  url       = {http://openaccess.thecvf.com/content/ICCV2023/html/Wu_MedKLIP_Medical_Knowledge_Enhanced_Language-Image_Pre-Training_for_X-ray_Diagnosis_ICCV_2023_paper.html}
}

@article{zhang2023kad,
  title     = {{Knowledge-Enhanced Visual-Language Pre-Training on Chest Radiology Images}},
  author    = {Zhang, Xiaohong and Wu, Chuang and Zhang, Yuxuan and Jin, Qiu and Tang, Yuxing and Duan, Jianglin and Bai, Haotian and Xu, Dong and Zhou, Hong-Yu and Sun, Linyuan and others},
  journal   = {Nat. Commun.},
  volume    = {14},
  pages     = {4542},
  year      = {2023},
  doi       = {10.1038/s41467-023-40260-7},
  url       = {https://doi.org/10.1038/s41467-023-40260-7}
}

@inproceedings{wang2022medclip,
  title     = {{MedCLIP: Contrastive Learning from Unpaired Medical Images and Text}},
  author    = {Wang, Zifeng and Wu, Zhenbang and Agarwal, Dinesh and Sun, Jimeng},
  booktitle = {Proc. Conf. Empirical Methods in Natural Language Processing (EMNLP)},
  pages     = {3876--3887},
  year      = {2022},
  address   = {Abu Dhabi, United Arab Emirates},
  publisher = {Association for Computational Linguistics},
  doi       = {10.18653/v1/2022.emnlp-main.256},
  url       = {https://aclanthology.org/2022.emnlp-main.256}
}

@inproceedings{chen2023ptunifier,
  title     = {{Towards Unifying Medical Vision-and-Language Pre-Training via Soft Prompts}},
  author    = {Chen, Zhihong and Diao, Shizhe and Wang, Benyou and Li, Guanbin and Wan, Xiang},
  booktitle = ICCV,
  year      = {2023},
  pages     = {23403--23413},
  url       = {https://openaccess.thecvf.com/content/ICCV2023/html/Chen_Towards_Unifying_Medical_Vision-and-Language_Pre-Training_via_Soft_Prompts_ICCV_2023_paper.html}
}

@inproceedings{xie2024pairaug,
  title     = {{PairAug: What Can Augmented Image-Text Pairs Do for Radiology?}},
  author    = {Xie, Yutong and Chen, Qi and Wang, Sinuo and To, Minh-Son and Lee, Iris and Khoo, Ee Win and Hendy, Kerolos and Koh, Daniel and Xia, Yong and Wu, Qi},
  booktitle = CVPR,
  year      = {2024},
  pages     = {11652--11657},
  url       = {https://openaccess.thecvf.com/content/CVPR2024/html/Xie_PairAug_What_Can_Augmented_Image-Text_Pairs_Do_for_Radiology_CVPR_2024_paper.html}
}

@article{shui2025organclip,
  title     = {{Large-scale and Fine-grained Vision-Language Pre-Training for Enhanced CT Image Understanding}},
  author    = {Shui, Zhongyi and Zhang, Jianpeng and Cao, Weiwei and Wang, Sinuo and Guo, Ruizhe and Lu, Le and Yang, Lin and Ye, Xianghua and Liang, Tingbo and Zhang, Qi and Zhang, Ling},
  journal   = {arXiv preprint arXiv:2501.14548},
  year      = {2025},
  url       = {https://arxiv.org/abs/2501.14548}
}

@article{blankemeier2024merlin,
  title     = {{Merlin: A Vision Language Foundation Model for 3D Computed Tomography}},
  author    = {Blankemeier, Louis and Cohen, Joseph Paul and Kumar, Ashwin and Van Veen, Dave and Gardezi, Syed Jamal Safdar and Paschali, Magdalini and Chen, Zhihong and Delbrouck, Jean-Benoit and Reis, Eduardo and Truyts, Cesar and others},
  journal   = {arXiv preprint arXiv:2406.06512},
  year      = {2024},
  url       = {https://arxiv.org/abs/2406.06512}
}

@inproceedings{radford2021learning,
  title={Learning transferable visual models from natural language supervision},
  author={Radford, Alec and Kim, Jong Wook and Hallacy, Chris and Ramesh, Aditya and Goh, Gabriel and Agarwal, Sandhini and Sastry, Girish and Askell, Amanda and Mishkin, Pamela and Clark, Jack and others},
  booktitle={International conference on machine learning},
  pages={8748--8763},
  year={2021},
  organization={PmLR}
}

@article{johnson2019mimic,
  title={MIMIC-CXR, a de-identified publicly available database of chest radiographs with free-text reports},
  author={Johnson, Alistair EW and Pollard, Tom J and Berkowitz, Seth J and Greenbaum, Nathaniel R and Lungren, Matthew P and Deng, Chih-ying and Mark, Roger G and Horng, Steven},
  journal={Scientific data},
  volume={6},
  number={1},
  pages={317},
  year={2019},
  publisher={Nature Publishing Group UK London}
}

@article{wasserthal2023totalsegmentator,
  title={TotalSegmentator: robust segmentation of 104 anatomic structures in CT images},
  author={Wasserthal, Jakob and Breit, Hanns-Christian and Meyer, Manfred T and Pradella, Maurice and Hinck, Daniel and Sauter, Alexander W and Heye, Tobias and Boll, Daniel T and Cyriac, Joshy and Yang, Shan and others},
  journal={Radiology: Artificial Intelligence},
  volume={5},
  number={5},
  pages={e230024},
  year={2023},
  publisher={Radiological Society of North America}
}

@article{han2022survey,
  title={A survey on vision transformer},
  author={Han, Kai and Wang, Yunhe and Chen, Hanting and Chen, Xinghao and Guo, Jianyuan and Liu, Zhenhua and Tang, Yehui and Xiao, An and Xu, Chunjing and Xu, Yixing and others},
  journal={IEEE transactions on pattern analysis and machine intelligence},
  volume={45},
  number={1},
  pages={87--110},
  year={2022},
  publisher={IEEE}
}

@article{kazemnejad2023impact,
  title     = {{The Impact of Positional Encoding on Length Generalization in Transformers}},
  author    = {Kazemnejad, Amirhossein and Padhi, Inkit and Natesan Ramamurthy, Karthikeyan and Das, Payel and Reddy, Siva},
  journal   = {arXiv preprint arXiv:2305.19466},
  year      = {2023},
  url       = {https://arxiv.org/abs/2305.19466}
}

@inproceedings{wang2025ctflow,
  title={CTFlow: Video-Inspired Latent Flow Matching for 3D CT Synthesis},
  author={Wang, Jiayi and Reynaud, Hadrien and Erick, Franciskus Xaverius and Kainz, Bernhard},
  booktitle={Proceedings of the IEEE/CVF International Conference on Computer Vision},
  pages={6750--6758},
  year={2025}
}

@article{reynaud2025echoflow,
  title={Echoflow: A foundation model for cardiac ultrasound image and video generation},
  author={Reynaud, Hadrien and Gomez, Alberto and Leeson, Paul and Meng, Qingjie and Kainz, Bernhard},
  journal={arXiv preprint arXiv:2503.22357},
  year={2025}
}

@article{touvron2023llama,
  title={Llama: Open and efficient foundation language models},
  author={Touvron, Hugo and Lavril, Thibaut and Izacard, Gautier and Martinet, Xavier and Lachaux, Marie-Anne and Lacroix, Timoth{\'e}e and Rozi{\`e}re, Baptiste and Goyal, Naman and Hambro, Eric and Azhar, Faisal and others},
  journal={arXiv preprint arXiv:2302.13971},
  year={2023}
}

@article{bai2023qwen,
  title={Qwen technical report},
  author={Bai, Jinze and Bai, Shuai and Chu, Yunfei and Cui, Zeyu and Dang, Kai and Deng, Xiaodong and Fan, Yang and Ge, Wenbin and Han, Yu and Huang, Fei and others},
  journal={arXiv preprint arXiv:2309.16609},
  year={2023}
}

@article{team2025kimi,
  title={Kimi k2: Open agentic intelligence},
  author={Team, Kimi and Bai, Yifan and Bao, Yiping and Chen, Guanduo and Chen, Jiahao and Chen, Ningxin and Chen, Ruijue and Chen, Yanru and Chen, Yuankun and Chen, Yutian and others},
  journal={arXiv preprint arXiv:2507.20534},
  year={2025}
}

@article{tschannen2025siglip,
  title={Siglip 2: Multilingual vision-language encoders with improved semantic understanding, localization, and dense features},
  author={Tschannen, Michael and Gritsenko, Alexey and Wang, Xiao and Naeem, Muhammad Ferjad and Alabdulmohsin, Ibrahim and Parthasarathy, Nikhil and Evans, Talfan and Beyer, Lucas and Xia, Ye and Mustafa, Basil and others},
  journal={arXiv preprint arXiv:2502.14786},
  year={2025}
}

@article{hamamci2024developing,
  title={{Developing Generalist Foundation Models from a Multimodal Dataset for 3D Computed Tomography}},
  author={Hamamci, Ibrahim Ethem and Er, Sezgin and Almas, Furkan and Simsek, Ayse Gulnihan and Esirgun, Sevval Nil and Dogan, Irem and Dasdelen, Muhammed Furkan and Durugol, Omer Faruk and Wittmann, Bastian and Amiranashvili, Tamaz and Simsar, Enis and Simsar, Mehmet and Erdemir, Emine Bensu and Alanbay, Abdullah and Sekuboyina, Anjany and Lafci, Berkan and Bluethgen, Christian and Ozdemir, Mehmet Kemal and Menze, Bjoern},
  journal={in press Nature Biomedical Engineering, arXiv preprint arXiv:2403.17834},
  year={2024},
  url={https://arxiv.org/abs/2403.17834}
}

@inproceedings{hamamci2024generatect,
  title={{GenerateCT: Text-Conditional Generation of 3D Chest CT Volumes}},
  author={Hamamci, Ibrahim Ethem and Er, Sezgin and Sekuboyina, Anjany and Simsar, Enis and Tezcan, Alperen and Simsek, Ayse Gulnihan and Esirgun, Sevval Nil and Almas, Furkan and Dogan, Irem and Dasdelen, Muhammed Furkan and others},
  booktitle=ECCV,
  pages={126--143},
  year={2024},
  publisher={Springer}
}

@inproceedings{hamamci2024ct2rep,
  title={{CT2Rep: Automated Radiology Report Generation for 3D Medical Imaging}},
  author={Hamamci, Ibrahim Ethem and Er, Sezgin and Menze, Bjoern},
  booktitle={MICCAI},
  pages={476--486},
  year={2024},
  publisher={Springer}
}

@article{su2023roformer,
  title     = {{RoFormer: Enhanced Transformer with Rotary Position Embedding}},
  author    = {Su, Jianlin and Lu, Yu and Pan, Shengfeng and Murtadha, Ahmed and Wen, Bo and Liu, Yunfeng},
  journal   = {arXiv preprint arXiv:2104.09864},
  year      = {2023},
  url       = {https://arxiv.org/abs/2104.09864}
}

@article{wei2025videorope,
  title     = {{VideoRoPE: What Makes for Good Video Rotary Position Embedding?}},
  author    = {Wei, Xilin and Liu, Xiaoran and Zang, Yuhang and Dong, Xiaoyi and Zhang, Pan and Cao, Yuhang and Tong, Jian and Duan, Haodong and Guo, Qipeng and Wang, Jiaqi and Qiu, Xipeng and Lin, Dahua},
  journal   = {arXiv preprint arXiv:2502.05173},
  year      = {2025},
  url       = {https://arxiv.org/abs/2502.05173}
}

@misc{wei2025videoropepp,
  title        = {{VideoRoPE++: Towards Better Video Rotary Position Embedding}},
  author       = {Wei, Xilin and Liu, Xiaoran and Zang, Yuhang and Ding, Shengyuan and Dong, Xiaoyi and Cao, Yuhang and Duan, Haodong and Guo, Qipeng and Wang, Jiaqi and Qiu, Xipeng and Lin, Dahua},
  year         = {2025},
  howpublished = {\url{https://github.com/Wiselnn570/VideoRoPE/blob/main/videorope_plus/VideoRoPE_plus.pdf}},
  doi          = {10.5281/zenodo.16529245}
}

@article{gao2024tcllava,
  title     = {{TC-LLaVA: Rethinking the Transfer from Image to Video Understanding with Temporal Considerations}},
  author    = {Gao, Mingze and Liu, Jingyu and Li, Mingda and Xie, Jiangtao and Liu, Qingbin and Zhao, Bo and Chen, Xi and Xiong, Hui},
  journal   = {arXiv preprint arXiv:2409.03206},
  year      = {2024},
  url       = {https://arxiv.org/abs/2409.03206}
}

@article{wang2024qwen2vl,
  title     = {{Qwen2-VL: Enhancing Vision-Language Model's Perception of the World at Any Resolution}},
  author    = {Wang, Peng and Bai, Shuai and Tan, Sinan and Wang, Shijie and Fan, Zhihao and Bai, Jinze and Chen, Keqin and Liu, Xuejing and Wang, Jialin and Ge, Wenbin and Fan, Yang and Dang, Kai and Du, Mengfei and Ren, Xuancheng and Men, Rui and Liu, Dayiheng and Zhou, Chang and Zhou, Jingren and Lin, Junyang},
  journal   = {arXiv preprint arXiv:2409.12191},
  year      = {2024},
  url       = {https://arxiv.org/abs/2409.12191}
}

@article{liu2025muon,
  title     = {{Muon is Scalable for LLM Training}},
  author    = {Liu, Jingyuan and Su, Jianlin and Yao, Xingcheng and Jiang, Zhejun and Lai, Guokun and Du, Yulun and Qin, Yidao and Xu, Weixin and Lu, Enzhe and Yan, Junjie and Chen, Yanru and Zheng, Huabin and Liu, Yibo and Liu, Shaowei and Yin, Bohong and He, Weiran and Zhu, Han and Wang, Yuzhi and Wang, Jianzhou and Dong, Mengnan and Zhang, Zheng and Kang, Yongsheng and Zhang, Hao and Xu, Xinran and Zhang, Yutao and Wu, Yuxin and Zhou, Xinyu and Yang, Zhilin},
  journal   = {arXiv preprint arXiv:2502.16982},
  year      = {2025},
  url       = {https://arxiv.org/abs/2502.16982}
}

@misc{jordan2024muon,
  title        = {{Muon: An Optimizer for Hidden Layers in Neural Networks}},
  author       = {Jordan, Keller and Jin, Yuchen and Boza, Vlado and You, Jiacheng and Cesista, Franz and Newhouse, Laker and Bernstein, Jeremy},
  year         = {2024},
  howpublished = {\url{https://kellerjordan.github.io/posts/muon/}}
}

@article{chen2023adamatch,
  title     = {{Fine-Grained Image-Text Alignment in Medical Imaging Enables Explainable Cyclic Image-Report Generation}},
  author    = {Chen, Wenting and Shen, Linlin and Lin, Jingyang and Luo, Jiebo and Li, Xiang and Yuan, Yixuan},
  journal   = {arXiv preprint arXiv:2312.08078},
  year      = {2023},
  url       = {https://arxiv.org/abs/2312.08078}
}

@article{liu2024imitate,
  title     = {{IMITATE: Clinical Prior Guided Hierarchical Vision-Language Pre-Training}},
  author    = {Liu, Che and Cheng, Sibo and Shi, Miaojing and Shah, Anand and Bai, Wenjia and Arcucci, Rossella},
  journal   = {arXiv preprint arXiv:2310.07355},
  year      = {2024},
  url       = {https://arxiv.org/abs/2310.07355}
}

@article{zhang2025velvetmed,
  title     = {{VELVET-Med: Vision and Efficient Language Pre-Training for Volumetric Imaging Tasks in Medicine}},
  author    = {Zhang, Ziyang and Yu, Yang and Yang, Xulei and Yeo, Si Yong},
  journal   = {arXiv preprint arXiv:2508.12108},
  year      = {2025},
  url       = {https://arxiv.org/abs/2508.12108}
}

@article{guo2025deepseekr1,
  author    = {Guo, D. and Yang, D. and Zhang, H. and others},
  title     = {DeepSeek-R1 incentivizes reasoning in LLMs through reinforcement learning},
  journal   = {Nature},
  volume    = {645},
  pages     = {633--638},
  year      = {2025},
  doi       = {10.1038/s41586-025-09422-z}
}

@misc{openai2025chatgpt5,
  author    = {OpenAI},
  title     = {ChatGPT-5},
  year      = {2025},
  howpublished = {\url{https://chat.openai.com/}},
  note      = {Large language model by OpenAI}
}

@article{simeoni2025dinov3,
  author        = {Sim{\'e}oni, Oriane and Vo, Huy V. and Seitzer, Maximilian and Baldassarre, Federico and Oquab, Maxime and Jose, Cijo and Khalidov, Vasil and Szafraniec, Marc and Yi, Seungeun and Ramamonjisoa, Micha{\"e}l and Massa, Francisco and Haziza, Daniel and Wehrstedt, Luca and Wang, Jianyuan and Darcet, Timoth{\'e}e and Moutakanni, Th{\'e}o and Sentana, Leonel and Roberts, Claire and Vedaldi, Andrea and Tolan, Jamie and Brandt, John and Couprie, Camille and Mairal, Julien and J{\'e}gou, Herv{\'e} and Labatut, Patrick and Bojanowski, Piotr},
  title         = {DINOv3},
  journal       = {arXiv preprint arXiv:2508.10104},
  year          = {2025},
  url           = {https://arxiv.org/abs/2508.10104}
}

@misc{duriantaco2025dinov3clip,
  author       = {duriantaco},
  title        = {dinov3clip: Caption‐free adapter that maps DINOv3 image embeddings into CLIP space},
  howpublished = {\url{https://github.com/duriantaco/dinov3clip}},
  year         = {2025}
}

@article{mcinnes2018umap,
  title={Umap: Uniform manifold approximation and projection for dimension reduction},
  author={McInnes, Leland and Healy, John and Melville, James},
  journal={arXiv preprint arXiv:1802.03426},
  year={2018}
}

@article{hamamci2024foundation,
  title={A foundation model utilizing chest ct volumes and radiology reports for supervised-level zero-shot detection of abnormalities},
  author={Hamamci, Ibrahim Ethem and Er, Sezgin and Almas, Furkan and Simsek, Ayse Gulnihan and Esirgun, Sevval Nil and Dogan, Irem and Dasdelen, Muhammed Furkan and Wittmann, Bastian and Simsar, Enis and Simsar, Mehmet and others},
  journal={CoRR},
  year={2024}
}

@inproceedings{zhai2023sigmoid,
  title={Sigmoid loss for language image pre-training},
  author={Zhai, Xiaohua and Mustafa, Basil and Kolesnikov, Alexander and Beyer, Lucas},
  booktitle={Proceedings of the IEEE/CVF international conference on computer vision},
  pages={11975--11986},
  year={2023}
}

@article{lee2024read,
  title={Read like a radiologist: Efficient vision-language model for 3d medical imaging interpretation},
  author={Lee, Changsun and Park, Sangjoon and Shin, Cheong-Il and Choi, Woo Hee and Park, Hyun Jeong and Lee, Jeong Eun and Ye, Jong Chul},
  journal={arXiv preprint arXiv:2412.13558},
  year={2024}
}

@article{xin2025med3dvlm,
  title={Med3dvlm: An efficient vision-language model for 3d medical image analysis},
  author={Xin, Yu and Ates, Gorkem Can and Gong, Kuang and Shao, Wei},
  journal={arXiv preprint arXiv:2503.20047},
  year={2025}
}

@article{ates2025dcformer,
  title={Dcformer: Efficient 3d vision-language modeling with decomposed convolutions},
  author={Ates, Gorkem Can and Xin, Yu and Gong, Kuang and Shao, Wei},
  journal={arXiv preprint arXiv:2502.05091},
  year={2025}
}

@article{wang2024enhancing,
  title={Enhancing vision-language models for medical imaging: bridging the 3D gap with innovative slice selection},
  author={Wang, Yuli and Dai, Yuwei and Jones, Craig and Sair, Haris and Shen, Jinglai and Loizou, Nicolas and Hsu, Wen-Chi and Imami, Maliha and Jiao, Zhicheng and Zhang, Paul and others},
  journal={Advances in Neural Information Processing Systems},
  volume={37},
  pages={99947--99964},
  year={2024}
}

@article{liu2023t3d,
  title={T3d: Towards 3d medical image understanding through vision-language pre-training},
  author={Liu, Che and Ouyang, Cheng and Chen, Yinda and Quilodr{\'a}n-Casas, Cesar C{\'e}sar and Ma, Lei and Fu, Jie and Guo, Yike and Shah, Anand and Bai, Wenjia and Arcucci, Rossella},
  journal={arXiv preprint arXiv:2312.01529},
  year={2023}
}

@inproceedings{ma20253d,
  title={3d-rpe: Enhancing long-context modeling through 3d rotary position encoding},
  author={Ma, Xindian and Liu, Wenyuan and Zhang, Peng and Xu, Nan},
  booktitle={Proceedings of the AAAI Conference on Artificial Intelligence},
  volume={39},
  number={23},
  pages={24804--24811},
  year={2025}
}

@article{huemann2025vision,
  title={Vision-language modeling in PET/CT for visual grounding of positive findings},
  author={Huemann, Zachary and Church, Samuel and Warner, Joshua D and Tran, Daniel and Tie, Xin and McMillan, Alan B and Hu, Junjie and Cho, Steve Y and Lubner, Meghan and Bradshaw, Tyler J},
  journal={arXiv preprint arXiv:2502.00528},
  year={2025}
}

@inproceedings{chen2024bimcv,
  title={Bimcv-r: A landmark dataset for 3d ct text-image retrieval},
  author={Chen, Yinda and Liu, Che and Liu, Xiaoyu and Arcucci, Rossella and Xiong, Zhiwei},
  booktitle={International Conference on Medical Image Computing and Computer-Assisted Intervention},
  pages={124--134},
  year={2024},
  organization={Springer}
}

@article{lin2024ct,
  title={Ct-glip: 3d grounded language-image pretraining with ct scans and radiology reports for full-body scenarios},
  author={Lin, Jingyang and Xia, Yingda and Zhang, Jianpeng and Yan, Ke and Lu, Le and Luo, Jiebo and Zhang, Ling},
  journal={arXiv preprint arXiv:2404.15272},
  year={2024}
}

@article{xu2025cads,
  title={CADS: A Comprehensive Anatomical Dataset and Segmentation for Whole-Body Anatomy in Computed Tomography},
  author={Xu, Murong and Amiranashvili, Tamaz and Navarro, Fernando and Fritsak, Maksym and Hamamci, Ibrahim Ethem and Shit, Suprosanna and Wittmann, Bastian and Er, Sezgin and Christ, Sebastian M and de la Rosa, Ezequiel and others},
  journal={arXiv preprint arXiv:2507.22953},
  year={2025}
}

@inproceedings{lin2014microsoft,
  title={Microsoft coco: Common objects in context},
  author={Lin, Tsung-Yi and Maire, Michael and Belongie, Serge and Hays, James and Perona, Pietro and Ramanan, Deva and Doll{\'a}r, Piotr and Zitnick, C Lawrence},
  booktitle={European conference on computer vision},
  pages={740--755},
  year={2014},
  organization={Springer}
}
}

\clearpage
\setcounter{page}{1}
\appendix
\maketitlesupplementary


\section{Muon Optimizer}

We use a single optimizer that applies \textbf{Muon} to matrix parameters with two or more dimensions and defaults to AdamW for everything else (\emph{e.g.,} biases, gains, embeddings). In this paper, we directly use the optimizer provided in the GitHub repository from \cite{jordan2024muon}. 
\textbf{Muon update:}
For parameter $\mathbf{w}$ with gradient $\mathbf{g}_t$, step size $\eta$, decay $\lambda$, and momentum $\beta$,
\begin{align}
\mathbf{u}_t &= \beta\,\mathbf{u}_{t-1} + (1-\beta)\,\mathbf{g}_t, \\
\mathbf{w} &\leftarrow (1-\eta\lambda)\,\mathbf{w}\;-\;\eta\,\mathbf{u}_t.
\end{align}
This is a non-adaptive, momentum-only update (no second moment or variance normalization), which we found preferable for large weight matrices. Other parameters use AdamW.

\begin{figure}[t]
    \centering
    \includegraphics[width=1.0\linewidth]{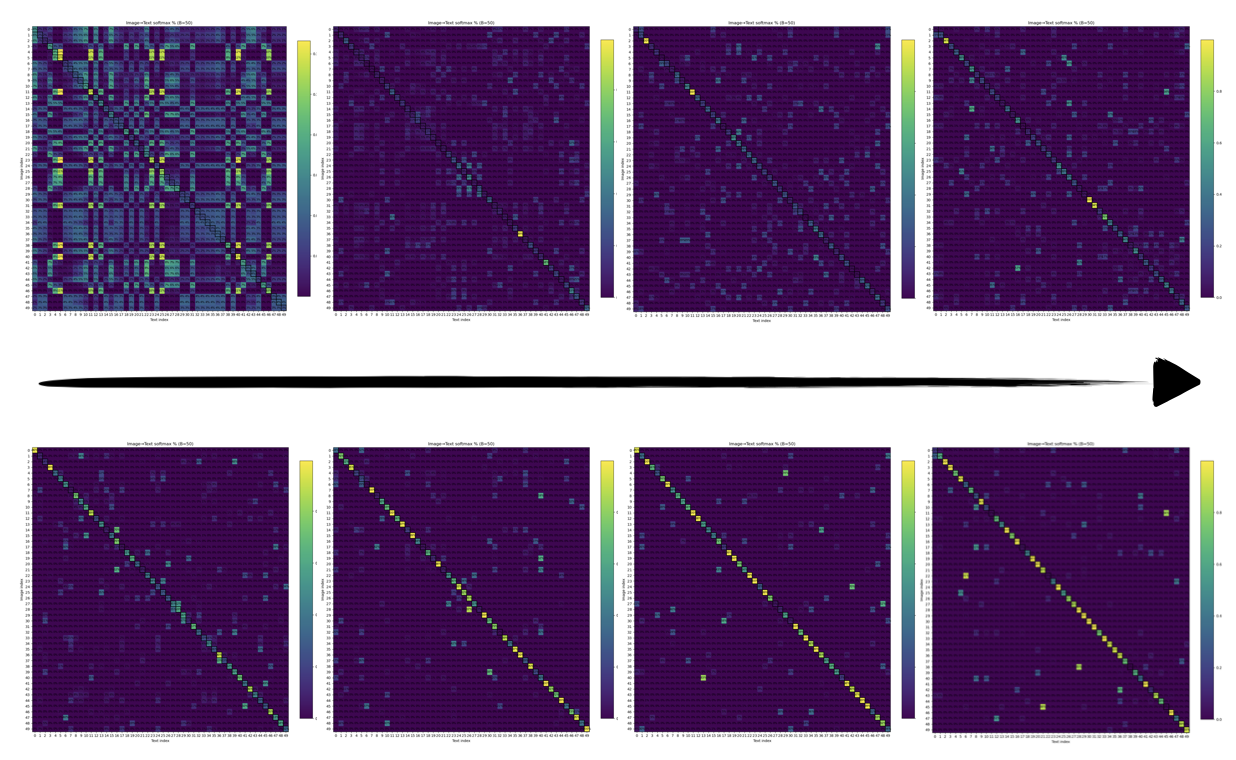}
    \caption{Heatmap Grids of Image-to-Text Retrieval (50-Val Set; Every 500 Steps): Adam in the Upper Row, Muon in the Lower Row}
    \label{fig:adam_vs_muon}
\end{figure}

\begin{figure}[t]
    \centering
    \includegraphics[width=1.0\linewidth]{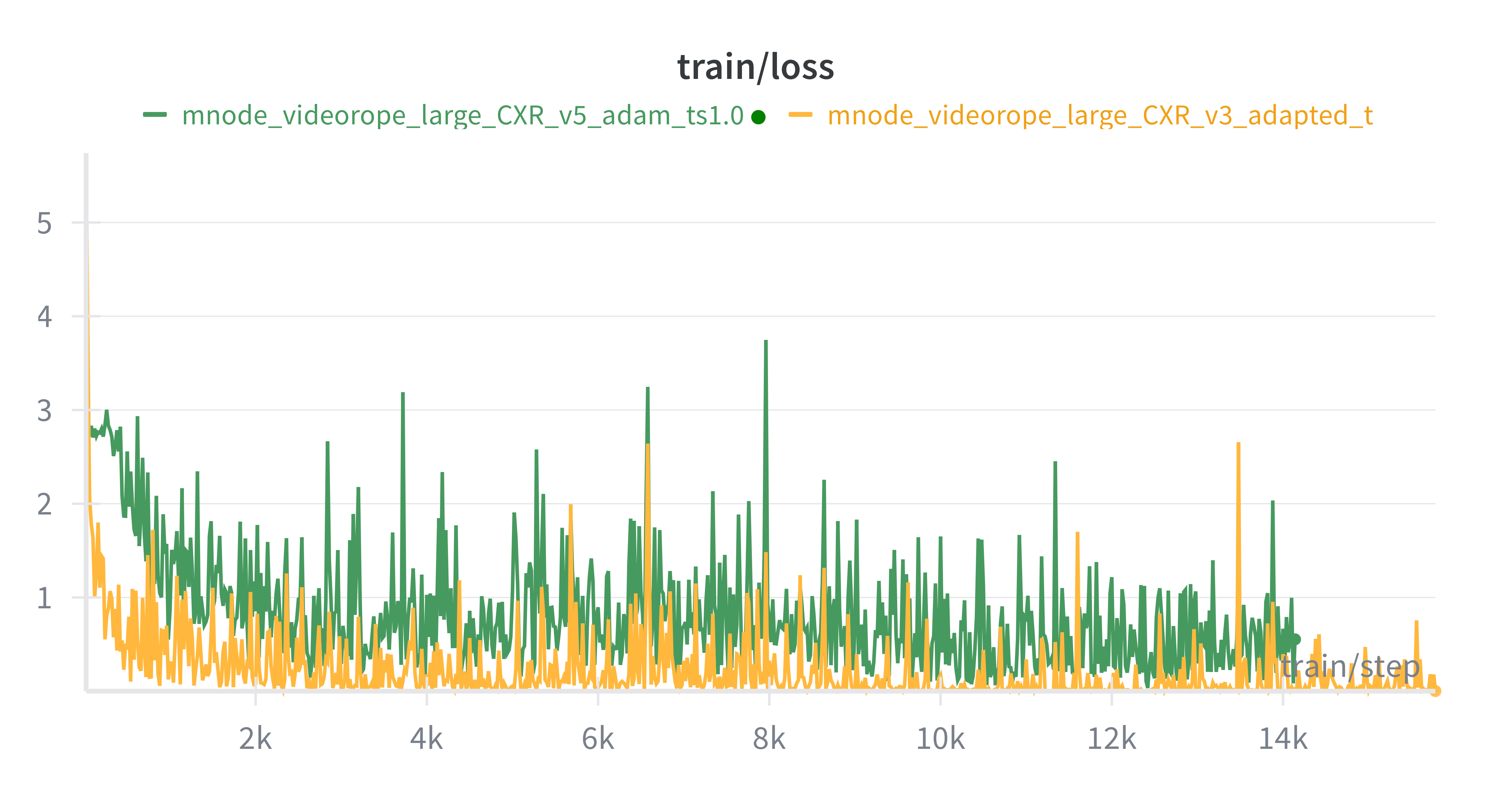}
    \caption{Training Loss vs. Steps — Adam (green) vs. Muon (yellow), Identical Model Configuration}
    \label{fig:loss_function}
\end{figure}

Across both visualizations, Muon consistently outperforms AdamW.
In Figure \ref{fig:adam_vs_muon}, the Muon training curve exhibits lower variance and converges to a more stable loss than AdamW, indicating smoother optimization and reduced gradient noise.
Figure \ref{fig:loss_function} further confirms this trend as the retrieval heatmaps under Muon become cleaner and more near-diagonal, while AdamW produces noisier, less structured similarity patterns.
Overall, Muon provides improved stability and alignment quality compared with pure AdamW under identical model settings.

\section{Feature Visualization}

\begin{figure*}[htbp]
    \centering
    \begin{subfigure}{0.24\textwidth}
        \includegraphics[width=\linewidth]{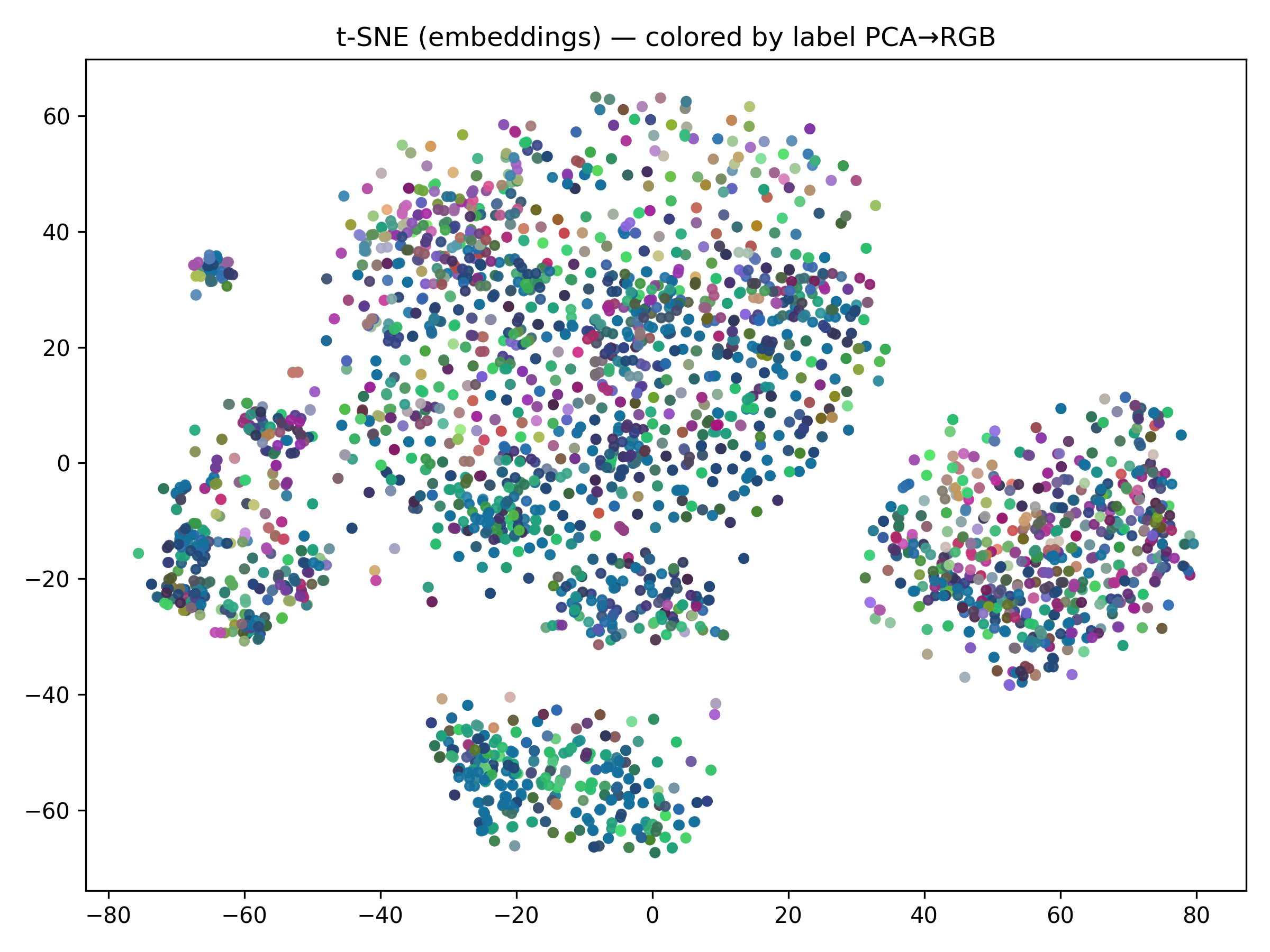}
        \caption{DINOV3-base}
    \end{subfigure}
    \begin{subfigure}{0.24\textwidth}
        \includegraphics[width=\linewidth]{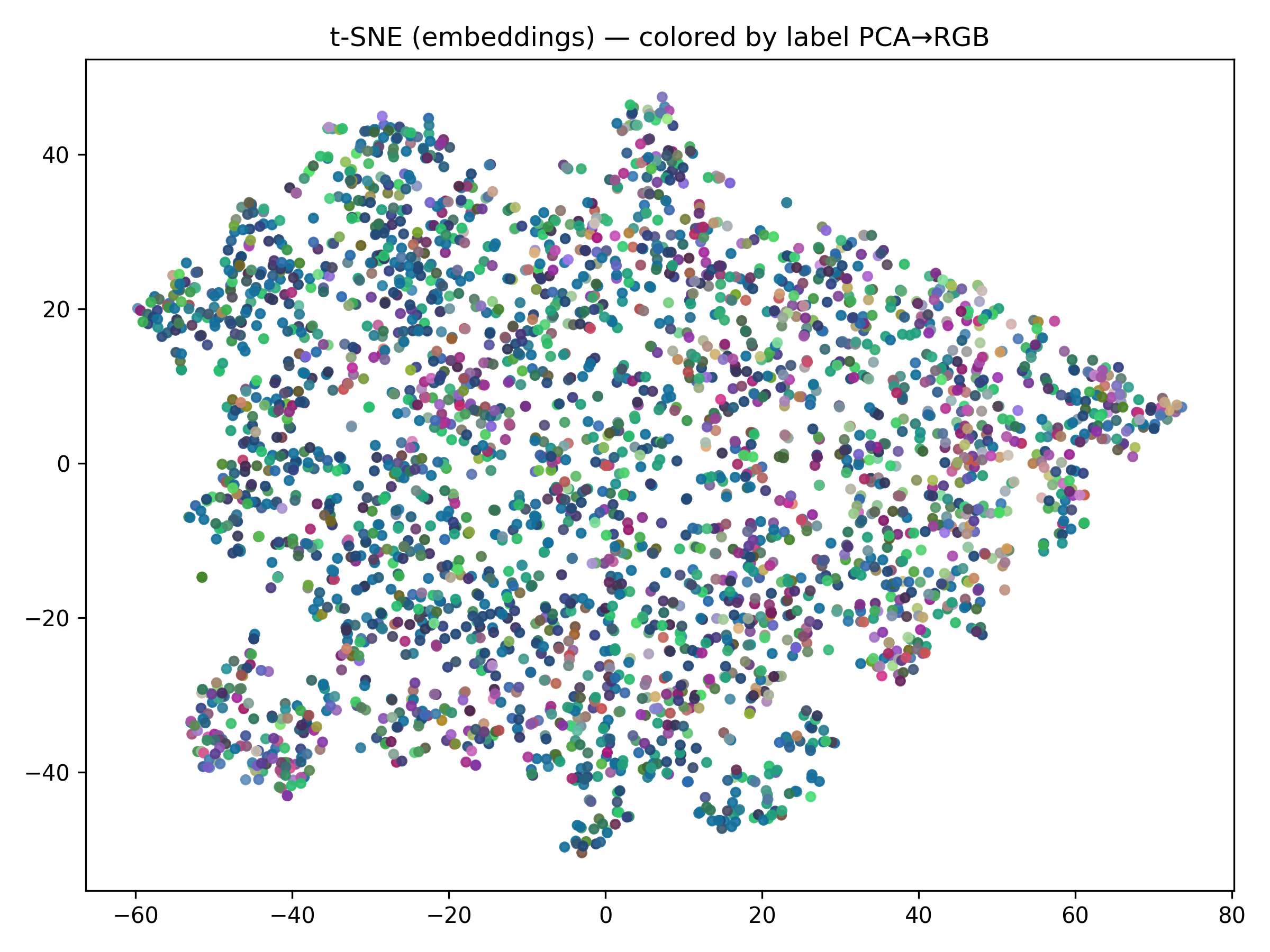}
        \caption{CT-CLIP}
    \end{subfigure}
    \begin{subfigure}{0.24\textwidth}
        \includegraphics[width=\linewidth]{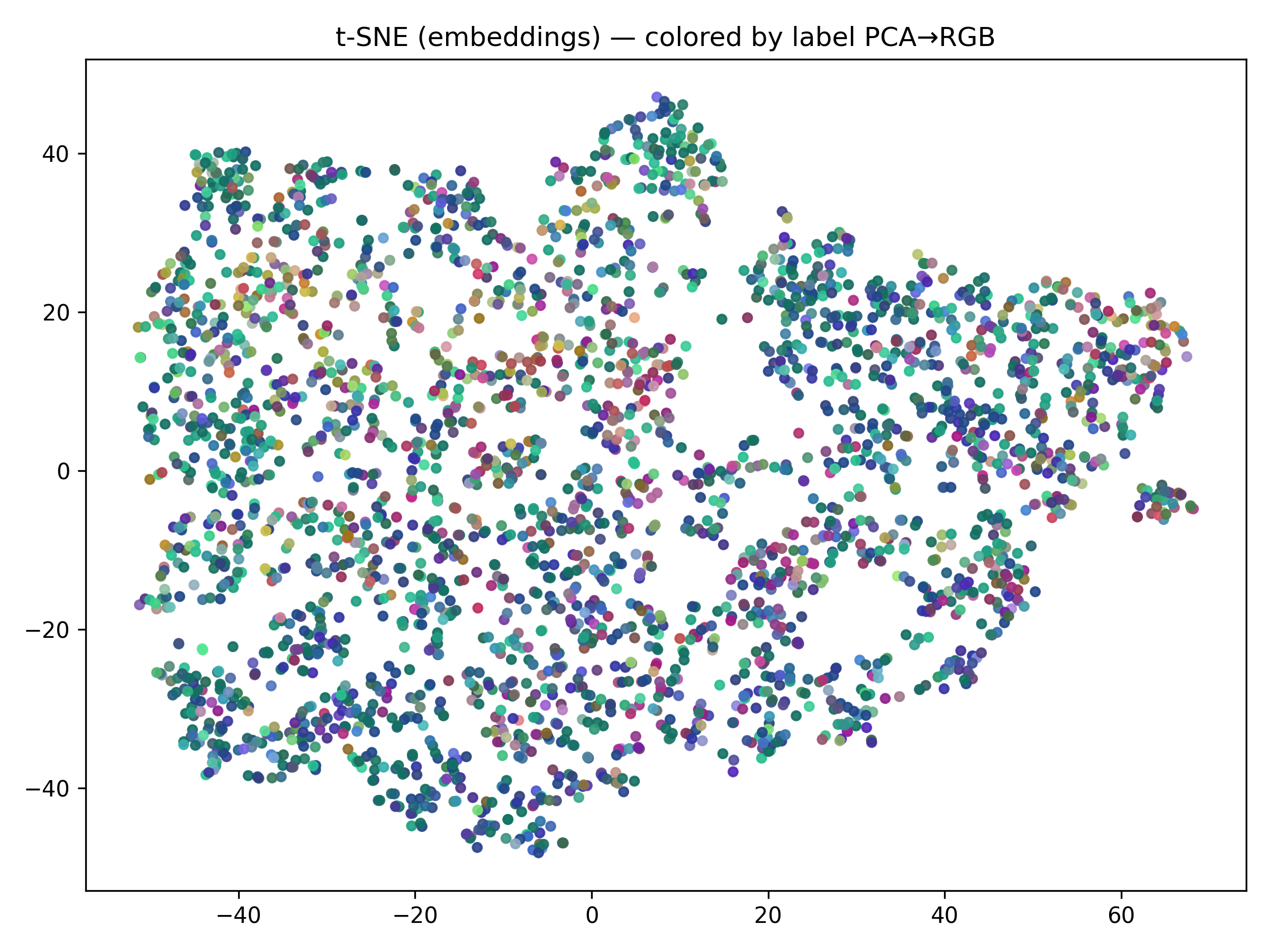}
        \caption{CT-Vocab (vocabulary-finetuned)}
    \end{subfigure}
    \begin{subfigure}{0.24\textwidth}
        \includegraphics[width=\linewidth]{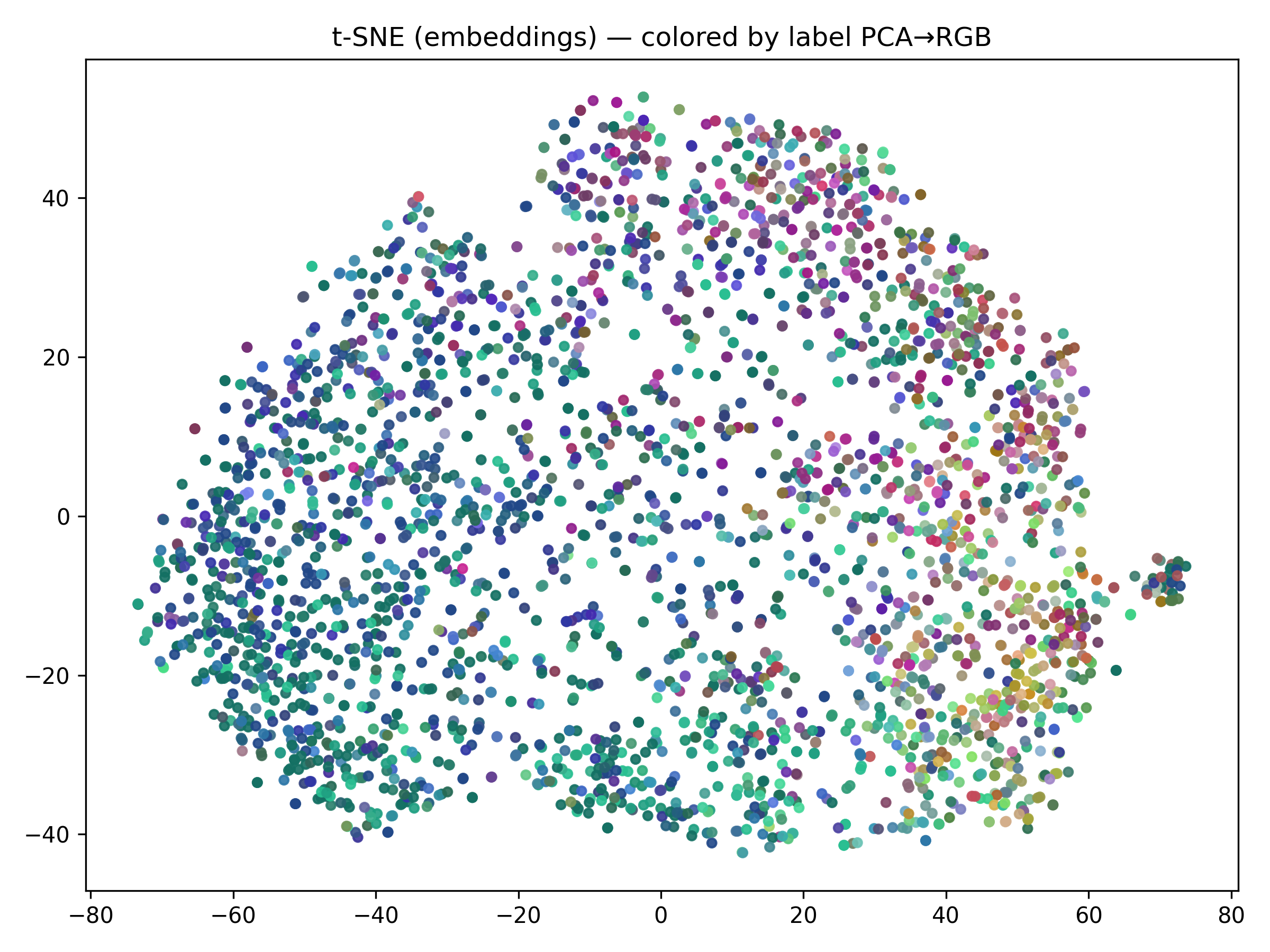}
        \caption{CT-LiPro (classification-tuned)}
    \end{subfigure}

    \begin{subfigure}{0.24\textwidth}
        \includegraphics[width=\linewidth]{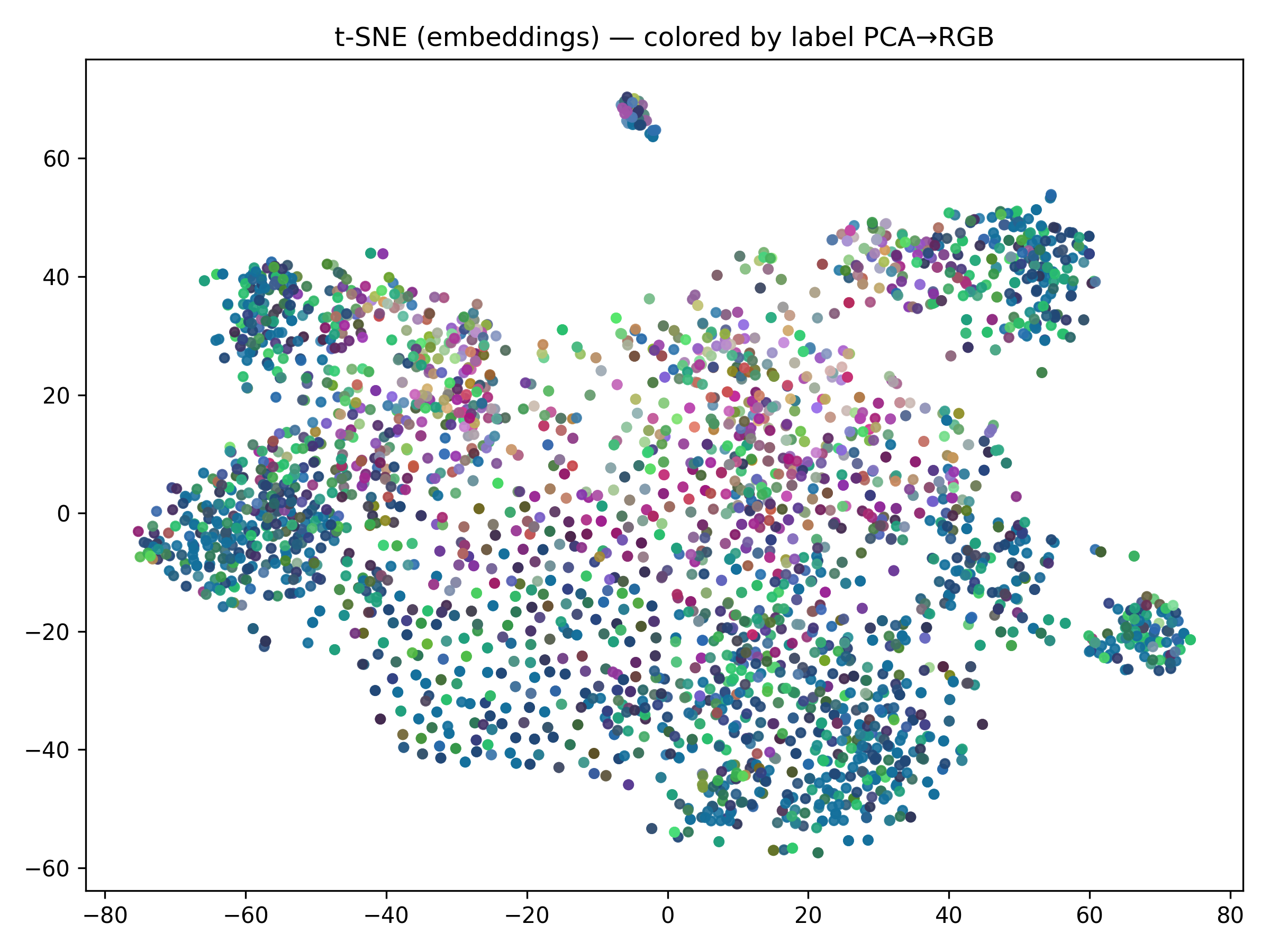}
        \caption{Ours (2k steps)}
    \end{subfigure}
    \begin{subfigure}{0.24\textwidth}
        \includegraphics[width=\linewidth]{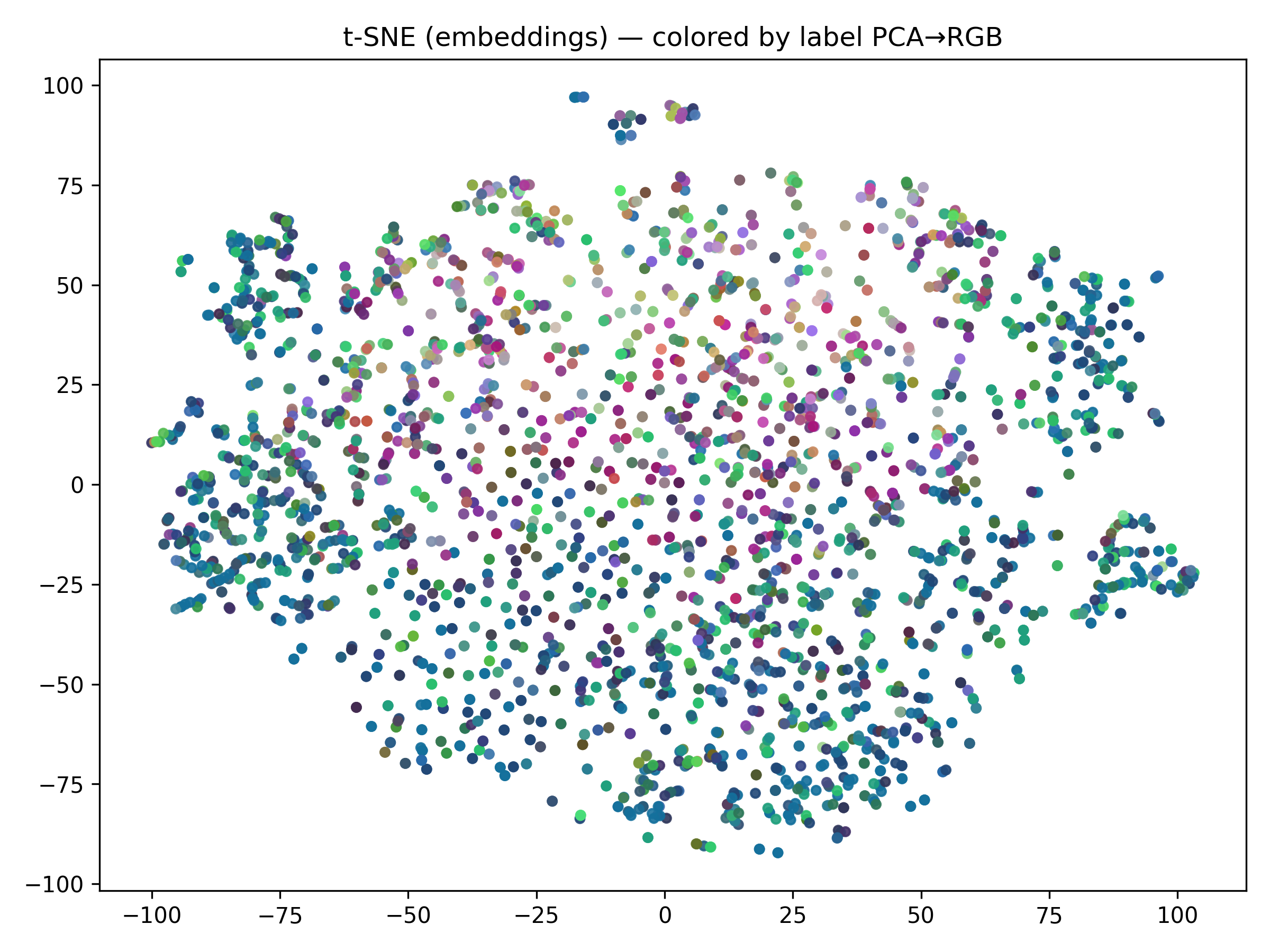}
        \caption{Ours (4k steps)}
    \end{subfigure}
    \begin{subfigure}{0.24\textwidth}
        \includegraphics[width=\linewidth]{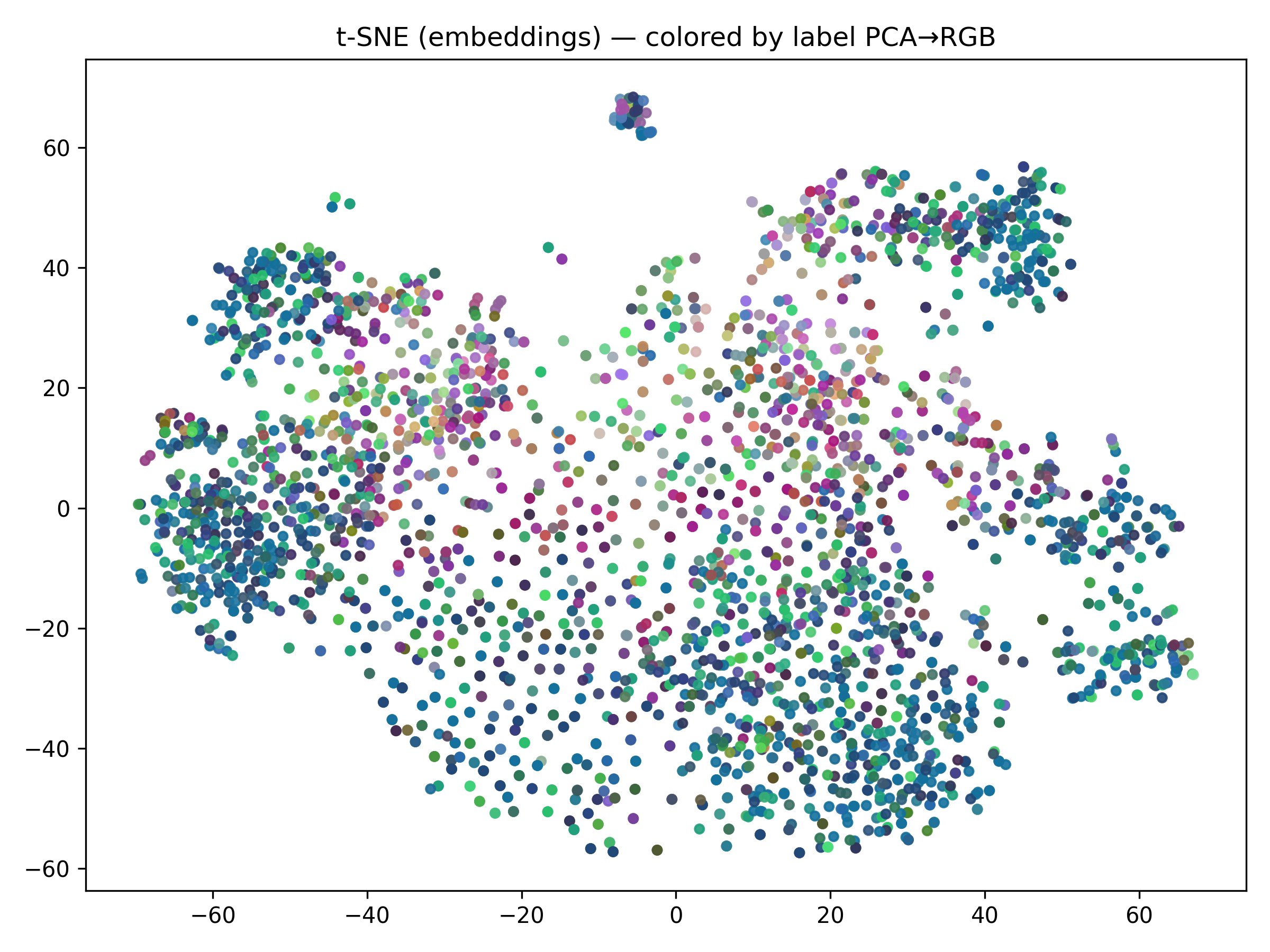}
        \caption{Ours (6k steps)}
    \end{subfigure}
    \begin{subfigure}{0.24\textwidth}
        \includegraphics[width=\linewidth]{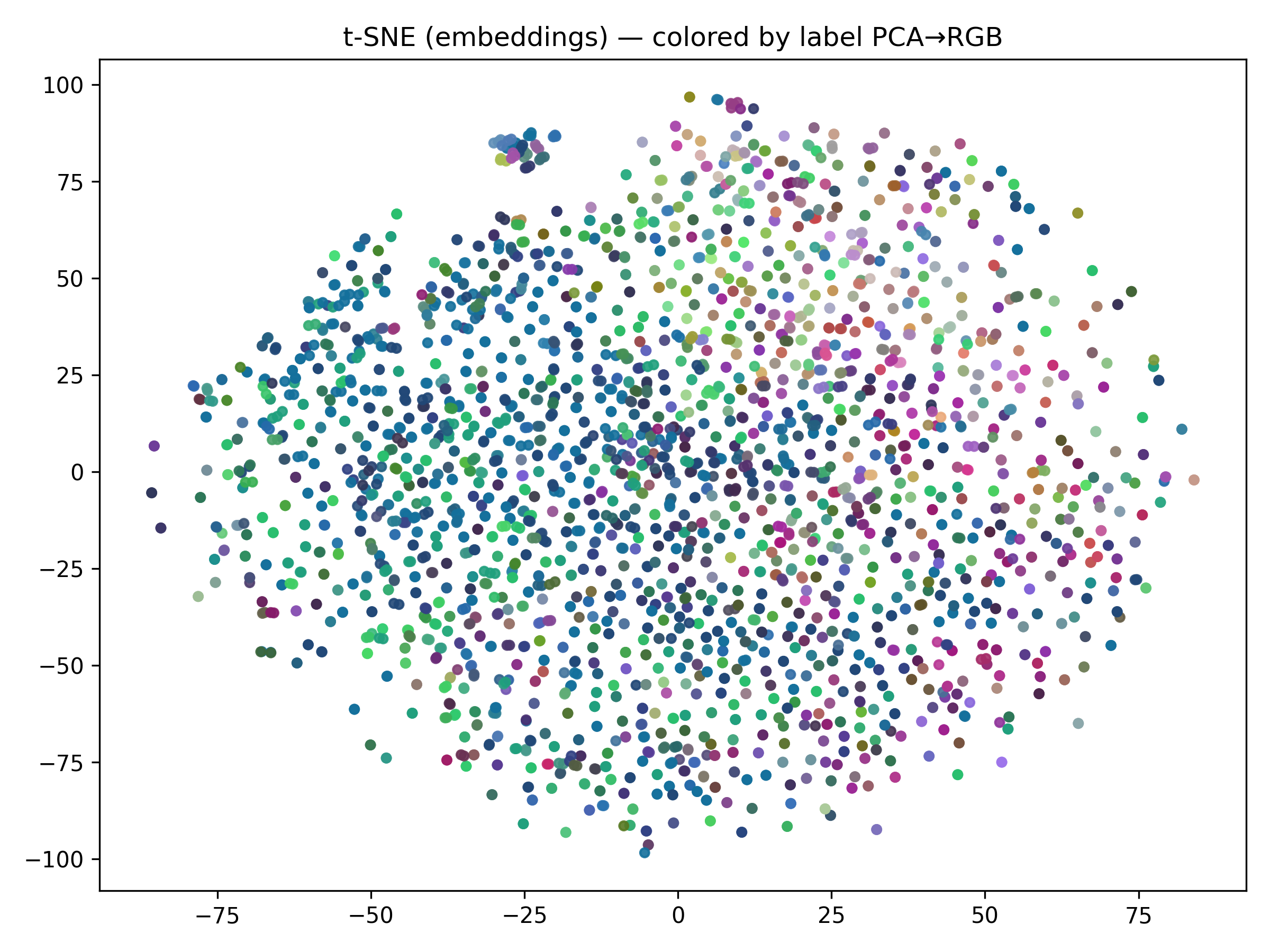}
        \caption{Ours (234{,}930 steps)}
    \end{subfigure}

    \caption{\textbf{t-SNE of evaluation embeddings} across baselines and our model. Colors denote abnormality classes (similar hues indicate semantically related labels). \textit{Top row:} DINOv3-base, CT-CLIP, CT-Vocab (vocabulary-finetuned), CT-LiPro (classification-finetuned). \textit{Bottom row:} Ours at 2k, 4k, 6k, and 234{,}930 training steps.}
    \label{fig:tNSE}
\end{figure*}

In Section~\ref{sec:results} Figure~\ref{fig:big_4x2}, we present UMAP feature visualizations. For completeness, we also evaluate t-SNE projections across different classes for the same embeddings. The results remain consistent with those in Figure~\ref{fig:tNSE}: the clusters produced by DINOv3 appear disorganized and randomly clustered, while those from CT-RATE and CT-Vocab exhibit mixed boundaries. In contrast, our model yields smoother and more coherent cluster distribution, showing behavior closely aligned with the class fine-tuned CT-LiPro model.

\section{Computational Cost and Throughput Comparison}

We report the computational efficiency of our model under different temporal lengths $T$ and compare it against DINOv3. 
For each setting, we measure the FLOP per sample, average latency, and effective FLOPs computed as 
$\text{FLOP}/(\text{latency in seconds})$. 
Note that for our model, $T=1$ shares the same cost as $T=16$ due to padding. 
Results are summarized in Table~\ref{tab:flop_throughput}.

Our model demonstrates consistently higher computational efficiency compared with DINOv3.
As shown in Table~\ref{tab:flop_throughput}, for all temporal lengths $T \ge 32$, our method requires substantially fewer FLOP per sample while achieving comparable or higher throughput.
Combined with the accuracy curves in section~\ref{sec:ablatioon} Fig.~\ref{fig:F1-curve}, it indicates that our model attains better performance using significantly lower computational resources than DINOv3, except when the slice count is extremely small ($T < 16$).

\begin{table*}[t]
\centering
\scriptsize
\caption{Comparison of computational cost and throughput for Ours vs. DINOv3 at different slices number. FLOP/sample corresponds to one forward pass per sample.}
\setlength{\tabcolsep}{6pt}
\begin{tabular}{c|ccc|ccc}
\toprule
& \multicolumn{3}{c|}{\textbf{Ours}} & \multicolumn{3}{c}{\textbf{DINOV3}} \\
\textbf{Slices Count} & \textbf{FLOP/sample} & \textbf{Latency (ms)} & \textbf{FLOP/s} & 
\textbf{FLOP/sample} & \textbf{Latency (ms)} & \textbf{FLOP/s} \\
\midrule
256 & 2.080 TFLOP & 147.692 & 14.09 & 5.720 TFLOP & 137.422 & 41.62 \\
128 & 0.834 TFLOP & 53.896 & 15.47 & 2.860 TFLOP & 69.674 & 41.04 \\
64  & 0.365 TFLOP & 24.588 & 14.86 & 1.430 TFLOP & 36.035 & 39.68 \\
32  & 0.170 TFLOP & 23.010 & 7.38  & 0.715 TFLOP & 18.440 & 38.78 \\
16  & 0.082 TFLOP & 22.758 & 3.59  & 0.358 TFLOP & 9.423  & 37.93 \\
1   & 0.082 TFLOP & 22.758 & 3.59  & 0.022 TFLOP & 12.555 & 1.78  \\
\bottomrule
\end{tabular}
\label{tab:flop_throughput}
\end{table*}

\section{Comparison of Padding Strategies}

In Section~\ref{sec:ablatioon}, we analyze the effect of input slice count variation. 
Because our patch size along the z-axis is fixed to 16, padding is required when evaluating cases with only 1 or 8 slices. 
We compared ``repeat padding'' and ``zero padding'', and ultimately adopted repeat padding because our quantitative results, presented in Table~\ref{tab:pad}, showed improved performance. 
In particular, when \textit{slice = 1}, repeat padding yields a substantial improvement, with precision increasing from 0.830 to 0.931.

\begin{figure*}[t]
    \centering

    \begin{subfigure}{0.9\linewidth}
        \centering
        \includegraphics[width=\linewidth]{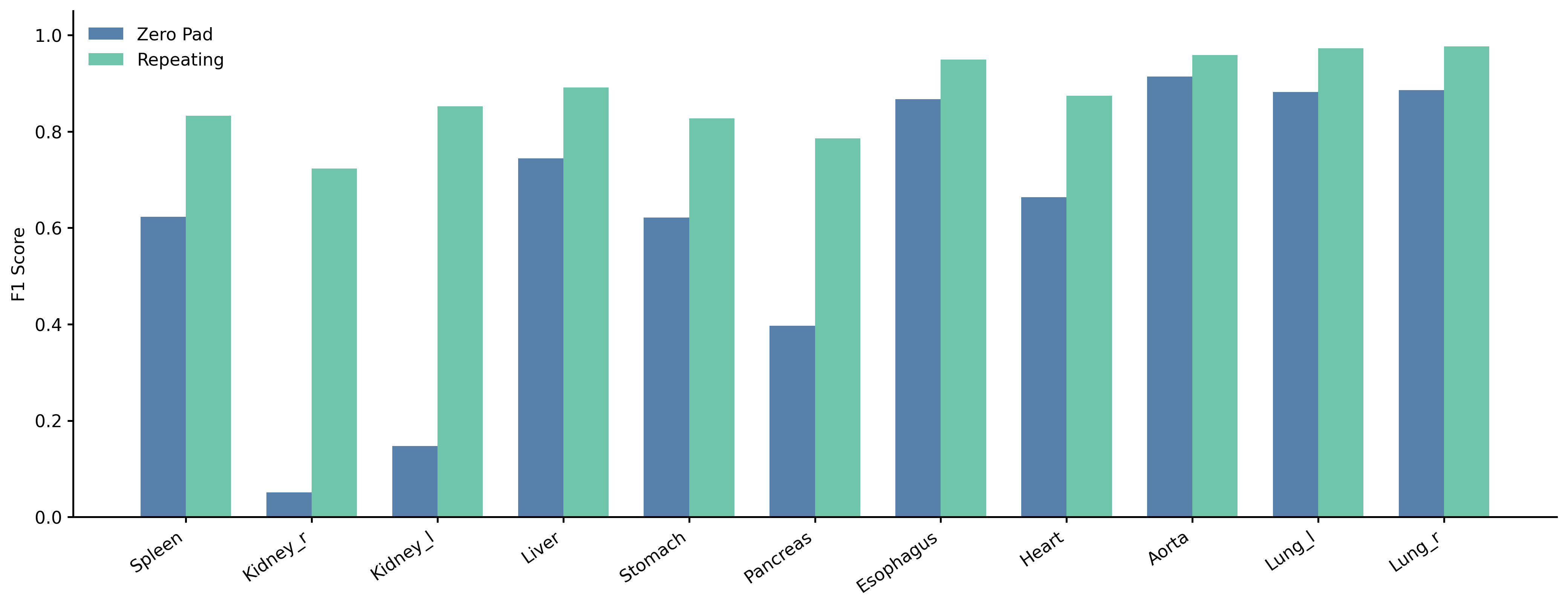}
        \caption{Per-organ F1 for Slice = 1}
        \label{fig:pad_slice1}
    \end{subfigure}

    \vspace{0.4cm} 

    \begin{subfigure}{0.9\linewidth}
        \centering
        \includegraphics[width=\linewidth]{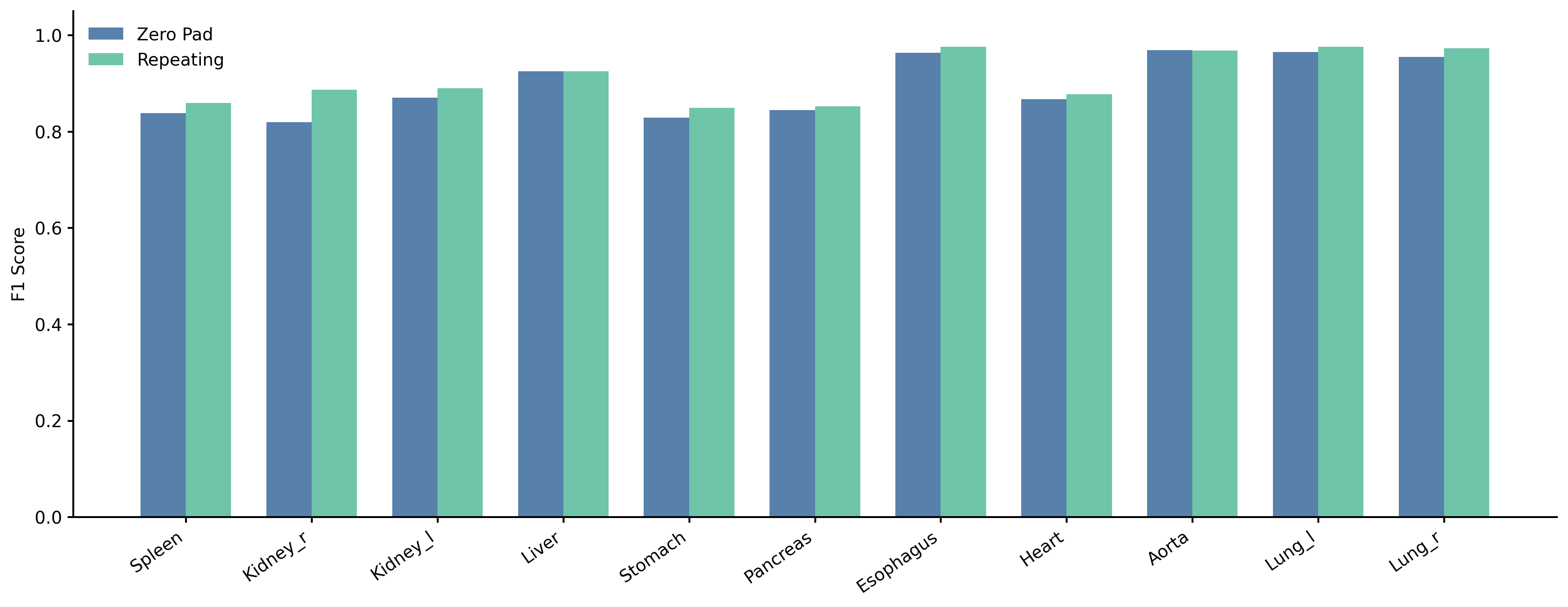}
        \caption{Per-organ F1 for Slice = 8}
        \label{fig:pad_slice8}
    \end{subfigure}

    \caption{Comparison of Zero Padding and Repeat Padding across organs under different slice counts.}
    \label{fig:pad_vertical}
\end{figure*}

More detailed per-organ comparisons are visualized in Fig.~\ref{fig:pad_vertical}. 
When \textit{slice = 1}, repeat padding significantly outperforms zero padding across almost all organs. 
Zero padding performs especially poorly on small or thin structures such as 
\textit{Right Kidney} (0.051~$\rightarrow$~0.723), 
\textit{Left Kidney} (0.147~$\rightarrow$~0.852), 
and \textit{Pancreas} (0.397~$\rightarrow$~0.786), 
while large organs such as 
\textit{Spleen} (0.623~$\rightarrow$~0.833) and 
\textit{Liver} (0.744~$\rightarrow$~0.891) also benefit noticeably. 
Overall, repeat padding consistently achieves higher F1 scores, with the largest gains observed on low-contrast or shape-sensitive organs.

When \textit{slice = 8}, the performance gap narrows, but the trend remains unchanged: repeat padding achieves the best results for nearly every organ, e.g., 
\textit{Right Kidney} (0.819~$\rightarrow$~0.887), 
\textit{Pancreas} (0.844~$\rightarrow$~0.852), and 
\textit{Left Lung} (0.965~$\rightarrow$~0.976). 
Although the improvements are smaller as there is more contextual information available, repeat padding still provides more stable and accurate predictions overall.

These results demonstrate that our model, which was not trained on empty slices, is sensitive to the content of each individual input slice.
Motivated by this conclusion, all downstream components that involve padding likewise employ the repeat strategy to provide as much contextual information as possible to the model.

\begin{table}[t]
\centering
\scriptsize
\caption{Comparison of padding strategies for different input slice counts.}
\begin{tabular}{c|c|ccccc}
\toprule
\textbf{Slices Num.} & \textbf{Pad method} & \textbf{Precision} & \textbf{Recall} & \textbf{F1} & \textbf{AUC} & \textbf{mAP} \\
\midrule
1 & Repeat   & 0.931 & 0.906 & 0.918 & 0.986 & 0.936 \\
1 & Zero Pad & 0.830 & 0.750 & 0.788 & 0.922 & 0.778 \\
\midrule
8 & Repeat   & 0.951 & 0.925 & 0.938 & 0.990 & 0.960 \\
8 & Zero Pad & 0.930 & 0.919 & 0.925 & 0.986 & 0.943 \\
\bottomrule
\end{tabular}
\label{tab:pad}
\end{table}

\section{1500 Volumes Report Retrieval}

\begin{table*}[t]
\centering
\caption{\textbf{1500 validation subset whole-body retrieval across different CLIP-based methods. For our proposed SigVLP
method, we employ reconstructed reports generated through our
pipeline.} 
Retrieval is performed between a 200-slice CT volume and the full radiology report.}
\label{tab:1500_retrieval}
\scriptsize
\begin{tabular}{lccccccc}
\toprule
\textbf{Model} & \textbf{R@1} & \textbf{R@5} & \textbf{R@10} & \textbf{R@50} & \textbf{R@100} & \textbf{MeanRank} & \textbf{mAP} \\
\midrule
SigLIPv2-L\cite{tschannen2025siglip}        & 0.002 & 0.003 & 0.007 & 0.033 & 0.065 & 751.77 & 0.006 \\
CT-CLIP\cite{hamamci2024developing}        & 0.007 & 0.029 & 0.045 & 0.161 & 0.263 & 378.39 & 0.024 \\
CT-Vocab\cite{hamamci2024developing}       & 0.001 & 0.004 & 0.007 & 0.035 & 0.071 & 729.09 & 0.006 \\
DINOv3-CLIP\cite{duriantaco2025dinov3clip}    & 0.001 & 0.003 & 0.008 & 0.030 & 0.064 & 751.70 & 0.006 \\
\textbf{Ours}  & \textbf{0.098} & \textbf{0.251} & \textbf{0.336} & \textbf{0.599} & \textbf{0.730} & \textbf{107.13} & \textbf{0.180} \\
\bottomrule
\end{tabular}
\end{table*}

\begin{figure*}[t]
    \centering

    \begin{subfigure}{0.32\textwidth}
        \centering
        \includegraphics[width=\linewidth]{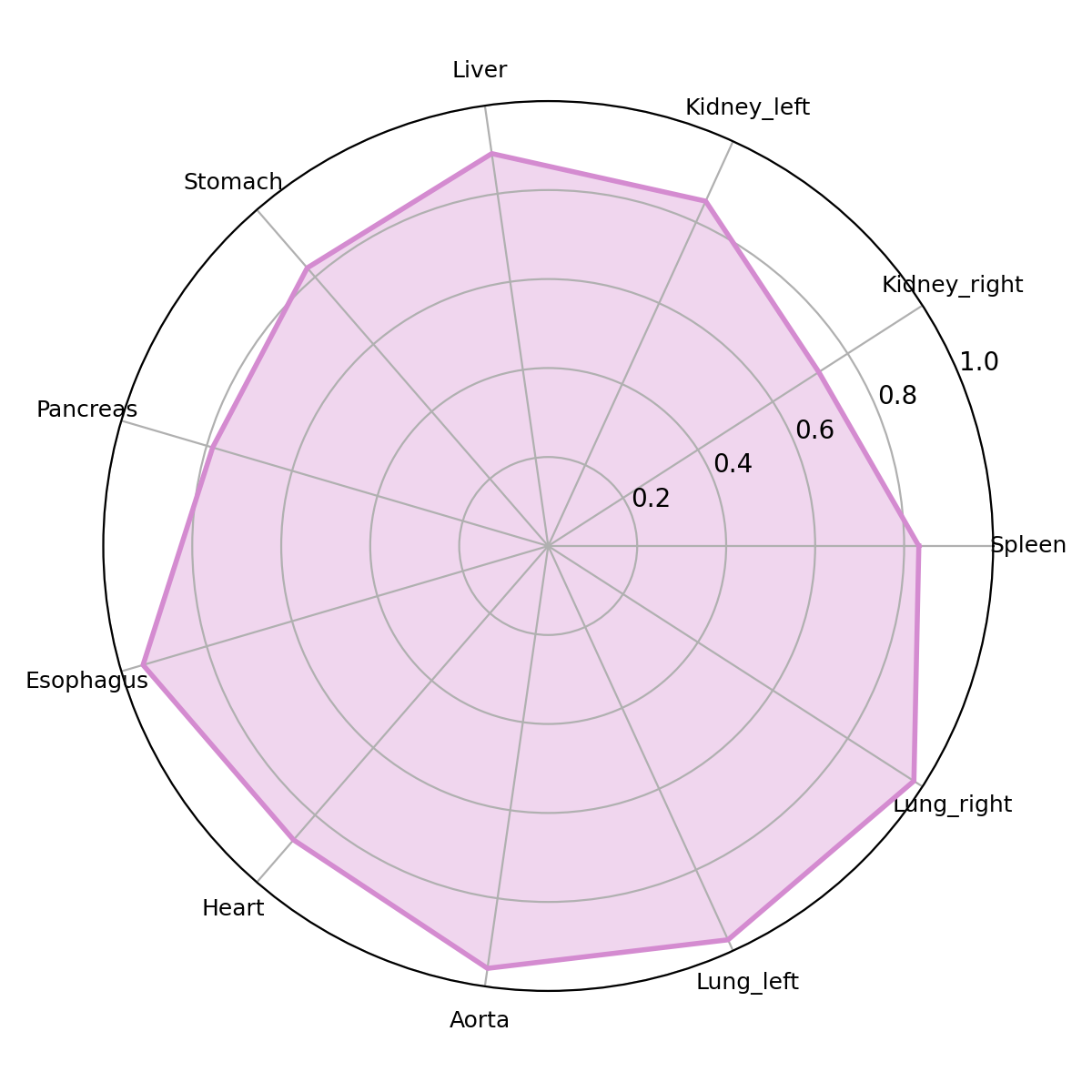}
        \caption{Slice = 1}
        \label{fig:radar_1}
    \end{subfigure}
    \begin{subfigure}{0.32\textwidth}
        \centering
        \includegraphics[width=\linewidth]{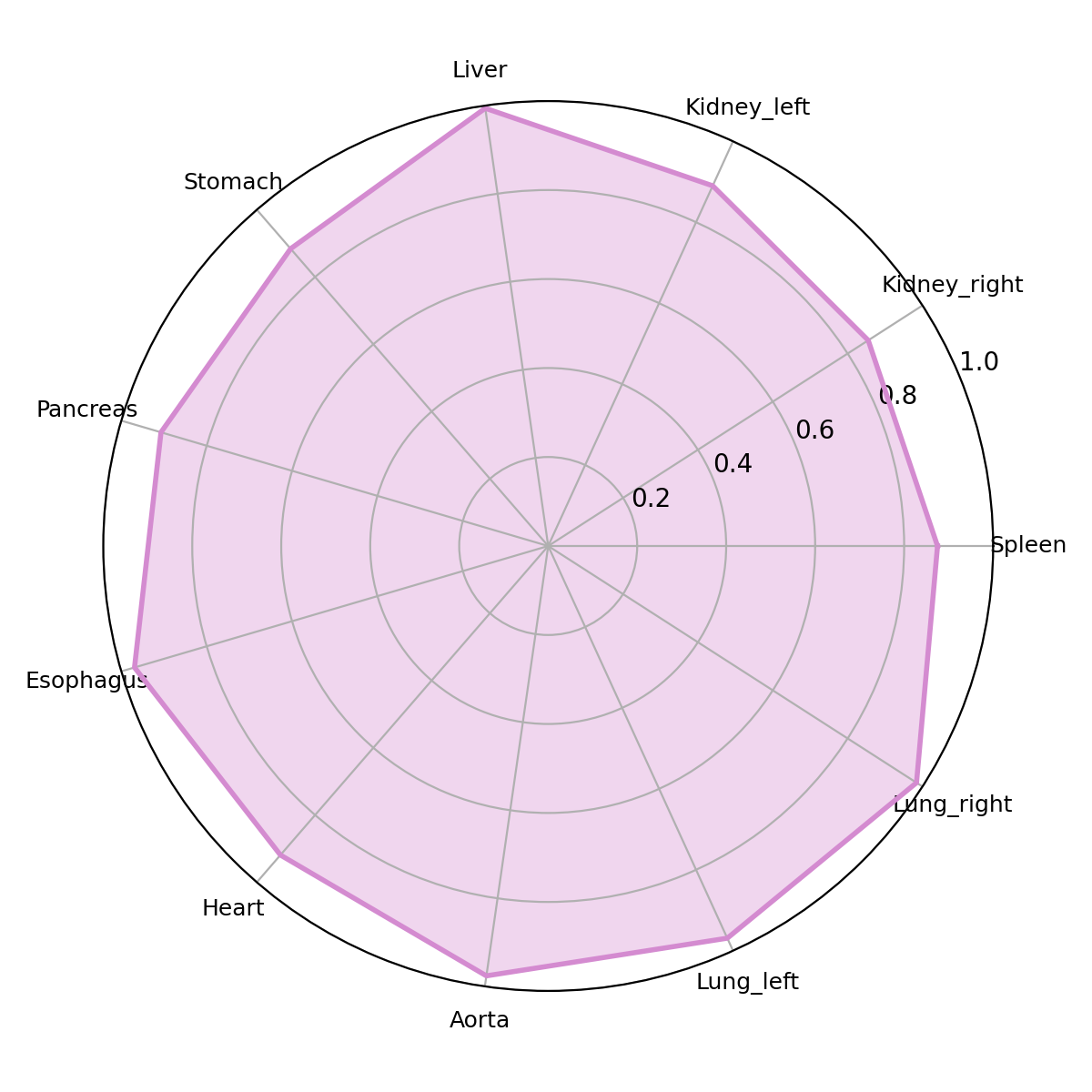}
        \caption{Slice = 16}
        \label{fig:radar_16}
    \end{subfigure}
    \begin{subfigure}{0.32\textwidth}
        \centering
        \includegraphics[width=\linewidth]{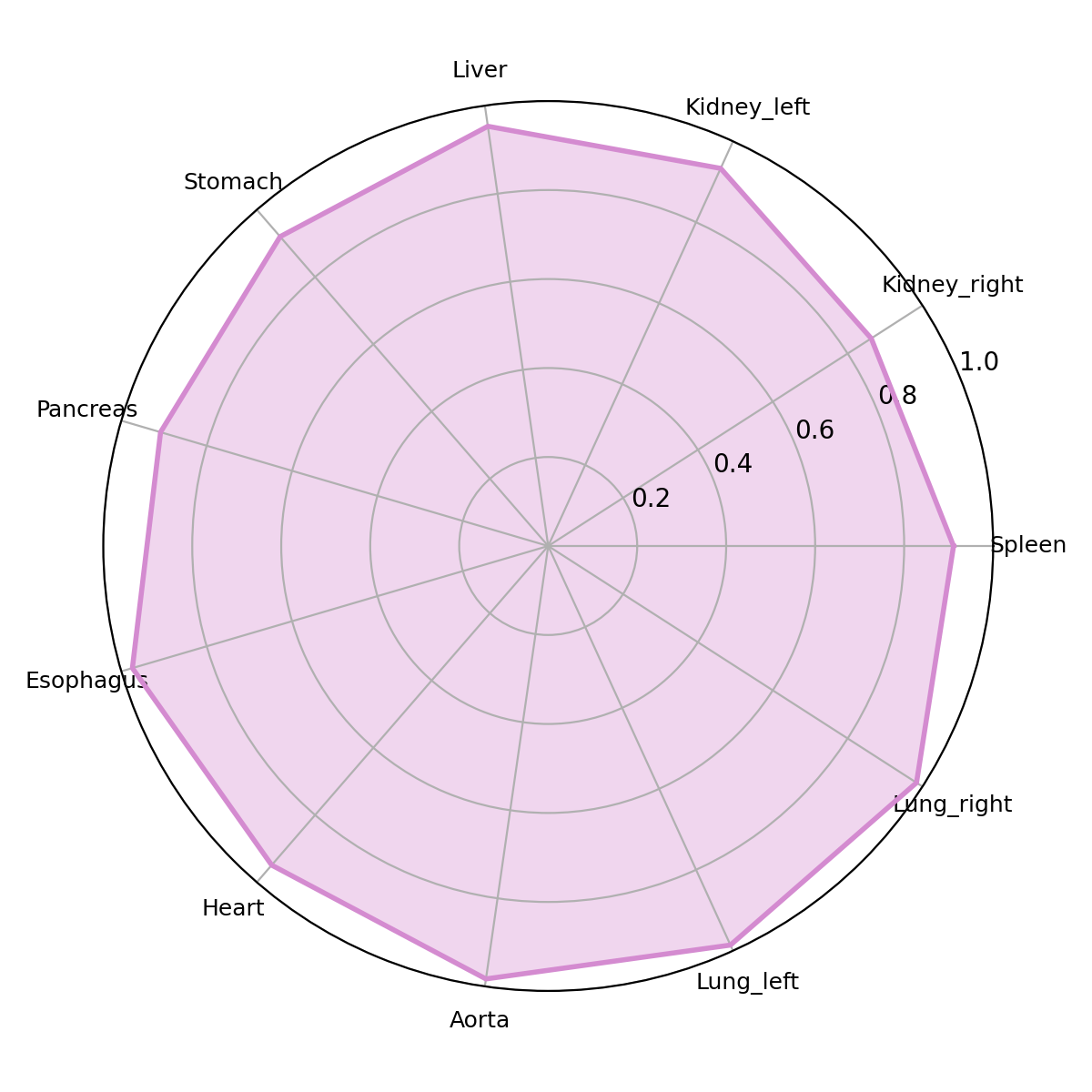}
        \caption{Slice = 32}
        \label{fig:radar_32}
    \end{subfigure}

    \vspace{8pt}

    \begin{subfigure}{0.32\textwidth}
        \centering
        \includegraphics[width=\linewidth]{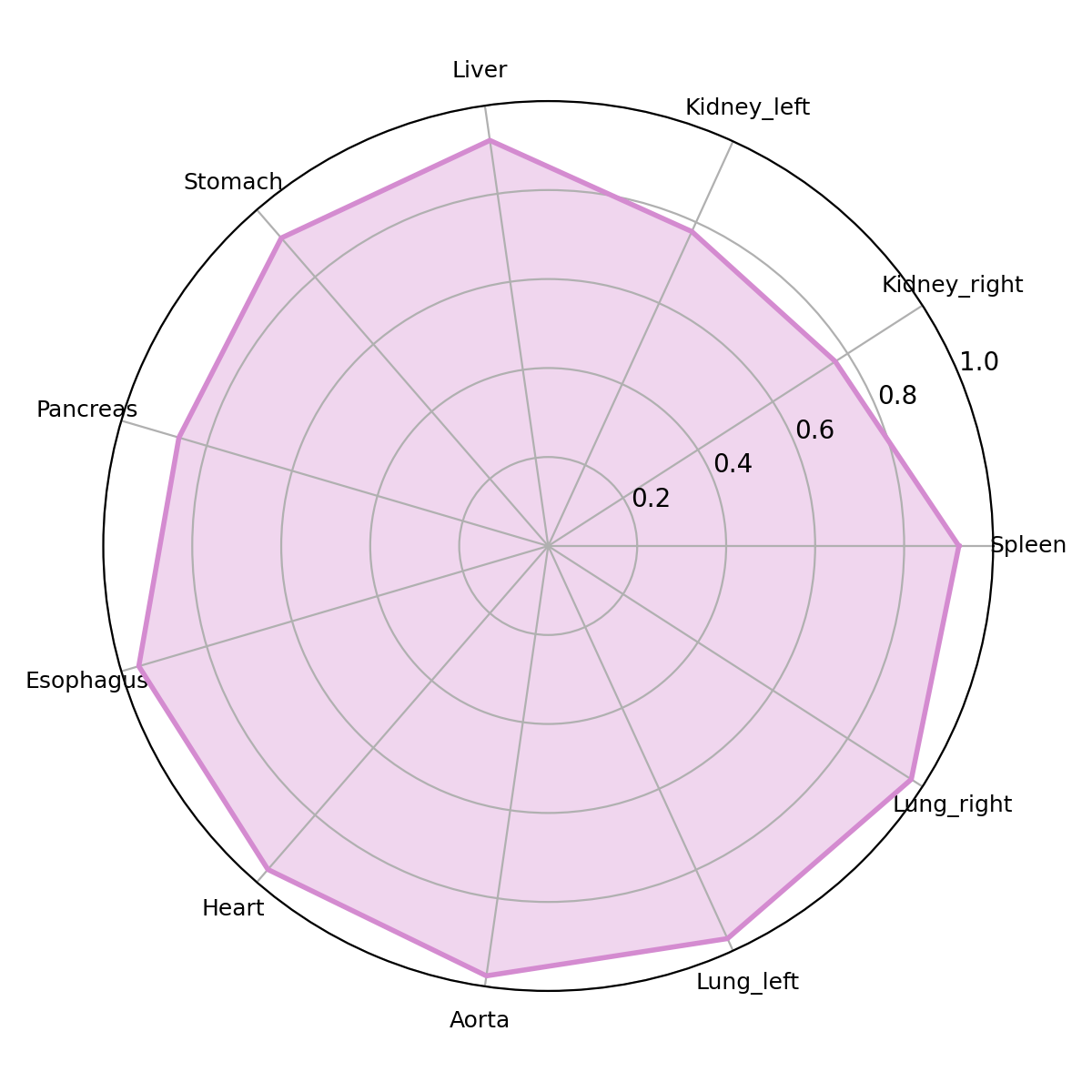}
        \caption{Slice = 64}
        \label{fig:radar_64}
    \end{subfigure}
    \begin{subfigure}{0.32\textwidth}
        \centering
        \includegraphics[width=\linewidth]{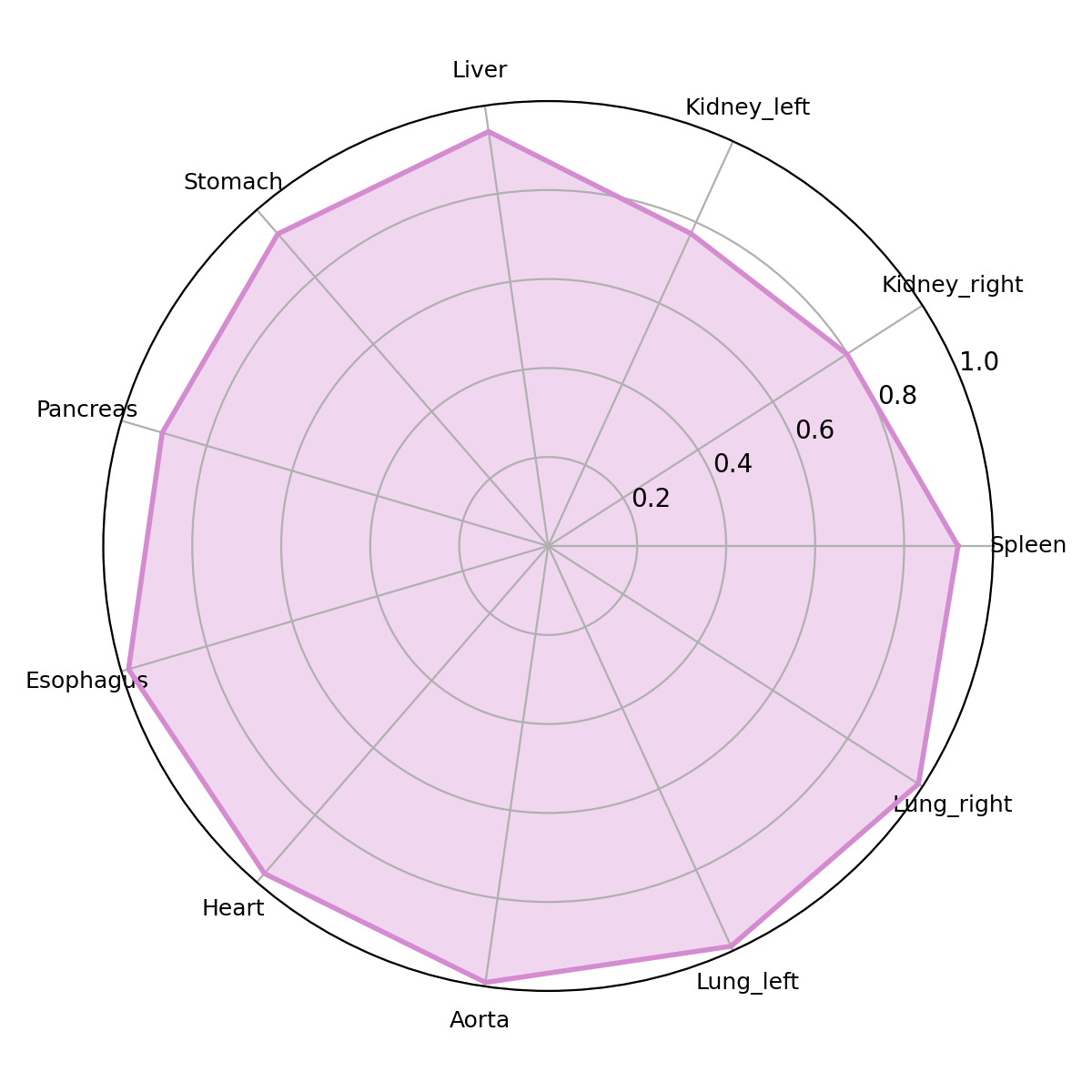}
        \caption{Slice = 128}
        \label{fig:radar_128}
    \end{subfigure}
    \begin{subfigure}{0.32\textwidth}
        \centering
        \includegraphics[width=\linewidth]{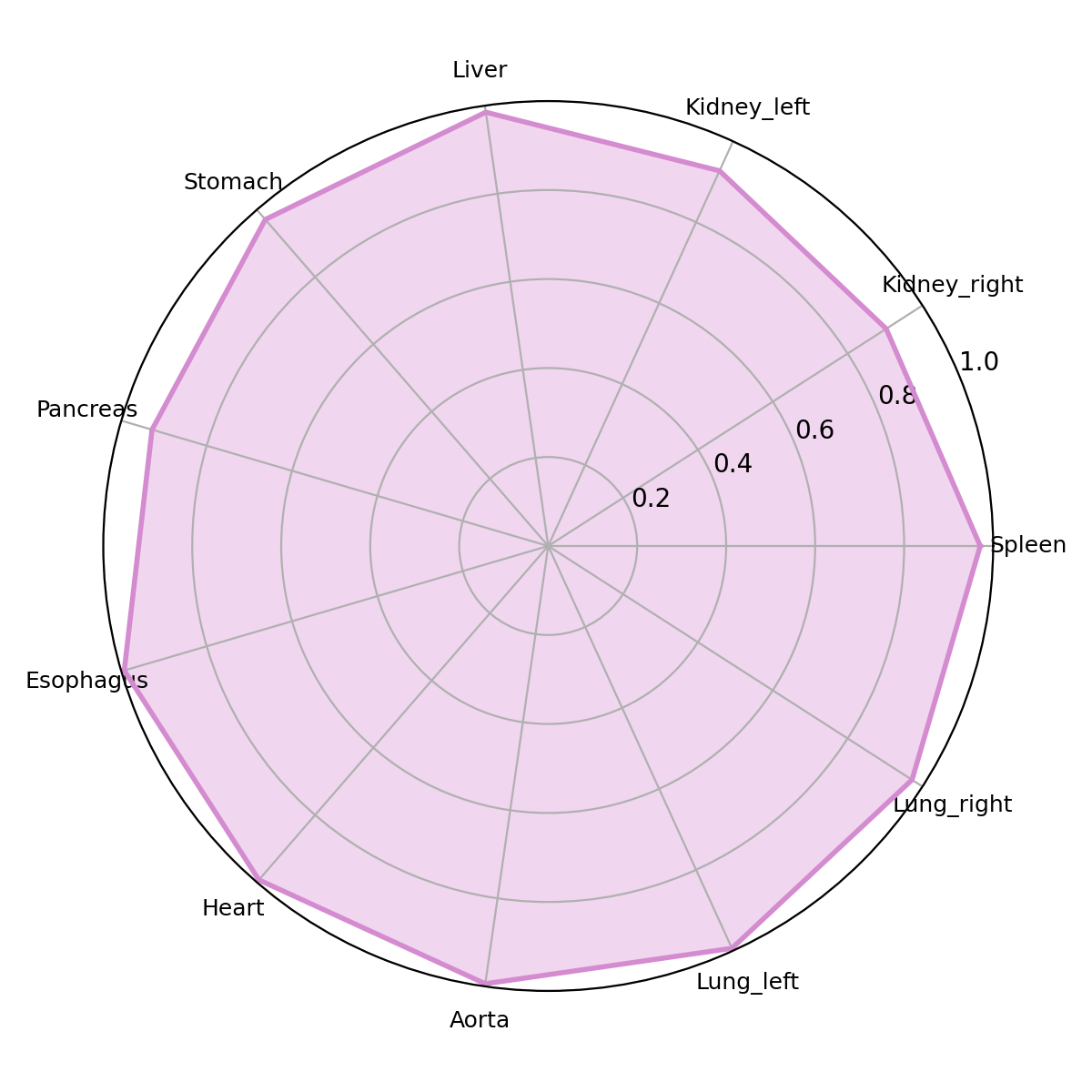}
        \caption{Slice = 256}
        \label{fig:radar_256}
    \end{subfigure}

    \caption{Organ-wise F1 radar plots across different numbers of input slices.}
    \label{fig:radar_all}
\end{figure*}

Additionally, in this paper, we evaluate retrieval performance on the full validation set.
However, the performance differences among baseline methods on the full set are extremely small, often below 0.001 across several metrics, making it difficult to draw reliable or meaningful conclusions from a single pass evaluation.
Therefore, in the main paper we primarily report results based on 100 bootstrap subsets, which provide a more stable, statistically robust, and discriminative comparison across methods.
For completeness, the full-set results are included in Table~\ref{tab:1500_retrieval} in this appendix.
The overall trends are fully consistent with those presented in the paper: our model outperforms all baselines across every retrieval metric.
Notably, despite the difficulty of retrieving a 200-slice volume from $1{,}500$ candidates using only a full radiology report, our method achieves a Recall@1 of 0.0998. This is remarkably high for large-scale radiology-report–to–CT retrieval and substantially exceeds existing state-of-the-art approaches.

\section{Organ-wise F1 Performance Across Varying Numbers of Input Slices}

To further examine how the amount of input 3D context influences organ-level performance, we visualize the F1 scores of 11 organs across six different slice counts (1, 16, 32, 64, 128, 256). The radar plots in Figure~\ref{fig:radar_all} reveal how prediction stability and anatomical completeness improve as more slices become available.

Across all six configurations, the overall trend is consistent with the observations made in Figure~\ref{fig:F1-curve}: When only one slice is provided, the performance distribution is highly uneven, while larger slice counts (64–256) yield more stable, high-magnitude F1 scores across all organs. This confirms that richer 3D context leads to more consistent anatomical reasoning and improved multi-organ segmentation performance.

\section{Structured Annotation}

To improve alignment between free-text radiology descriptions and our model’s structured prediction space, we employed GPT-5 Mini to convert the original free-text reports
into standardized organ-level annotations. The LLM maps each mentioned anatomical region into a predefined schema, and leverages categorical status labels (\emph{e.g.,} normal, abnormal, not examined) along with corresponding summarized findings. In clinical practice, the label \textit{not examined} implicitly suggests that the corresponding organ is generally considered healthy unless otherwise indicated.

\clearpage
\onecolumn

\begin{tcolorbox}[breakable, colback=white, colframe=black, arc=2mm, boxrule=0.5pt]

{\ttfamily
\small

\newcommand{\statusabn}{\textcolor{red}{abnormal}}
\newcommand{\statusnorm}{\textcolor{green!50!black}{normal}}
\newcommand{\statusnot}{\textcolor{gray}{not\_examined}}

\noindent\textbf{train\_7\_a\_1.json\{}\\
\quad "Adrenal gland": \{ "status": "\statusnorm", "findings": "Bilateral adrenal gland calibration was normal and no space-occupying lesion was detected." \},\\
\quad "Aorta": \{ "status": "\statusnorm", "findings": "No dilatation was detected in the thoracic aorta; calibration of thoracic main vascular structures is natural." \},\\
\quad "Brain": \{ "status": "\statusnot", "findings": "not\_examined" \},\\
\quad "Breast": \{ "status": "\statusnot", "findings": "not\_examined" \},\\
\quad "Clavicle": \{ "status": "\statusnot", "findings": "not\_examined" \},\\
\quad "Colon": \{ "status": "\statusnot", "findings": "not\_examined" \},\\
\quad "Esophagus": \{ "status": "\statusnorm", "findings": "Thoracic esophagus calibration was normal and no significant pathological wall thickening was detected." \},\\
\quad "Femur": \{ "status": "\statusnot", "findings": "not\_examined" \},\\
\quad "Gallbladder": \{ "status": "\statusnot", "findings": "not\_examined" \},\\
\quad "Gluteus muscles": \{ "status": "\statusnot", "findings": "not\_examined" \},\\
\quad "Great vessels": \{ "status": "\statusnorm", "findings": "Calibration of thoracic main vascular structures is natural; no dilatation of the thoracic aorta." \},\\
\quad "Heart": \{ "status": "\statusnorm", "findings": "Heart contour size is natural; pericardial thickening-effusion was not detected." \},\\
\quad "Hip/Pelvis": \{ "status": "\statusnot", "findings": "not\_examined" \},\\
\quad "Humerus": \{ "status": "\statusnot", "findings": "not\_examined" \},\\
\quad "Iliopsoas": \{ "status": "\statusnot", "findings": "not\_examined" \},\\
\quad "Inferior vena cava": \{ "status": "\statusnot", "findings": "not\_examined" \},\\
\quad "Kidney": \{ "status": "\statusnot", "findings": "not\_examined" \},\\
\quad "Liver": \{ "status": "\statusabn", "findings": "Liver size increased (hepatomegaly). Other upper abdominal sections within the examination area are normal." \},\\
\quad "Lung": \{ "status": "\statusabn", "findings": "Pleuroparenchymal sequelae increase in density and paracicatricial bronchiectasis in the right upper lobe; increased pleuroparenchymal sequelae density in the left lower lobe; calcified nonspecific parenchymal nodules 3–3.5 mm in both lungs; no pleural effusion." \},\\
\quad "Lymph nodes": \{ "status": "\statusabn", "findings": "A few calcified lymph nodes/nodules (3.5 mm left middle lobe, 3 mm right upper lobe); no pathological mediastinal/hilar lymphadenopathy." \},\\
\quad "Pancreas": \{ "status": "\statusnot", "findings": "not\_examined" \},\\
\quad "Paraspinal muscles": \{ "status": "\statusnot", "findings": "not\_examined" \},\\
\quad "Pericardium": \{ "status": "\statusnorm", "findings": "Pericardial thickening-effusion was not detected." \},\\
\quad "Pleura": \{ "status": "\statusnorm", "findings": "Bilateral pleural thickening-effusion was not detected." \},\\
\quad "Portal vein and splenic vein": \{ "status": "\statusnot", "findings": "not\_examined" \},\\
\quad "Prostate": \{ "status": "\statusnot", "findings": "not\_examined" \},\\
\quad "Pulmonary vessels": \{ "status": "\statusnot", "findings": "not\_examined" \},\\
\quad "Ribs": \{ "status": "\statusnot", "findings": "not\_examined" \},\\
\quad "Scapula": \{ "status": "\statusnot", "findings": "not\_examined" \},\\
\quad "Skull": \{ "status": "\statusnot", "findings": "not\_examined" \},\\
\quad "Small intestine": \{ "status": "\statusnot", "findings": "not\_examined" \},\\
\quad "Spinal cord": \{ "status": "\statusnot", "findings": "not\_examined" \},\\
\quad "Spine/Vertebrae": \{ "status": "\statusnorm", "findings": "No lytic-destructive lesion was detected in bone structures." \},\\
\quad "Spleen": \{ "status": "\statusnot", "findings": "not\_examined" \},\\
\quad "Sternum": \{ "status": "\statusnot", "findings": "not\_examined" \},\\
\quad "Stomach": \{ "status": "\statusnot", "findings": "not\_examined" \},\\
\quad "Thyroid gland": \{ "status": "\statusnot", "findings": "not\_examined" \},\\
\quad "Trachea": \{ "status": "\statusnorm", "findings": "Trachea and both main bronchi are open; no occlusive pathology was detected." \},\\
\quad "Urinary bladder": \{ "status": "\statusnot", "findings": "not\_examined" \},\\
\quad "general": "Mediastinal structures were evaluated as suboptimal since the examination was unenhanced; an intravascular catheter projects superiorly to the vena cava."\\
\textbf{\}}
}

\end{tcolorbox}

\end{document}